\documentclass[journal, 10pt]{IEEEtran}
\usepackage{etoolbox}
\usepackage{dblfloatfix}

\makeatletter
\patchcmd{\@thm}{\trivlist}{\trivlist\setlength{\itemsep}{0pt}}{}{}
\makeatother
\usepackage{ulem}
\usepackage{algorithm}
\usepackage{algorithmic}
\IEEEoverridecommandlockouts

\usepackage{cite}
\usepackage{amsmath,amssymb,amsfonts}
\usepackage{amsthm}
\usepackage{amsmath}
\allowdisplaybreaks[4]
\usepackage{algorithmic}
\usepackage{graphicx}
\usepackage{caption}
\usepackage{subcaption}
\usepackage{makecell}
\usepackage{textcomp}
\usepackage{xcolor}
\usepackage{booktabs}
\usepackage{xr-hyper}
\usepackage{lipsum}

\usepackage{booktabs}
\usepackage{multirow}
\newtheorem{theorem}{Theorem}
\newtheorem{lemma}{Lemma}
\newtheorem{corollary}{Corollary}
\newtheorem{assumption}{Assumption}

\def\BibTeX{{\rm B\kern-.05em{\sc i\kern-.025em b}\kern-.08em
    T\kern-.1667em\lower.7ex\hbox{E}\kern-.125emX}}
\makeatletter
\newtheorem*{rep@lemma}{\rep@title}
\newcommand{\rep@title}{}

\newenvironment{replemma}[1]{%
  \renewcommand{\rep@title}{Lemma~\ref{#1}}%
  \begin{rep@lemma}
}{%
  \end{rep@lemma}
}
\makeatother

\makeatletter
\newtheorem*{rep@corollary}{\rep@corotitle}
\newcommand{\rep@corotitle}{}

\newenvironment{repcorollary}[1]{%
  \renewcommand{\rep@corotitle}{Corollary~\ref{#1}}%
  \begin{rep@corollary}
}{%
  \end{rep@corollary}
}
\makeatother

\makeatletter
\newtheorem*{rep@theorem}{\rep@theotitle}
\newcommand{\rep@theotitle}{}

\newenvironment{reptheorem}[1]{%
  \renewcommand{\rep@theotitle}{Theorem~\ref{#1}}%
  \begin{rep@theorem}
}{%
  \end{rep@theorem}
}
\makeatother
\begin{document}
\title{Federated Self-Supervised Modulation Classification under Non-IID and Imbalanced Data}

\author{
    \IEEEauthorblockN{
        Usman Akram, 
        Yiyue Chen, 
        and Haris Vikalo
    }
    \\
    \IEEEauthorblockA{Department of Electrical and Computer Engineering, University of Texas at Austin, Austin, TX, USA\\
    usman.akram@austin.utexas.edu, yiyuechen@utexas.edu, hvikalo@ece.utexas.edu}
}
\maketitle

\begin{abstract}
Automatic modulation classification (AMC) is a core enabler of cognitive wireless systems, providing spectrum awareness and supporting adaptive communication at the network edge. However, training AMC models on centrally aggregated data incurs high communication overhead, raises privacy concerns, and often lacks robustness to real-world conditions. We propose FedSSL-AMC, a federated self-supervised framework for learning AMC models from sparsely labeled, distributed I/Q time-series data. Participating clients collaboratively train a causal, time-dilated CNN encoder using triplet-loss self-supervision on unlabeled signals, followed by lightweight local SVMs trained on limited labeled samples. This enables communication-round-efficient, robust representation learning under class imbalance and channel variability. We establish convergence guarantees for a proximal variant of the encoder-training procedure and derive a separability bound for the downstream classifier under feature noise. Experiments on synthetic and over-the-air datasets demonstrate improvements over supervised FL baselines across all three datasets and nearly all evaluated settings involving heterogeneous SNRs, carrier-frequency offsets, and non-IID label distributions.
\end{abstract}


\section{Introduction}

Next-generation wireless networks must support high-density connectivity and dynamic spectrum access as IoT growth increases demand on limited spectrum resources. Achieving this goal requires spectrum intelligence, i.e., real-time awareness of channel conditions and signal types to enable interference mitigation, adaptive modulation, and opportunistic spectrum access. Automatic modulation classification (AMC), an essential tool for acquiring spectral awareness, has received considerable attention in recent years, with numerous deep-learning approaches proposed~\cite{lin2017application,tu2018semi,hu2019deep,wang2020lightamc,wang2020deep,qi2020automatic,huang2020visualizing,wang2020automatic,ke2021real}. However, most of these methods rely on centralized training, which incurs substantial bandwidth costs, raises privacy concerns, and often yields models that lack robustness to channel mismatch. Federated learning (FL) offers an alternative by enabling distributed training across edge clients without requiring raw I/Q samples to be shared~\cite{mcmahan2017communication,shi2020signal,zhao2022semisupervised}. However, the performance of standard FL methods can degrade under class imbalance and non-IID client data, both of which commonly arise in AMC due to variability in spectrum conditions and deployment settings. These challenges are further exacerbated by the scarcity of labeled data and the relative abundance of unlabeled I/Q streams in practical wireless systems.

An early deep-learning approach to AMC~\cite{zhang2019automatic} transforms I/Q sequences into time-frequency images, including Smoothed Pseudo Wigner-Ville and Born-Jordan distributions, and applies convolutional neural networks (CNNs) to fuse them with handcrafted features. LightAMC~\cite{wang2020lightamc} improves the efficiency of such frameworks through compressive-sensing-based pruning of CNN models. SplitAMC~\cite{park2023splitamc} transmits intermediate activations (``smashed data'') and gradients instead of raw signals to reduce latency and improve robustness. To address class imbalance,~\cite{gao2023modulation} augments underrepresented classes and combines CNNs with LSTMs for temporal modeling.

Federated learning methods for AMC include FedeAMC~\cite{wang2021federated}, which employs a balanced cross-entropy loss to mitigate class imbalance, and FedBKD~\cite{qi2022fedbkd}, which uses bidirectional knowledge distillation to address data and model heterogeneity. Other approaches include federated incremental learning with private local classes~\cite{qi2022collaborative} and personalized FL based on MetaSGD~\cite{rahman2024improved,li2017meta}, which optimizes client-specific learning rates. Several complementary lines of work focus on communication-efficient FL over wireless networks. For example,~\cite{dist_survey} surveys methods for reducing uplink and downlink costs;~\cite{RSPBLMUC} exploits correlations between gradient subspaces to compress local updates;~\cite{10648926} uses unsourced massive access and approximate message passing for over-the-air model aggregation; and~\cite{10436088} develops schedule-aware FL over mixed terrestrial-satellite links with heterogeneous latency and bandwidth.

While these works investigate model compression and/or personalization, none explore self-supervised representation learning for AMC in federated settings. Self-supervised learning (SSL) is a natural fit for RF signal inference because labeled data are scarce while unlabeled I/Q sequences are abundant. Prior centralized studies have shown strong promise for SSL-based AMC~\cite{liu2021self,yang2025revolutionizing}. In federated settings, SSL not only reduces reliance on labeled data but can also mitigate distribution drift among heterogeneous clients~\cite{zhuangdivergence,chen2024fed}. Building on these insights, we develop a federated self-supervised approach for AMC in realistic deployment settings.

Specifically, to address the limitations of supervised FL under label scarcity and data heterogeneity, we propose a federated self-supervised framework for AMC on distributed I/Q data. Our approach, \textbf{FedSSL-AMC}, collaboratively pretrains a shared encoder on unlabeled I/Q streams using a contrastive objective, promoting consistent latent representations across clients. A causal, time-dilated CNN serves as the backbone, capturing long-range temporal dependencies in modulation patterns. To limit annotation and communication overhead, each client trains a lightweight support vector machine locally using a small labeled set.

Key components of our approach and contributions include:
\begin{itemize}
    \item \textbf{Self-supervised pretraining with lightweight adaptation}: We develop a causal, time-dilated CNN trained via federated averaging using a triplet-loss objective on unlabeled I/Q sequences. Each client then fits a lightweight support vector machine (SVM) on its own labeled subset for personalized classification. While similar federated SSL schemes with SVM heads have been proposed in other domains, this is the first demonstration of their effectiveness for modulation classification.

    \item \textbf{Theoretical analysis}: We provide convergence guarantees for a proximal variant of the federated self-supervised encoder trained on noisy signals, characterizing how signal noise, client heterogeneity, and inexact local updates affect descent of the global objective. For smooth encoder activations, the noise-dependent terms vanish as the SNR increases. For the downstream classifier, we derive a separability condition that quantifies the SNR required for reliable classification.

    \item \textbf{Empirical results}: We benchmark FedSSL-AMC on both synthetic and over-the-air AMC datasets. It outperforms supervised FL baselines across all three datasets and nearly all evaluated settings, including heterogeneous SNRs, carrier-frequency offsets, and non-IID label partitions.

    \item \textbf{Resource footprint}: We quantify model size, MFLOPs, and communication-round requirements, characterizing the tradeoff between classification accuracy, local computation, and communication cost.
\end{itemize}

To our knowledge, this is the first work to develop a self-supervised learning (SSL) framework for automatic modulation classification in a federated setting. The proposed approach addresses key challenges in cognitive wireless systems, including label scarcity, client heterogeneity, and privacy constraints, that centralized SSL methods are not well-suited to handle. While our primary focus is representation learning under heterogeneous and label-sparse conditions, FedSSL-AMC also reduces the number of communication rounds required during training. Consistent with prior observations that larger local datasets can improve FL efficiency~\cite{yangachieving}, our experiments in Section~\ref{4B} show that, on the synthetic dataset, FedSSL-AMC achieves higher accuracy than the supervised FL baselines using 10 communication rounds, compared with the 1{,}000-round protocol adopted from FedeAMC, at the expense of greater local computation per round.

The remainder of the paper is organized as follows. Section~II introduces the proposed FedSSL-AMC framework. Section~III presents the theoretical analysis of encoder convergence and downstream separability under noise and distribution shift. Section~IV reports experimental results on synthetic and over-the-air datasets, and Section~V concludes the paper.\footnote{A preliminary version of this work appeared in the 2025 IEEE Workshop on Signal Processing Advances in Wireless Communications (SPAWC)~\cite{11143450}. This journal version substantially extends the workshop paper through new theoretical results, a generalized framework, and additional experiments.}

\label{sec:intro}



\section{Federated Self-Supervised Learning for AMC}

The proposed method addresses two major challenges in federated AMC: non-IID data across clients and limited access to labeled I/Q samples. To address both, we decouple representation learning from downstream classification: a shared encoder is pretrained on unlabeled data through federated self-supervised learning, after which each client locally trains a lightweight classifier on a small labeled subset.

\subsection{Representation learning on I/Q sequences}

\begin{figure}[!htbp]
    \centering
    \begin{subfigure}[b]{\linewidth}
        \centering
        \includegraphics[width=0.85\linewidth]{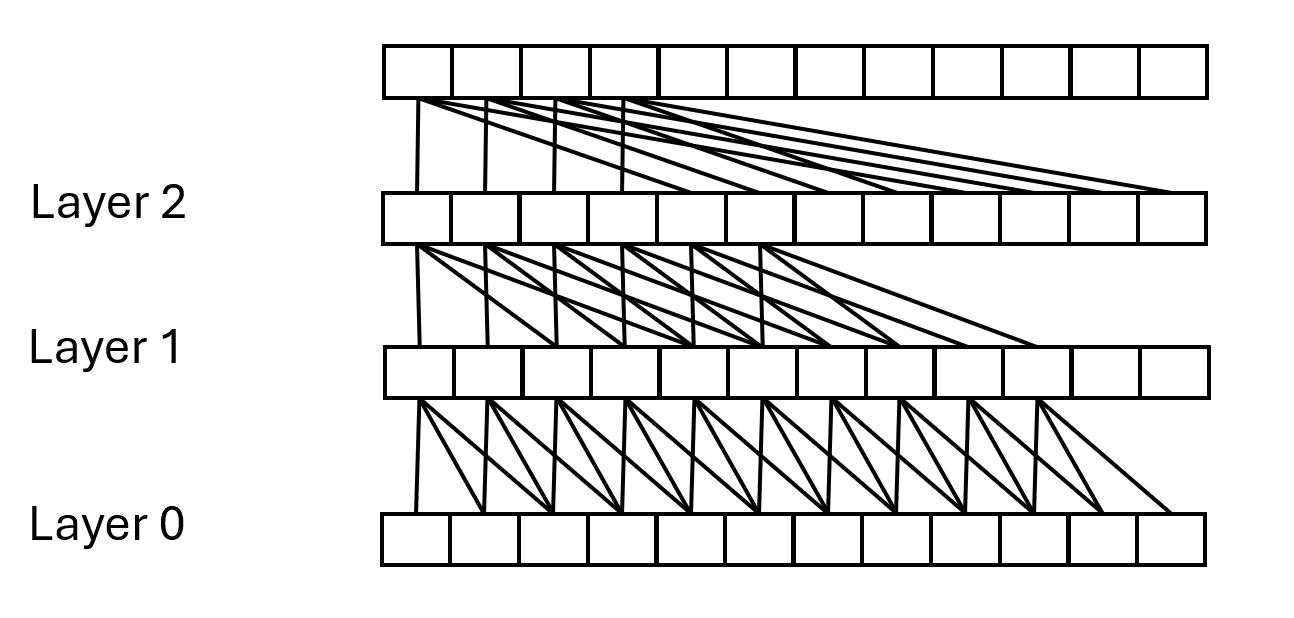}
        \caption{\it Growth of the receptive field.}
        \label{fig:1a}
    \end{subfigure}
    \hfill
    \vspace{-0.1in}
    \begin{subfigure}[b]{\linewidth}
        \centering
        \includegraphics[width=0.475\linewidth]{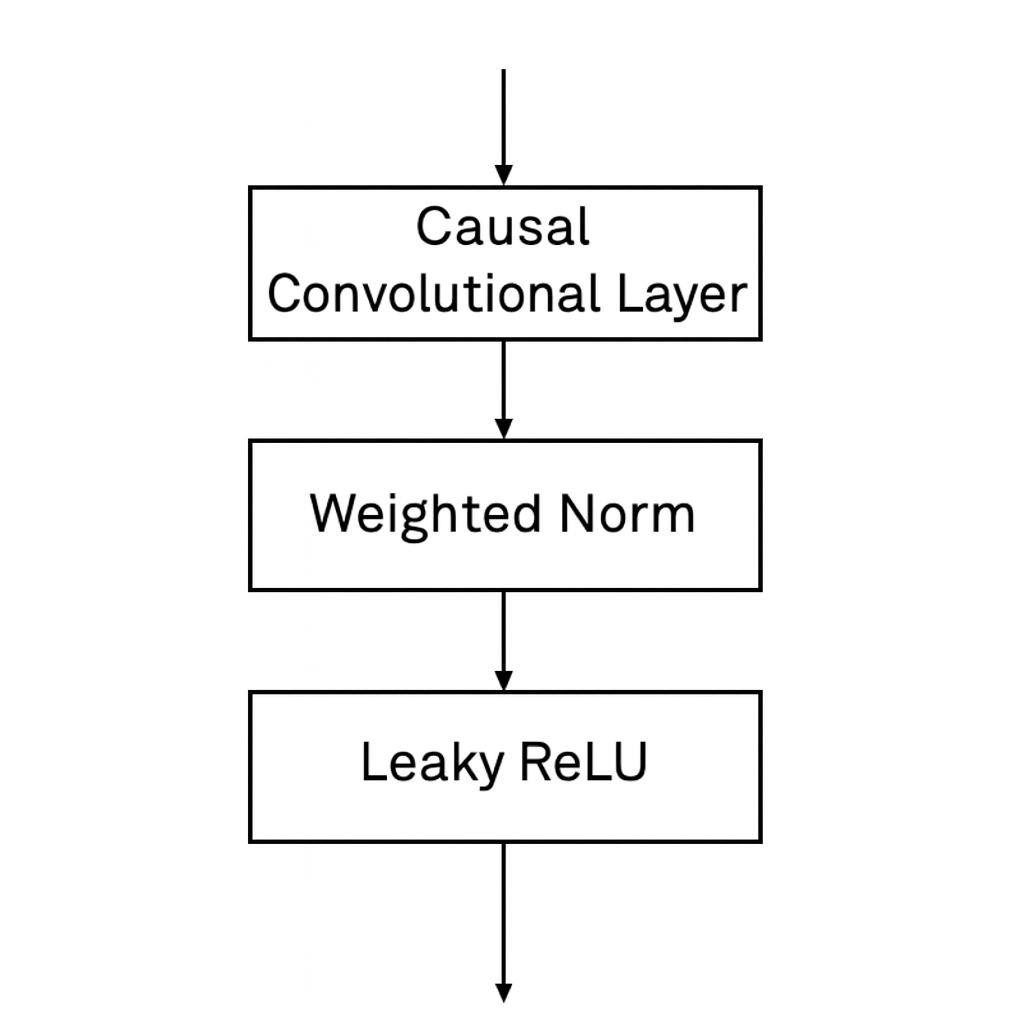}
        \caption{\it Convolutional block with weighted normalization and leaky ReLU activation.}

        \label{fig:1b}
    \end{subfigure}
    \caption{\it Time-dilated stacked convolutional layers. In layer $l$, the neurons connected to a neuron in layer $l+1$ are spaced $2^l$ apart, resulting in a receptive field that expands exponentially with network depth.}
    \label{fig:1}
\end{figure}

To learn representations from I/Q sequences, we map each time-series input in $\mathbb{R}^{2 \times T}$ to a compact feature vector in $\mathbb{R}^{d}$, where ideally $d \ll 2T$. We adopt the self-supervised time-series representation learning framework of~\cite{franceschi2019unsupervised}, which uses a causal convolutional neural network with time dilation. The encoder architecture, illustrated in Fig.~\ref{fig:1}, consists of stacked convolutional layers in which a neuron in the $k^{\text{th}}$ layer connects to neurons in the preceding layer with spacing $2^{k-1}$. This exponentially dilated causal design offers the following benefits:

\begin{enumerate}
    \item \textbf{Long-range temporal modeling.} Exponentially dilated causal convolutions capture long-range dependencies through an expanded receptive field.

    \item \textbf{Scalability and efficiency.} The CNN architecture supports efficient parallelization and scales to long sequences.

    \item \textbf{Online inference capability.} The causal structure ensures that adding a new input element at test time requires recomputing only a small portion of the computational graph, enabling efficient online deployment.

\end{enumerate}

As shown in \cite{franceschi2019unsupervised}, self-supervised representation learning for time-series data can be achieved using a contrastive triplet loss. Motivated by \textit{word2vec} \cite{mikolov2013distributed}, the main idea is to bring similar samples closer in the feature space while pushing dissimilar ones apart. Specifically, for each randomly sampled reference sample, a positive sample should lie close in the learned embedding space, while $K$ negative samples should be far from it. Formally, let $f(x;\Theta)$ denote the representation of input $x$ produced by a deep encoder parameterized by $\Theta$. Omitting the parameterization for brevity, the triplet loss is
\begin{align}
\mathcal{L}(x^{ref}, x^{pos}, \{x^{neg}_k\}_{k=1}^K) 
&= -\log\left(\sigma\left(f(x^{ref})^T f(x^{pos})\right)\right) \nonumber \\
& \!\!\! - \sum_{k=1}^K \log\left(\sigma\left(-f(x^{ref})^T f(x^{neg}_k)\right)\right),
\label{eq1}
\end{align}
where $\sigma(\cdot)$ is the sigmoid function. The loss encourages the inner product between $f(x^{ref})$ and $f(x^{pos})$ to be large and positive, while penalizing similarity with each $f(x^{neg}_k)$. When $f(x^{ref})^T f(x^{pos}) \gg 1$, the first term vanishes, i.e.,
\[
\log\left(\sigma(f(x^{ref})^T f(x^{pos}))\right) \approx 0.
\]
Similarly, when $f(x^{ref})^T f(x^{neg}_k) \ll -1$, it holds that
\[
\log\left(\sigma(-f(x^{ref})^T f(x^{neg}_k))\right) \approx 0.
\]
Thus, minimizing the triplet loss aligns related samples while separating unrelated ones in the learned representation space.

In our implementation, $x^{pos}$ is a randomly sampled subsequence from the same I/Q sequence as $x^{ref}$, while each $x^{neg}_k$ is drawn from a different I/Q sequence. This construction exploits the local temporal consistency of wireless signals, since segments from the same sequence are more likely to share modulation patterns, and enables the model to learn meaningful representations without labels.

\subsection{Federated self-supervised learning}

Self-supervised learning addresses limited labeled data by decoupling representation learning from downstream classification. In the federated setting, clients collaboratively train a shared feature encoder using standard federated averaging (FedAvg)~\cite{mcmahan2017communication} without exchanging raw I/Q data, and then train lightweight classifiers locally on small labeled subsets. This forms the basis of \textbf{FedSSL-AMC}, which combines federated self-supervised representation learning with client-specific classification under class imbalance and data heterogeneity. We use a time-dilated causal CNN as the encoder and train it on unlabeled I/Q streams using the triplet loss in~\eqref{eq1}. After encoder training, each client fits a support vector machine (SVM) on its labeled data. Algorithm~\ref{alg1} summarizes the full procedure.

\begin{algorithm}
\caption{FedSSL-AMC} \label{alg1}
\begin{algorithmic}[1]
\STATE \textbf{Input:} Number of rounds $T_{\text{rounds}}$, number of clients $C$, initial encoder parameters $\boldsymbol{\Theta}^0$ (causal CNN with time dilation)
\FOR{each round $t = 1, 2, \dots, T_{\text{rounds}}$}
    \FOR{each client $c = 1, 2, \dots, C$}
        \STATE Client $c$ downloads current global encoder parameters $\boldsymbol{\Theta}^{t-1}$
        \STATE Client $c$ updates encoder parameters $\boldsymbol{\Theta}_c^t$ using local unlabeled I/Q data and triplet loss \eqref{eq1}
        \STATE Client $c$ uploads updated parameters $\boldsymbol{\Theta}_c^t$ to the server
    \ENDFOR
    \STATE Server aggregates updates according to
    \[
    \boldsymbol{\Theta}^{t} = \sum_{c=1}^C \frac{|\mathcal{D}_c|}{|\mathcal{D}|} \boldsymbol{\Theta}_c^t,
    \]
    where $|\mathcal{D}_c|$ is the number of local unlabeled samples on client $c$, and $|\mathcal D|=\sum_{c=1}^{C}|\mathcal D_c|$.
\ENDFOR

\FOR{each client $c = 1, 2, \dots, C$}
    \STATE Client $c$ receives final encoder $\boldsymbol{\Theta}^{T_{\text{rounds}}}$ from the server
    \STATE Client $c$ trains a local SVM classifier $\mathcal{T}_c$ on encoder outputs $f(x;\boldsymbol{\Theta}^{T_{\text{rounds}}})$ for labeled inputs $x$
\ENDFOR
\end{algorithmic}
\end{algorithm}

\label{sec:proposed method}
\section{Analysis of Contrastive Encoder Training}

This section provides theoretical guarantees for the proposed federated self-supervised learning framework. We first establish that the causal CNN encoder is Lipschitz continuous with respect to both its input signals and model parameters (Section~\ref{convergence_rep}), which allows us to bound the sensitivity of the contrastive loss to additive channel noise. We then characterize how the expected gradient norm scales with the signal-to-noise ratio (SNR), quantifying the effect of noise on encoder training. For analytical tractability, the convergence analysis considers a proximal variant of the local contrastive objective rather than the FedAvg implementation in Algorithm~\ref{alg1}. Under full client participation, we characterize how signal noise, client heterogeneity, and inexact local updates affect descent of the corresponding global objective, with tighter SNR-dependent guarantees obtained for smooth encoder activations. Finally, in Section~\ref{SVManalysis}, we derive a separability condition for the downstream SVM classifier trained on noisy encoder features.

\subsection{Convergence analysis of representation learning}\label{convergence_rep}

We begin by introducing the notation and assumptions used to analyze the federated self-supervised representation learning process. Let us denote the input sequence length by $T$ and the output length at the $l^{\text{th}}$ encoder layer by $T_l$. Let $P$ and $\gamma$ denote the expected signal power per channel and the SNR, respectively. The following assumptions ensure that the encoder output remains bounded and make the analysis tractable.

\begin{assumption}
The magnitude of every encoder weight is bounded by a constant $R$.
\end{assumption}

We analyze the proposed federated self-supervised framework with a causal CNN encoder using either ReLU or Softplus activations. The analysis considers contrastive triplet-loss training under heterogeneous client data and noisy signal conditions. Using Lipschitz continuity of the encoder and bounds on the noise-induced gradient perturbation, we characterize how signal noise, client heterogeneity, and inexact local updates affect descent of the global objective for the proximal variant considered in the analysis.

\subsubsection{Results for a causal CNN encoder with ReLU activation}

For the causal CNN encoder trained with triplet loss and ReLU
activation, we derive bounds on the expected gradient norm as a
function of the SNR. Building on these results, we adapt the convergence
analysis of~\cite{li2020federated} to the proximal variant introduced
in Section~III, characterizing descent of the global objective under
full client participation, noisy signals, and client heterogeneity.
Full proofs are deferred to the supplementary material. We begin with
the following assumptions.

\begin{assumption}
 The magnitude of each encoder input channel is clipped at {\color{black}{$\zeta \sqrt{P}$}}, where $\zeta \in \mathbb{R}_{++}$ and $P$ denotes the expected signal power per channel.
\end{assumption}

\begin{assumption}
The encoder is a $\Psi$-layer causal CNN with kernel size $\chi$, unit stride, $\hat{c}$ channels per layer, and ReLU activations.
\end{assumption}

The results extend to settings where the kernel size, stride, and number of channels vary across layers, provided the activation functions are $L_{\mathrm{act}}$-Lipschitz for some constant $L_{\mathrm{act}}$.

We express the time-dilated convolution as a matrix multiplication, so that the output activation of layer $l$ becomes
\begin{align}
    \bar{r}_{l} = \operatorname{ReLU}(\bar{\Theta}_l \bar{r}_{l-1}).
    \label{eqnactivation}
\end{align}
The input to layer $l$, denoted by $\bar{r}_{l-1}$, is defined as
\begin{align}
\bar{r}_{l-1}
=
\begin{bmatrix}
r_{l-1}^{(1,1)}, \dots, r_{l-1}^{(1,T_{l-1})}, \mathbf{0}_{\text{pad}}, \dots,
r_{l-1}^{(\hat{c},1)}, \dots, r_{l-1}^{(\hat{c},T_{l-1})}, \mathbf{0}_{\text{pad}}
\end{bmatrix}^{T},
\end{align}
where $T_{l-1}$ denotes the input sequence length per channel at layer $l-1$, and $\mathbf{0}_{\text{pad}}$ denotes the zero-padding required to represent the dilated convolution structure.

The matrix $\bar{\Theta}_l$ denotes the weight matrix at layer $l$ and is constructed as
\begin{align}
\bar{\Theta}_l =
\begin{bmatrix}
\bar{\Theta}^{(1,1)}_l & \bar{\Theta}^{(1,2)}_l & \dots & \bar{\Theta}^{(1,\hat{c})}_l \\
\bar{\Theta}^{(2,1)}_l & \bar{\Theta}^{(2,2)}_l & \dots & \bar{\Theta}^{(2,\hat{c})}_l \\
\vdots & \vdots & \ddots & \vdots \\
\bar{\Theta}^{(\hat{c},1)}_l & \bar{\Theta}^{(\hat{c},2)}_l & \dots & \bar{\Theta}^{(\hat{c},\hat{c})}_l
\end{bmatrix}.
\end{align}
Each submatrix $\bar{\Theta}^{(i,j)}_l$ represents the convolutional mapping from input channel $j$ to output channel $i$ and has the block Toeplitz form
\begin{align}
\bar{\Theta}^{(i,j)}_l =
\begin{bmatrix}
\bar{\Theta}^{(i,j,1)}_l & \dots & \bar{\Theta}^{(i,j,\chi)}_l & 0 & \dots & 0 \\
\vdots & \ddots & \vdots & \vdots & \ddots & \vdots \\
0 & \dots & 0 & \bar{\Theta}^{(i,j,1)}_l & \dots & \bar{\Theta}^{(i,j,\chi)}_l
\end{bmatrix},
\end{align}
where the shifts between rows reflect the time-dilation structure. Each row block is defined as
\begin{align}
\bar{\Theta}^{(i,j,k)}_l =
\begin{bmatrix}
\bar{\theta}^{(i,j,k)}_l & \mathbf{0}_{2^l-1}^T
\end{bmatrix},
\end{align}
where $\mathbf{0}_{2^l-1}^T$ denotes a row vector of $2^l-1$ zeros.

We assume that the spectral norm of each layer weight matrix is uniformly bounded.

\begin{assumption}
\label{SpectralNormBound}
For all $l = 1,\dots,\Psi$, the spectral norm of $\bar{\Theta}_l$ satisfies
\[
\|\bar{\Theta}_l\|_2
=
\sup_{x\neq 0}
\frac{\|\bar{\Theta}_l x\|_2}{\|x\|_2}
\le R_{\mathrm{mat}} .
\]
\end{assumption}

Under Assumption~\ref{SpectralNormBound}, the encoder is Lipschitz continuous with respect to its input.

\begin{lemma}
\label{input_enc_lip}
The encoder function
$f(x;\bar{\Theta}_1,\dots,\bar{\Theta}_\Psi)$ is $L_x$-Lipschitz
with respect to $x$, where $L_x=(R_{\mathrm{mat}})^\Psi$; that is,
\[
\left\|
f(x;\bar{\Theta}_1,\dots,\bar{\Theta}_\Psi)
-
f(x';\bar{\Theta}_1,\dots,\bar{\Theta}_\Psi)
\right\|_2
\leq
L_x\|x-x'\|_2.
\]
\end{lemma}

For the remainder of this section, let $\Theta$ denote the stacked
parameter vector
$\Theta =
\begin{bmatrix}
\bar{\theta}_1^T & \dots & \bar{\theta}_\Psi^T
\end{bmatrix}^T$.
To establish Lipschitz continuity with respect to the model parameters,
we rewrite~\eqref{eqnactivation} as
\begin{align}
\bar{r}_l
=
\operatorname{ReLU}
\left(
\bar{\mathbf{R}}_{l-1}\bar{\theta}_l
\right),
\label{eqnactivationparam}
\end{align}
where
\begin{align}
\bar{\theta}_l
=
\begin{bmatrix}
(\bar{\theta}_l^{(1,1)})^T &
\dots &
(\bar{\theta}_l^{(1,\hat{c})})^T &
\dots &
(\bar{\theta}_l^{(\hat{c},1)})^T &
\dots &
(\bar{\theta}_l^{(\hat{c},\hat{c})})^T
\end{bmatrix}^T,
\end{align}
with
\begin{align}
\bar{\theta}_l^{(i,j)}
=
\begin{bmatrix}
\bar{\theta}_l^{(i,j,1)} &
\dots &
\bar{\theta}_l^{(i,j,\chi)}
\end{bmatrix}^T.
\end{align}
The corresponding block-diagonal activation matrix is
\begin{align}
\bar{\mathbf{R}}_l
=
\operatorname{diag}
\left(
\underbrace{
\hat{\mathbf{R}}_l,\dots,\hat{\mathbf{R}}_l
}_{\hat{c}\ \text{times}}
\right),
\end{align}
where
$\hat{\mathbf{R}}_l
=
\begin{bmatrix}
\bar{\mathbf{R}}_{1,l} &
\bar{\mathbf{R}}_{2,l} &
\dots &
\bar{\mathbf{R}}_{\hat{c},l}
\end{bmatrix}$,
and
\begin{align}
\bar{\mathbf{R}}_{i,l}
=
\begin{bmatrix}
r_{l-1}^{(i,1)}
&
r_{l-1}^{(i,1+2^l)}
&
\dots
&
r_{l-1}^{(i,1+2^l(\chi-1))}
\\
r_{l-1}^{(i,2)}
&
r_{l-1}^{(i,2+2^l)}
&
\dots
&
r_{l-1}^{(i,2+2^l(\chi-1))}
\\
\vdots & \vdots & \ddots & \vdots
\\
r_{l-1}^{(i,T_{l-1}-2^l(\chi-1))}
&
r_{l-1}^{(i,T_{l-1}-2^l(\chi-2))}
&
\dots
&
r_{l-1}^{(i,T_{l-1})}
\end{bmatrix}.
\end{align}

This formulation allows us to establish Lipschitz continuity of the encoder with respect to its model parameters.

\begin{lemma}\label{lem_param_lip}
The encoder function
\[
f(x;\bar{\Theta}_1,\dots,\bar{\Theta}_\Psi)
\equiv
f(x;\bar{\theta}_1,\dots,\bar{\theta}_\Psi)
\]
is $L_{\theta}(l)$-Lipschitz with respect to the parameter vector $\bar{\theta}_l$, where
\color{black}\[
L_{\theta}(l)
=
\sqrt{T_l\hat{c}\chi}\,\zeta\sqrt{P}\,
(\hat{c}\chi R)^{l-1}
(R_{\mathrm{mat}})^{\Psi-l}.
\]
\end{lemma}

We can immediately use the above lemma to bound the norm of the Jacobian of the encoder function with respect to the encoder parameters.

\begin{corollary}\label{col_jac_bound}
Wherever it exists, the Jacobian of the encoder function
$f(x;\bar{\theta}_1,\dots,\bar{\theta}_\Psi)$ with respect to
$\bar{\theta}_l$ has spectral norm bounded by $L_{\theta}(l)$; formally,
\[
\left\|J_{\bar{\theta}_l}(x)\right\|_2
\leq
L_{\theta}(l).
\]
\end{corollary}

Given the preceding assumptions and lemmas, we can also bound the encoder output.
\begin{lemma} \label{enc_out_bound}
The encoder output is bounded {\color{black}{element-wise}} by
$\zeta\sqrt{P}\,(\hat{c}\chi R)^\Psi$.
\end{lemma}

A direct consequence of Lemma~\ref{enc_out_bound} is that the gradients of the loss with respect to the encoder outputs are bounded. Let
$z^{ref}=f(x^{ref};\Theta)$,
$z^{pos}=f(x^{pos};\Theta)$, and
$z^{neg}_k=f(x^{neg}_k;\Theta)$.
Then the following corollary holds.

\begin{corollary}\label{col_grad_loss_bound}
There exists a constant $G>0$ such that
\begin{align}
\left\|\nabla_{z^{ref}}
\mathcal{L}(z^{ref},z^{pos},\{z^{neg}_k\}_{k=1}^K)\right\|_2
&\leq G, \nonumber \\
\left\|\nabla_{z^{pos}}
\mathcal{L}(z^{ref},z^{pos},\{z^{neg}_k\}_{k=1}^K)\right\|_2
&\leq G, \nonumber \\
\left\|\nabla_{z^{neg}_k}
\mathcal{L}(z^{ref},z^{pos},\{z^{neg}_k\}_{k=1}^K)\right\|_2
&\leq G,
\qquad k=1,\dots,K.
\nonumber
\end{align}
\end{corollary}

Finally, we show that the triplet loss is smooth with respect to the encoder outputs.

\begin{lemma}\label{Lemma Smooth Loss}
The triplet loss
$\mathcal{L}(z^{ref},z^{pos},\{z^{neg}_k\}_{k=1}^K)$
is $\Gamma$-smooth with respect to the concatenated vector
$[z^{ref},z^{pos},z_1^{neg},\dots,z_K^{neg}]$, where
\[
\Gamma
=
\sqrt{K+2}\,
\max\{\Gamma_{ref},\Gamma_{pn}\},
\]
and
{\setlength{\abovedisplayskip}{4pt}
 \setlength{\belowdisplayskip}{4pt}
 \setlength{\abovedisplayshortskip}{2pt}
 \setlength{\belowdisplayshortskip}{2pt}
\begin{align}
\Gamma_{ref}
&=
2(T_{\Psi}\zeta)^2P(\hat{c}\chi R)^{2\Psi}K^2 +
\left(
4(T_{\Psi}\zeta)^2P(\hat{c}\chi R)^{2\Psi}+1
\right)K
\nonumber\\
&\quad+
2(T_{\Psi}\zeta)^2P(\hat{c}\chi R)^{2\Psi}+1,
\\
\Gamma_{pn}
&=
2(K+1)(T_{\Psi}\zeta)^2P(\hat{c}\chi R)^{2\Psi}+1.
\end{align}}
\end{lemma}

Let $n^{ref}$, $n^{pos}$, and $n_k^{neg}$ denote the noise in the reference, positive, and negative sequences, respectively; define
$x^{ref'}=x^{ref}+n^{ref}$,$x^{pos'}=x^{pos}+n^{pos}$, and
$x_k^{neg'}=x_k^{neg}+n_k^{neg}$.
The corresponding encoder outputs are $z^{ref'}=f(x^{ref'};\Theta)$, $z^{pos'}=f(x^{pos'};\Theta)$, $z_k^{neg'}=f(x_k^{neg'};\Theta)$. Using the Lipschitz continuity of the encoder with respect to its inputs and parameters, together with the smoothness of the triplet loss and the Jacobian bound, we obtain the following theorem. It bounds the expected $\ell_2$ norms of the gradients at a given SNR $\gamma$, where the expectation is taken over the additive signal noise.

\begin{theorem} \label{grad_SNR_bound}
Let $\gamma$ denote the SNR and $P$ the expected signal power per channel. Let $z^{ref'} = f(x^{ref'};\Theta)$, $z^{pos'} = f(x^{pos'};\Theta)$, and $z_k^{neg'} = f(x_k^{neg'};\Theta)$ denote the noise-perturbed embeddings, and let $z^{ref}$, $z^{pos}$, and $z_k^{neg}$ denote the corresponding noiseless embeddings. Under the preceding assumptions and lemmas, the following bounds hold, where the expectation is taken over the additive signal noise.

\vspace{0.05in}

\noindent
(1) The expected $\ell_2$-norm of the gradient of the triplet loss $\mathcal{L}$ with respect to the layer-$l$ encoder parameters $\bar{\theta}_l$ satisfies:
\begin{align}
& \mathbb{E}\left[ \left\| \nabla_{\bar{\theta}_l} \mathcal{L}(z^{ref'}, z^{pos'}, \{z_k^{neg'}\}_{k=1}^K) \right\|_2 \right] \nonumber \\
&\leq \left\| \nabla_{\bar{\theta}_l} \mathcal{L}(z^{ref}, z^{pos}, \{z_k^{neg}\}_{k=1}^K) \right\|_2 + 2(K+2) L_{\theta}(l) G \nonumber \\
&\quad + L_{\theta}(l) \Gamma L_x \sqrt{T(K+2)^3 \gamma^{-1} P}.
\end{align}

\vspace{0.025in}

\noindent 
(2) The corresponding expected squared $\ell_2$-norm satisfies:
\begin{align}
& \mathbb{E}\left[ \left\| \nabla_{\bar{\theta}_l} \mathcal{L}(z^{ref'}, z^{pos'}, \{z_k^{neg'}\}_{k=1}^K) \right\|_2^2 \right] \nonumber \\
&\leq \color{black} 2\left\| \nabla_{\bar{\theta}_l} \mathcal{L}(z^{ref}, z^{pos}, \{z_k^{neg}\}_{k=1}^K) \right\|_2^2 + 16(K+2)^2 L_{\theta}^2(l) G^2 \nonumber \\
&\quad + 4 L_{\theta}^2(l) \Gamma^2 L_x^2 T(K+2)^3 \gamma^{-1} P.
\end{align}

\vspace{0.025in}

\noindent 
(3) The expected $\ell_2$-norm of the gradient of the triplet loss with respect to the full encoder parameter vector $\Theta$ satisfies:
\begin{align}
& \mathbb{E}\left[ \left\| \nabla_{\Theta} \mathcal{L}(z^{ref'}, z^{pos'}, \{z_k^{neg'}\}_{k=1}^K) \right\|_2 \right] \nonumber \\
&\leq \left\| \nabla_{\Theta} \mathcal{L}(z^{ref}, z^{pos}, \{z_k^{neg}\}_{k=1}^K) \right\|_2 + 2(K+2)G \sum_{l=1}^{\Psi} L_{\theta}(l) \nonumber \\
&\quad + \Gamma L_x \sqrt{T(K+2)^3 \gamma^{-1} P} \sum_{l=1}^{\Psi} L_{\theta}(l).
\end{align}

\vspace{0.025in}

\noindent 
(4) The corresponding expected squared $\ell_2$-norm satisfies:
\begin{align}
& \mathbb{E}\left[ \left\| \nabla_{\Theta} \mathcal{L}(z^{ref'}, z^{pos'}, \{z_k^{neg'}\}_{k=1}^K) \right\|_2^2 \right] \nonumber \\
&\leq \color{black} 2\left\| \nabla_{\Theta} \mathcal{L}(z^{ref}, z^{pos}, \{z_k^{neg}\}_{k=1}^K) \right\|_2^2 \nonumber \\
&\quad + 16(K+2)^2 G^2 \sum_{l=1}^{\Psi} L_{\theta}^2(l)  \nonumber \\
&\quad   + 4 \Gamma^2 L_x^2 T(K+2)^3 \gamma^{-1} P \sum_{l=1}^{\Psi} L_{\theta}^2(l).
\end{align}
Here, $L_{\theta}(l)$ is the Lipschitz constant of the encoder with respect to its layer-$l$ parameters, $G$ bounds the gradient of the loss with respect to the encoder outputs (Corollary~\ref{col_grad_loss_bound}), $\Gamma$ is the smoothness constant of the triplet loss (Lemma~\ref{Lemma Smooth Loss}), $L_x$ is the Lipschitz constant of the encoder with respect to its inputs, $T$ is the signal length, and $\Psi$ is the number of encoder layers.
\end{theorem}

So far, we have characterized how noise perturbations in the encoder
input affect the gradient norms. We next analyze the effect of SNR on
the convergence of the global contrastive objective in a federated
setting with heterogeneous client data. This analysis requires
regularity of the local objectives with respect to the encoder
parameters.

\begin{assumption}
\label{lower_b_exists}
For every client $c$, the local objective $\mathcal{L}_c(\Theta)$ is
$L_{\mathrm{loss}}^{\mathrm{all}}$-smooth with respect to the encoder
parameter vector $\Theta$, i.e.,
\[
\|\nabla \mathcal{L}_c(\Theta)
-\nabla \mathcal{L}_c(\Theta')\|_2
\leq
L_{\mathrm{loss}}^{\mathrm{all}}
\|\Theta-\Theta'\|_2
\]
for all $\Theta$ and $\Theta'$. In addition, there exists a constant
$L_->0$ such that every local objective $\mathcal{L}_c$ is
$L_-$-weakly convex. Equivalently, wherever the Hessian exists,
\[
\nabla^2 \mathcal{L}_c(\Theta) \succeq -L_- I,
\]
where $I$ denotes the identity matrix.
\end{assumption}

Many existing convergence analyses, including~\cite{Onconvergence}, assume strong convexity of the objective. This assumption is not appropriate here because neural encoders and contrastive objectives are generally nonconvex. We therefore build on the FedProx convergence framework~\cite{li2020federated}, which introduces a proximal term in each local objective to control client drift under heterogeneous data. Although our experiments use standard federated averaging for encoder training, the analysis applies to the corresponding FedProx-style variant. The resulting guarantees therefore characterize a closely related proximal implementation rather than Algorithm~\ref{alg1} directly.

For the federated analysis, we assume full client participation. At communication round $t$, client $c$ minimizes the regularized local objective
\begin{align}
h_c(\Theta,\Theta^t)
=
\mathcal{L}_c(\Theta)
+
\frac{\mu_{\mathrm{prox}}}{2}
\|\Theta-\Theta^t\|_2^2,
\end{align}
where $\mu_{\mathrm{prox}}>0$ controls the deviation of the local model from the current global model $\Theta^t$. The corresponding local update is
\begin{align}
\hat{\Theta}_c^{t+1}
=
\arg\min_{\Theta}
h_c(\Theta,\Theta^t).
\end{align}

We define the local objective $\mathcal{L}_c(\Theta)$ as the expected contrastive loss under client $c$'s data distribution and additive noise:
\begin{align}
\mathcal{L}_c(\Theta)
=
\mathbb{E}_{x\sim\mathcal{D}_c,\,n}
\big[
\mathcal{L}(\Theta;x+n)
\big].
\end{align}
Here,
$x=\{x^{\mathrm{ref}},x^{\mathrm{pos}},x^{\mathrm{neg}}_1,\dots,x^{\mathrm{neg}}_K\}$
is drawn from the local distribution $\mathcal{D}_c$, and
$n=\{n^{\mathrm{ref}},n^{\mathrm{pos}},n^{\mathrm{neg}}_1,\dots,n^{\mathrm{neg}}_K\}$
denotes additive Gaussian noise applied independently to the samples in the triplet. The notation $\mathcal{L}(\Theta;x+n)$ is shorthand for
\[
\mathcal{L}
\big(
f(x^{\mathrm{ref}}+n^{\mathrm{ref}};\Theta),
f(x^{\mathrm{pos}}+n^{\mathrm{pos}};\Theta),
\{f(x_k^{\mathrm{neg}}+n_k^{\mathrm{neg}};\Theta)\}_{k=1}^K
\big).
\]

We also define the corresponding clean local objective as
\begin{align}
\mathcal{L}_c^{\mathrm{clean}}(\Theta)
=
\mathbb{E}_{x\sim\mathcal{D}_c}
\big[
\mathcal{L}(\Theta;x)
\big],
\end{align}
where $\mathcal{L}(\Theta;x)$ denotes the same triplet loss evaluated on uncorrupted inputs. The clean global objective is
\begin{align}
\mathcal{L}_{\mathrm{global}}(\Theta)
&=
\sum_{c=1}^C
p_c\,
\mathcal{L}_c^{\mathrm{clean}}(\Theta)
=
\mathbb{E}_{c}
\big[
\mathcal{L}_c^{\mathrm{clean}}(\Theta)
\big].
\end{align}
The weights satisfy $p_c\geq 0$ and $\sum_{c=1}^C p_c=1$; under uniform client weighting, $p_c=1/C$. If each local objective is $L_{\mathrm{loss}}^{\mathrm{all}}$-smooth, then their weighted average is also $L_{\mathrm{loss}}^{\mathrm{all}}$-smooth.

Our goal is to characterize, under full client participation, how signal
noise, client heterogeneity, and inexact local updates affect descent of
the clean global objective. To this end, we adopt a reformulation based on the FedProx analysis of~\cite{li2020federated}, which accommodates nonconvex local objectives through proximal regularization.

\begin{assumption}\label{inexact_ass}
Let $\eta_{\mathrm{ie}}\in[0,1]$ denote the inexactness level. At communication round $t$, each client computes a local solution $\Theta_c^{t+1}$ satisfying
\[
\left\|
\nabla h_c(\Theta_c^{t+1},\Theta^t)
\right\|_2
\leq
\eta_{\mathrm{ie}}
\left\|
\nabla h_c(\Theta^t,\Theta^t)
\right\|_2.
\]
The server then averages the local solutions to obtain the next global iterate:
\begin{align}
\Theta^{t+1}
=
\frac{1}{C}
\sum_{c=1}^{C}
\Theta_c^{t+1}.
\end{align}
\end{assumption}

\noindent
We next define local dissimilarity. The clean local objectives are
$B_{\mathrm{diss}}$-locally dissimilar at $\Theta$ if
\[
\mathbb{E}_c\!\left[
\left\|\nabla \mathcal{L}_c^{\mathrm{clean}}(\Theta)\right\|_2^2
\right]
\leq
B_{\mathrm{diss}}^2
\left\|\nabla \mathcal{L}_{\mathrm{global}}(\Theta)\right\|_2^2.
\]
This condition bounds the heterogeneity of the client gradients relative to the global gradient. We can now state the convergence result under full client participation and bounded local dissimilarity.

\begin{theorem} \label{fedprox_th4_adapt}
Assume that the clean local objectives are $B_{\text{diss}}$-locally dissimilar at $\Theta^t$ {\color{black}{, and that
\[
B_{\mathrm{diss}}\eta_{\mathrm{ie}}<1.
\]}} Then for any SNR $\gamma$, signal power $P$, and heterogeneous client data distributions, there exists a sufficiently large proximal coefficient $\mu_{\text{prox}} > L_-$ such that
\begin{align}
\mathcal{L}_{\text{global}}(\Theta^{t+1}) 
&\leq \mathcal{L}_{\text{global}}(\Theta^t)
- \mathcal{O}\left(\frac{1}{\mu_{\text{prox}}}\right)
\|\nabla \mathcal{L}_{\text{global}}(\Theta^t)\|^2_2 \\
&\quad + \mathcal{O}\left(\frac{1}{\mu_{\text{prox}}}\right) \cdot \mathrm{Err}(\gamma, P, B_{\text{diss}}, \eta_{\text{ie}}, \Theta^t),
\end{align}
where the error term captures the effects of signal noise, client heterogeneity, and inexact local updates. Its full expression is provided in the supplementary material.
\end{theorem}

\subsubsection{Results for a causal CNN encoder with Softplus activation}

\setlength{\abovedisplayskip}{2pt}
\setlength{\belowdisplayskip}{2pt}
\setlength{\abovedisplayshortskip}{2pt}
\setlength{\belowdisplayshortskip}{2pt}
{The bounds in Theorems~\ref{grad_SNR_bound} and~\ref{fedprox_th4_adapt} contain additive terms that do not depend on the SNR. This looseness arises because, with ReLU activations, the encoder Jacobian is not generally Lipschitz continuous with respect to the input. By replacing ReLU with a smooth activation such as Softplus, we can obtain tighter bounds in which these SNR-independent terms are eliminated. The following lemma formalizes the required Jacobian regularity.

\begin{lemma}\label{jac_liptz}
Suppose the activation function is Softplus, or another twice continuously differentiable function with bounded first and second derivatives. Under the preceding bounded-input and bounded-parameter assumptions, the Jacobian of the encoder
$f(x;\bar{\theta}_1,\ldots,\bar{\theta}_\Psi)$
with respect to $\bar{\theta}_l$ is $\Gamma_J$-Lipschitz in $x$ for some constant $\Gamma_J>0$; that is,
\[
\left\|
J_{\bar{\theta}_l}(x)
-
J_{\bar{\theta}_l}(x')
\right\|_2
\leq
\Gamma_J
\left\|x-x'\right\|_2.
\]
\end{lemma}

Since Softplus is also $1$-Lipschitz, as is ReLU, the preceding Lipschitz and boundedness results remain applicable. Together with Lemma~\ref{jac_liptz}, this allows us to derive the following theorems. Their proofs closely follow those of Theorems~\ref{grad_SNR_bound} and~\ref{fedprox_th4_adapt} and are provided in the supplementary material.

\begin{theorem} \label{grad_SNR_bound_softplus} Suppose the encoder uses Softplus, or another activation function satisfying the conditions of Lemma~\ref{jac_liptz}. Let $z^{ref'}$, $z^{pos'}$, $\{z^{neg'}_k\}_{k=1}^K$ denote the noise-perturbed embeddings, and $z^{ref}, z^{pos}, \{z^{neg}_k\}_{k=1}^K$ their noiseless counterparts. Then the excess terms in the expected gradient-norm bounds scale as $\mathcal{O}(\gamma^{-1/2}P^{1/2})$, while those in the expected squared-gradient-norm bounds scale as $\mathcal{O}(\gamma^{-1}P)$. Specifically,
\begin{align}
&\mathbb{E}\Big[ \big\| \nabla_{\bar{\theta}_l} \mathcal{L}(z^{ref'}, z^{pos'}, \{z^{neg'}_k\}) \big\|_2 \Big] \nonumber \\
&\leq \big\| \nabla_{\bar{\theta}_l} \mathcal{L}(z^{ref}, z^{pos}, \{z^{neg}_k\}) \big\|_2 + (K+2)\Gamma_J G \sqrt{T\gamma^{-1}P} \nonumber \\
&\quad + L_{\theta}(l)\Gamma L_x \sqrt{T(K+2)^3\gamma^{-1}P}, 
\end{align}
\begin{align}
&\mathbb{E}\Big[ \big\| \nabla_{\bar{\theta}_l} \mathcal{L}(z^{ref'}, z^{pos'}, \{z^{neg'}_k\}) \big\|_2^2 \Big] \nonumber \\
&\leq \color{black} 2\big\| \nabla_{\bar{\theta}_l} \mathcal{L}(z^{ref}, z^{pos}, \{z^{neg}_k\}) \big\|_2^2 + 4T\gamma^{-1}P(K+2)^2 \Gamma_J^2 G^2 \nonumber \\
&\quad + 4L_{\theta}(l)^2 \Gamma^2 L_x^2 T(K+2)^3 \gamma^{-1}P,
\end{align}
\begin{align}
&\mathbb{E}\Big[ \big\| \nabla_{\Theta} \mathcal{L}(z^{ref'}, z^{pos'}, \{z^{neg'}_k\}) \big\|_2 \Big] \nonumber \\
&\leq \big\| \nabla_{\Theta} \mathcal{L}(z^{ref}, z^{pos}, \{z^{neg}_k\}) \big\|_2 + (K+2)\Gamma_J G \Psi \sqrt{T\gamma^{-1}P} \nonumber \\
&\quad + \Gamma L_x \sqrt{T(K+2)^3 \gamma^{-1}P} \sum_{l=1}^{\Psi} L_{\theta}(l),
\end{align}
\begin{align}
&\mathbb{E}\Big[ \big\| \nabla_{\Theta} \mathcal{L}(z^{ref'}, z^{pos'}, \{z^{neg'}_k\}) \big\|_2^2 \Big] \nonumber \\
&\leq \color{black} 2\big\| \nabla_{\Theta} \mathcal{L}(z^{ref}, z^{pos}, \{z^{neg}_k\}) \big\|_2^2 + 4T\gamma^{-1}P (K+2)^2 \Psi \Gamma_J^2 G^2 \nonumber \\
&\quad + 4 \Gamma^2 L_x^2 T(K+2)^3 \gamma^{-1}P \sum_{l=1}^{\Psi} L_{\theta}(l)^2.
\end{align}
\end{theorem}

Theorem~\ref{grad_SNR_bound_softplus} shows that, with smooth activations, the gradient deviations induced by noise admit tighter bounds than in the ReLU case. Specifically, the excess first-order terms scale as $\mathcal{O}(\gamma^{-1/2}P^{1/2})$, while the excess second-order terms scale as $\mathcal{O}(\gamma^{-1}P)$, owing to the Lipschitz continuity of the encoder Jacobians. This leads to the following convergence guarantee for noisy federated training with smooth activations.

\begin{theorem}
\label{fedprox_th4_adapt_softplus}
Assume that the clean local objectives are
$B_{\mathrm{diss}}$-locally dissimilar at $\Theta^t$, and that
$B_{\mathrm{diss}}\eta_{\mathrm{ie}}<1$. Then for any signal power
$P$ and heterogeneous client data distributions, there exist a
sufficiently large proximal coefficient
$\mu_{\mathrm{prox}}>L_-$ and a sufficiently large SNR $\gamma$ such
that
\begin{align}
&\mathcal{L}_{\mathrm{global}}(\Theta^{t+1})
\leq
\mathcal{L}_{\mathrm{global}}(\Theta^t)
-
\mathcal{O}\left(\frac{1}{\mu_{\mathrm{prox}}}\right)
\|\nabla \mathcal{L}_{\mathrm{global}}(\Theta^t)\|_2^2
\nonumber\\
&\quad+
\mathcal{O}\left(\frac{1}{\mu_{\mathrm{prox}}}\right)
\cdot
\mathcal{O}\left(\gamma^{-1/2}P^{1/2}\right)
\|\nabla \mathcal{L}_{\mathrm{global}}(\Theta^t)\|_2
\nonumber\\
&\quad+
\mathcal{O}\left(\frac{1}{\mu_{\mathrm{prox}}}\right)
\cdot
\mathcal{O}\left(\gamma^{-1}P\right).
\end{align}
The hidden constants depend on the client heterogeneity and inexactness
level, while the explicit SNR-dependent terms quantify the effect of
signal noise. The full expression is provided in the supplementary
material.
\end{theorem}

The preceding results lead to three observations. First,
Theorem~\ref{grad_SNR_bound_softplus} shows that, as
$\gamma\to\infty$, the excess terms in the noisy gradient-norm bounds
vanish, so the bounds approach their clean-signal counterparts.
Second, in the same high-SNR regime,
Theorem~\ref{fedprox_th4_adapt_softplus} recovers the
full-participation FedProx-type descent behavior of Theorem~4
in~\cite{li2020federated}, whereby a sufficiently large proximal
coefficient $\mu_{\mathrm{prox}}$ ensures a decrease in the global
objective. Finally, because the constants in the noise-dependent
bounds grow with encoder depth, channel width, and kernel size, larger
architectures may require stronger proximal regularization to maintain
stable descent across communication rounds.

\subsection{Linear separability under a hard-margin SVM} \label{SVManalysis}

We now analyze the encoder outputs under hard-margin linear classification. In the noise-free setting, we assume that the features produced by the causal CNN encoder are $(\mu,\rho)$-separable. Specifically, for each input signal $r_q$, the corresponding feature vector $\phi(r_q)$ satisfies
$\|\phi(r_q)\|_2 \leq \rho$, and the labeled dataset
$\mathcal{D}=\{(\phi(r_q),y_q)\}_{q=1}^{\bar{Q}}$
contains $\bar{Q}$ binary-labeled examples with
$y_q\in\{+1,-1\}$. By definition of $(\mu,\rho)$-separability, there exist a unit-norm weight vector $\theta^*_{\mathrm{svm},w}$ and a bias $\theta^*_{\mathrm{svm},\mathrm{bias}}$ such that
\[
y_q \left( \theta^{*T}_{\mathrm{svm},w} \, \phi(r_q) + \theta^*_{\mathrm{svm},\mathrm{bias}} \right) \geq \mu.
\]
In practical AMC settings, the received signals are corrupted by noise,
and the resulting encoder features may no longer remain linearly
separable. Let $\gamma_{\mathrm{enc}}$ denote the SNR at the encoder
output, and assume that it increases monotonically with the input SNR
$\gamma$. We model the noisy encoder output as
$\phi_q=\phi(r_q)+w_q$, where $\phi(r_q)$ is the noise-free feature
vector and $w_q$ is zero-mean Gaussian noise satisfying
\[
\operatorname{Cov}(w_q)
\preceq
\rho\gamma_{\mathrm{enc}}^{-1}I.
\]
Let
\[
\tilde{w}_q
=
\theta_{\mathrm{svm},w}^{*T}w_q.
\]
The signed margin of the noisy feature under the clean-data separating
hyperplane is
\begin{align}
y_q\left(
\theta_{\mathrm{svm},w}^{*T}\phi_q
+
\theta_{\mathrm{svm},\mathrm{bias}}^*
\right)
&=
y_q\left(
\theta_{\mathrm{svm},w}^{*T}\phi(r_q)
+
\theta_{\mathrm{svm},\mathrm{bias}}^*
\right)
+
y_q\tilde{w}_q.
\end{align}
Since the clean signed margin is at least $\mu$, the noisy feature
remains correctly separated whenever
\begin{align}
y_q\tilde{w}_q>-\mu.
\end{align}
Because $\|\theta_{\mathrm{svm},w}^*\|_2=1$,
$\tilde{w}_q$ is Gaussian with zero mean and variance bounded by
\[
\operatorname{Var}(\tilde{w}_q)
=
\theta_{\mathrm{svm},w}^{*T}
\operatorname{Cov}(w_q)
\theta_{\mathrm{svm},w}^*
\leq
\rho\gamma_{\mathrm{enc}}^{-1}.
\]
Assuming that the noise vectors are independent across samples, the
probability that all noisy features remain correctly separated by the
clean-data hyperplane satisfies
\begin{align}
&\Pr\left(
y_q\left(
\theta_{\mathrm{svm},w}^{*T}\phi_q
+
\theta_{\mathrm{svm},\mathrm{bias}}^*
\right)>0,
\ \forall q
\right)
\nonumber\\
&\geq
\left[
1-
Q\left(
\frac{\mu}
{\sqrt{\rho\gamma_{\mathrm{enc}}^{-1}}}
\right)
\right]^{\bar Q}.
\end{align}

This leads to the following guarantee.

\begin{theorem}
Let
$\mathcal{D}=\{(\phi(r_q),y_q)\}_{q=1}^{\bar{Q}}$
be a clean dataset that is $(\mu,\rho)$-separable under the causal CNN
encoder. Assume that the feature-noise vectors are independent across
samples. Then, for any $\epsilon>0$, there exists a threshold
$\delta(\epsilon,\bar{Q})>0$ such that, if
$\gamma_{\mathrm{enc}}>\delta(\epsilon,\bar{Q})$, all samples in the
noisy dataset
$\mathcal{D}'=\{(\phi_q,y_q)\}_{q=1}^{\bar{Q}}$
are correctly separated by the clean-data hyperplane with probability
at least $1-\epsilon$.
\end{theorem}

\noindent\textbf{Proof.}
From the preceding bound, the probability that all noisy samples remain
correctly separated by the clean-data hyperplane satisfies
\[
\Pr\left(
y_q\left(
\theta_{\mathrm{svm},w}^{*T}\phi_q
+
\theta_{\mathrm{svm},\mathrm{bias}}^*
\right)>0,
\ \forall q
\right)
\geq
\left[
1-
Q\left(
\frac{\mu}
{\sqrt{\rho\gamma_{\mathrm{enc}}^{-1}}}
\right)
\right]^{\bar{Q}}.
\]
As $\gamma_{\mathrm{enc}}\to\infty$, the argument of the $Q$-function
tends to infinity, and hence the lower bound tends to $1$. Therefore,
for any fixed $\epsilon>0$, there exists a finite threshold
$\delta(\epsilon,\bar{Q})$ such that the probability is at least
$1-\epsilon$ whenever
$\gamma_{\mathrm{enc}}>\delta(\epsilon,\bar{Q})$.
\hfill $\blacksquare$

\section{Experimental Results}

\subsection{Experimental settings}

We evaluate \textbf{FedSSL-AMC} against seven supervised learning baselines and four self-supervised learning baselines. The supervised baselines are:
\begin{itemize}
\item \textbf{FedeAMC-CNN}~\cite{wang2021federated} addresses class imbalance using class-weighted cross-entropy. It employs the CNN architecture in Table~\ref{tab:baseline_network_architecture}, which is also used by all supervised baselines for a consistent comparison.

\item \textbf{PMSGD-CNN}~\cite{rahman2024improved} selects a client-specific learning rate through a grid search to mitigate data heterogeneity. Each client evaluates learning rates in $\{0.01,0.001,0.0001\}$ using $20\%$ of its training samples and selects the value yielding the lowest training loss.

\item \textbf{FedAvg-CNN} denotes standard federated averaging.

\item \textbf{FedProx-CNN}~\cite{li2020federated} adds the proximal term
$\frac{\mu_{\text{prox}}}{2}
\|\theta_{\text{local}}-\theta_{\text{global}}\|_2^2$
to limit local-model drift under non-IID data. We set
$\mu_{\text{prox}}=0.01$.

\item \textbf{FedDyn-CNN}~\cite{acar2021federated} uses a dynamic regularizer to reduce mismatch between local and global objectives. We set the regularization strength to $0.01$.

\item \textbf{FedAH-CNN}~\cite{10924883} maintains both a global and a locally adapted copy of the output layer. After each local round, the two are combined through element-wise aggregation with trainable weights.

\item \textbf{FedRep-CNN}~\cite{collins2021exploiting} alternates between training a shared feature extractor, whose parameters are globally aggregated, and a personalized output layer that remains local. After representation training, the feature extractor is frozen and the local output layer is trained.
\end{itemize}

For the self-supervised learning baselines, we include:
\begin{itemize}
\item \textbf{SimCSE-CNN+SVM}, inspired by~\cite{gao-etal-2021-simcse}, forms positive pairs by passing the same input through dropout. Apart from this positive-pair construction, it follows the same training pipeline as \textbf{FedSSL-AMC}.

\item \textbf{SemiAMC}~\cite{liu2021self} adapts a centralized semi-supervised contrastive AMC method to the federated setting. It uses I/Q rotation for data augmentation and combines convolutional layers with a two-layer LSTM to capture local and long-range temporal structure. As in \textbf{FedSSL-AMC}, the encoder is frozen after representation learning.

\item \textbf{DAE}~\cite{ke2021real} adapts an LSTM-based denoising autoencoder for centralized AMC to the federated setting. Representation learning minimizes reconstruction loss, while the subsequent labeled-data phase minimizes a weighted combination of reconstruction and cross-entropy losses. To remain faithful to the original method, the encoder is not frozen during this second phase.

\item \textbf{CSSL-AMC}~\cite{10857965} adapts a centralized self-supervised contrastive AMC method to distributed settings. It augments high SNR data with additive Gaussian noise and uses ResNet-18. We report the original protocol, \textbf{CSSL-AMC (Enc. not frozen)}, which performs end-to-end fine-tuning after representation learning, and \textbf{CSSL-AMC (Enc. frozen)}, which freezes the encoder while training the classifier on labeled data. The frozen encoder variant enables a protocol-consistent comparison with \textbf{SemiAMC} and \textbf{FedSSL-AMC}, which evaluate pretrained representations with limited downstream adaptation.
\end{itemize}

Of the eleven baselines above, \textbf{FedeAMC-CNN}, \textbf{PMSGD-CNN}, \textbf{CSSL-AMC}, \textbf{SemiAMC}, and \textbf{DAE} were originally proposed for automatic modulation classification, and only \textbf{FedeAMC-CNN} and \textbf{PMSGD-CNN} were designed for federated learning. The remaining methods are our adaptations for the distributed AMC setting.

Across all experiments, \textbf{FedSSL-AMC} uses a 10-layer causal CNN encoder with kernel size 3 and a 320-dimensional output representation. During contrastive training, each anchor is paired with one positive and 10 negative samples. For \textbf{FedSSL-AMC} and all self-supervised baselines, representation learning runs for $T_{\text{rounds}}=10$ communication rounds. In each round, clients perform 15,000 local training steps with batch size 50 using Adam~\cite{kingma2014adam} and a learning rate of 0.001.

\vspace{0.05in}

\noindent
\textbf{Datasets.} We evaluate the framework on three datasets.
\begin{itemize}
\item \textbf{Custom synthetic dataset:} Following the signal model in~\cite{wang2021federated}, we generate a synthetic dataset with received baseband signal
\begin{align}
r[n] = A e^{j(\Delta\theta + 2\pi \Delta f \frac{n}{N})} s[n] + w[n],
\end{align}
where $s[n]$ is the transmitted symbol at time index $n$, $A$ is the channel gain, $\Delta\theta$ is the carrier phase offset, $\Delta f$ is the normalized carrier-frequency offset, and $N$ is the number of I/Q symbols per sequence. The additive noise $w[n]$ is circularly symmetric complex Gaussian. For each sequence, we draw $A_0 \sim \operatorname{Rayleigh}(1)$ and set $A=\min\{A_0,1\}$, while $\Delta\theta \sim \mathcal{U}(0,\pi/16)$ and $\Delta f=0.01$. The signal-to-noise ratio is
\begin{align}
\text{SNR} = 10 \log_{10} \left(
\frac{\sum_{n=0}^{N-1} \left| A e^{j(\Delta\theta + 2\pi \Delta f \frac{n}{N})} s[n] \right|^2}
{\sum_{n=0}^{N-1} |w[n]|^2}
\right),
\end{align}
with SNR sampled uniformly from $\mathcal{U}(-10,10)$ during both training and evaluation.

\begin{table}[htbp]
    \centering
     \caption{Network architecture used by the supervised learning baselines.}
    \label{tab:baseline_network_architecture}
    \begin{tabular}{c c l}
        \toprule
        \textbf{No.} & \textbf{Type} & \textbf{Structure} \\
        \midrule
        -- & -- & Input (I/Q samples and labels) \\
        1 & Conv & Conv1D (128, 16) + BN + ReLU + Dropout (0.1) \\
        2 & Conv & Conv1D (64, 8) + BN + ReLU + Dropout (0.1) \\
        3 & FC   & Dense (256) + BN + ReLU + Dropout (0.5) \\
        4 & FC   & Dense (128) + BN + ReLU + Dropout (0.5) \\
        5 & FC   & Dense (4) + Softmax \\
        \bottomrule
    \end{tabular}
   
\end{table}

\begin{figure}[!htbp]
    \centering
    \includegraphics[width=\linewidth, keepaspectratio]{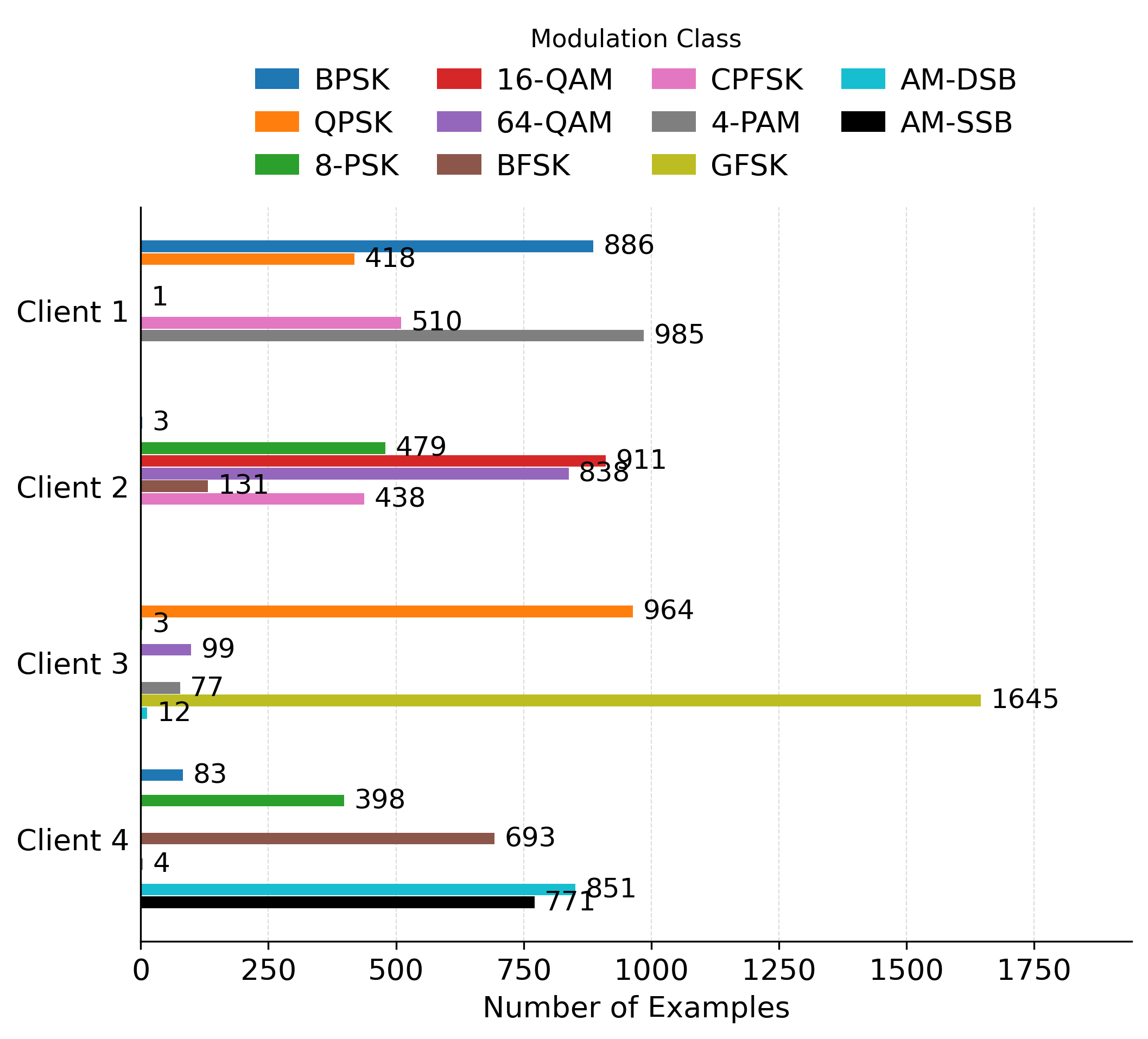}
    \caption{Client-wise label distributions in the MIGOU dataset under Dirichlet partitioning with $\bar{\alpha}=0.1$, illustrating highly skewed non-IID partitions across clients.}
    \label{fig:dist_migou_dirichlet}
\end{figure}

\begin{table}[!htbp]
    \centering
    \small
    \setlength{\tabcolsep}{3.5pt}
    \caption{Client-averaged classification accuracy across the three datasets. Results for the synthetic and DeepSig datasets are also averaged over SNR.}
    \label{tab:acc_all_datasets}
    \begin{tabular}{lccc}
        \toprule
        \textbf{Method}
        & \textbf{Synthetic}
        & \textbf{MIGOU}
        & \textbf{DeepSig} \\
        \midrule

        \multicolumn{4}{l}{\textit{Supervised / non-SSL FL baselines}} \\

        \textbf{FedAvg-CNN}
        & 41.61 & 37.19 & 25.12 \\

        \textbf{FedeAMC}
        & 27.34 & 33.20 & 4.02 \\

        \textbf{FedProx-CNN}
        & 40.82 & 37.67 & 16.82 \\

        \textbf{FedDyn-CNN}
        & 36.63 & 67.17 & 18.37 \\

        \textbf{PMSGD-CNN}
        & 43.92 & 32.88 & 20.24 \\

        \textbf{FedAH-CNN}
        & 61.05 & 78.41 & 28.32 \\

        \textbf{FedRep-CNN}
        & 61.31 & 79.52 & 24.11 \\

        \midrule
        \multicolumn{4}{l}{\textit{SSL / representation-learning methods}} \\

        \textbf{SimCSE-CNN+SVM}
        & 57.30 & 76.57 & 23.09 \\

        \textbf{CSSL-AMC (enc. not frozen)}
        & 49.69 & 74.08 & 32.66 \\

        \textbf{CSSL-AMC (enc. frozen)}
        & 30.83 & 36.37 & 15.39 \\

        \textbf{SemiAMC}
        & 36.13 & 63.61 & 33.32 \\

        \textbf{DAE}
        & 37.63 & 27.71 & 12.79 \\

        \textbf{FedSSL-AMC (Ours)}
        & \textbf{66.69} & \textbf{84.55} & \textbf{36.08} \\

        \bottomrule
    \end{tabular}
    
\end{table}


\begin{figure*}[!t]
    \centering
    \begin{subfigure}[t]{0.32\textwidth}
        \centering
        \includegraphics[width=\linewidth]{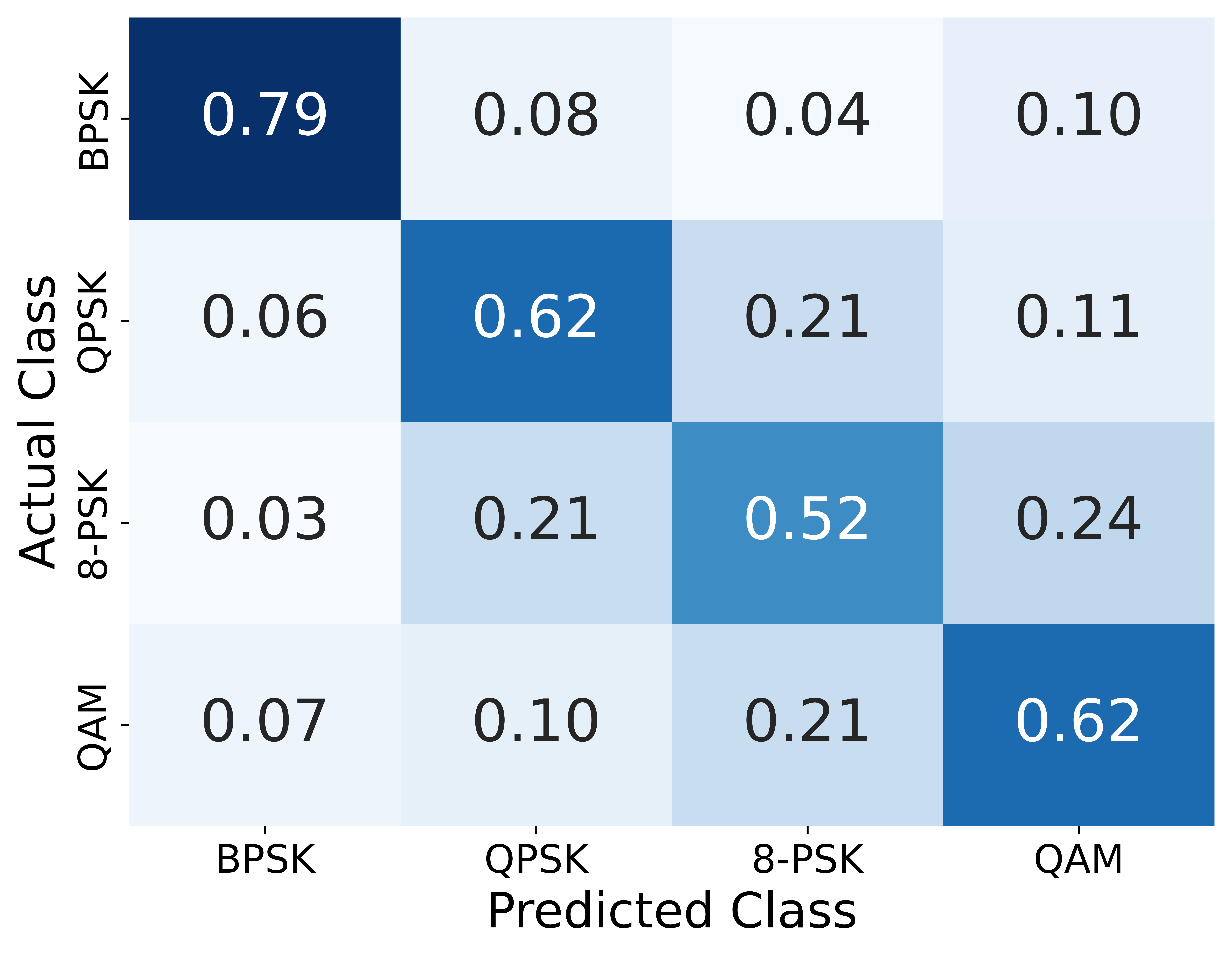}
        \caption{FedSSL-AMC}
        \label{fig:confssl}
    \end{subfigure}
    \hfill
    \begin{subfigure}[t]{0.32\textwidth}
        \centering
        \includegraphics[width=\linewidth]{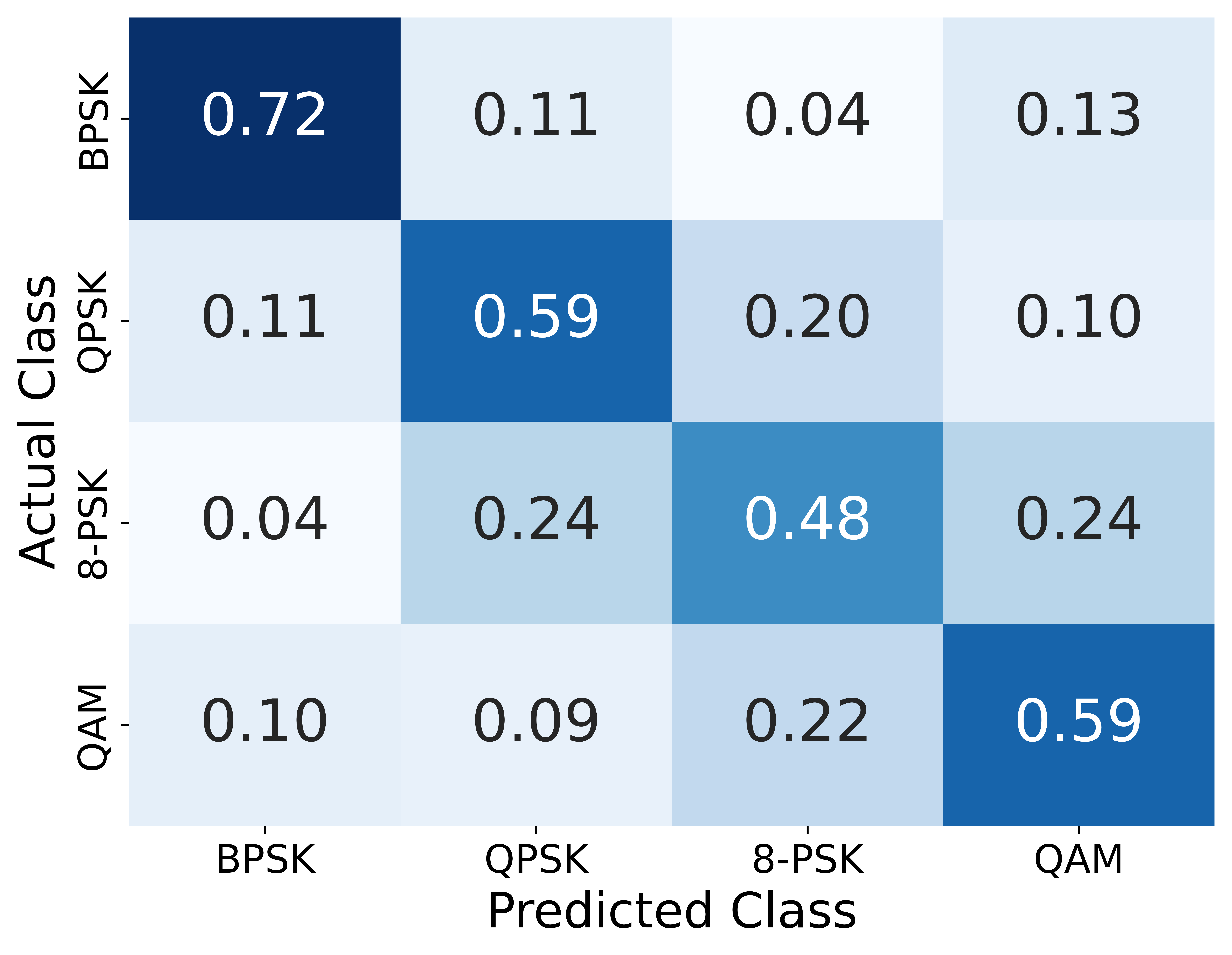}
        \caption{SimCSE-CNN+SVM}
        \label{fig:confsimcse}
    \end{subfigure}
    \hfill
    \begin{subfigure}[t]{0.32\textwidth}
        \centering
        \includegraphics[width=\linewidth]{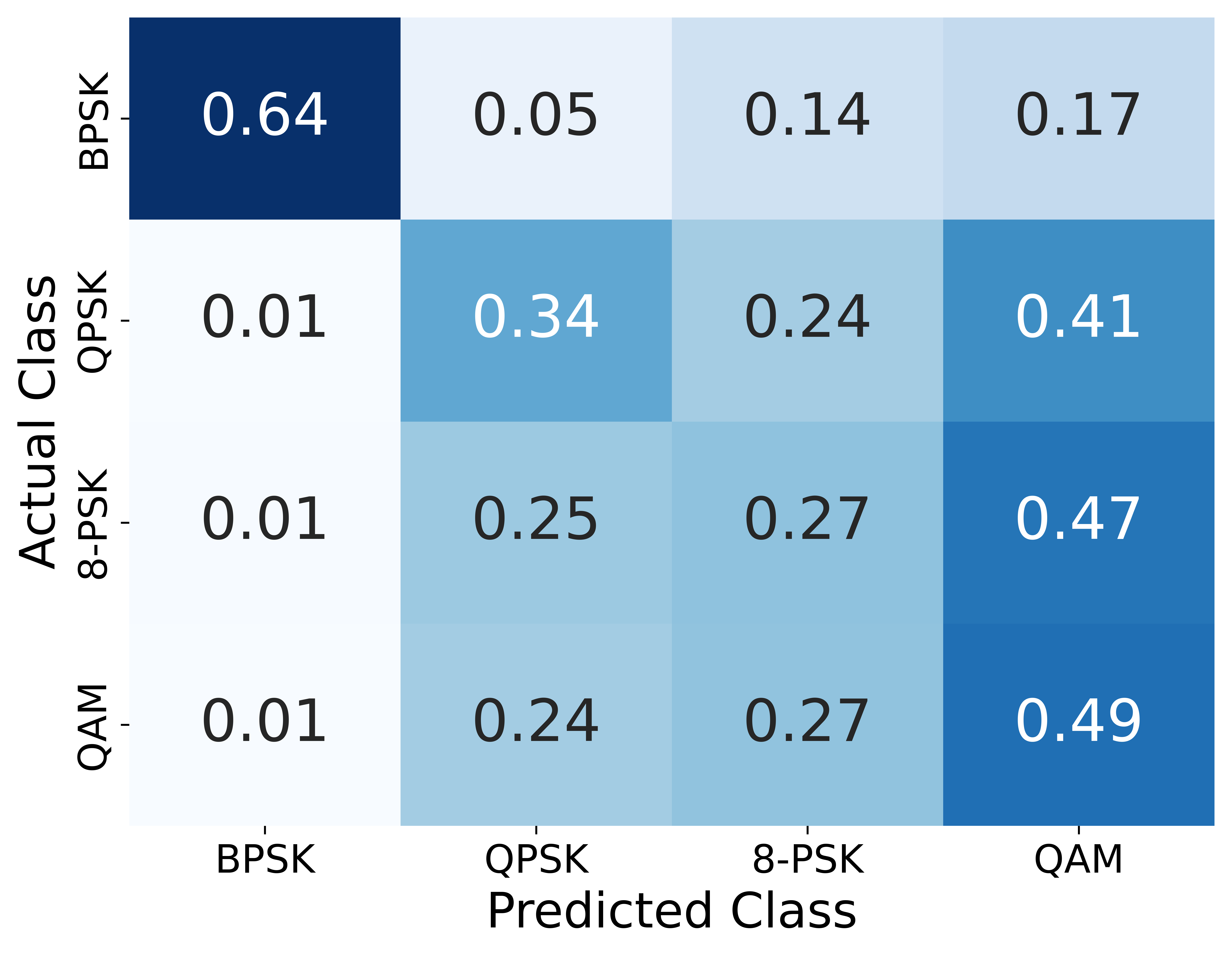}
        \caption{FedDyn-CNN}
        \label{fig:feddyn}
    \end{subfigure}

    \caption{Client- and SNR-averaged confusion matrices for FedSSL-AMC,
    SimCSE-CNN+SVM, and FedDyn-CNN with 14{,}000 labeled samples per client.}
    \label{fig:confmat}
\end{figure*}

\begin{table}[!htbp]
\centering
\small
\setlength{\tabcolsep}{2.8pt}
\caption{Client-averaged test accuracy on the synthetic dataset as the number of labeled training samples per client varies.}
\label{tab_num_labeled}
\begin{tabular}{lccccc}
\toprule
\textbf{Method}
& \textbf{2.8k}
& \textbf{4.2k}
& \textbf{7.0k}
& \textbf{9.8k}
& \textbf{14.0k} \\
\midrule

\multicolumn{6}{l}{\textit{Supervised / non-SSL FL baselines}} \\
\textbf{FedAvg-CNN}   & 41.61 & 41.52 & 41.25 & 41.98 & 42.62 \\
\textbf{FedeAMC}      & 27.34 & 27.34 & 41.42 & 42.88 & 43.69 \\
\textbf{FedProx-CNN}  & 40.82 & 43.15 & 42.91 & 43.29 & 44.72 \\
\textbf{FedDyn-CNN}   & 36.63 & 37.83 & 37.38 & 40.09 & 44.28 \\
\textbf{PMSGD-CNN}    & 43.92 & 42.64 & 45.20 & 44.92 & 47.61 \\
\textbf{FedAH-CNN}    & 50.15 & 54.16 & 60.87 & 62.38 & 63.92 \\
\textbf{FedRep-CNN}   & 61.31 & 60.08 & 60.87 & 62.38 & 63.53 \\

\midrule
\multicolumn{6}{l}{\textit{SSL / representation-learning methods}} \\
\textbf{SimCSE-CNN+SVM}          & 57.30 & 60.26 & 59.73 & 59.56 & 59.64 \\
\textbf{CSSL-AMC (enc. not frozen)} & 49.69 & 49.17 & 49.46 & 51.51 & 53.72 \\
\textbf{CSSL-AMC (enc. frozen)}  & 30.83 & 39.02 & 38.23 & 38.76 & 42.10 \\
\textbf{SemiAMC}                 & 36.13 & 35.67 & 35.52 & 35.83 & 35.43 \\
\textbf{DAE}                     & 37.63 & 45.00 & 54.59 & 47.69 & 49.30 \\
\textbf{FedSSL-AMC (Ours)}       & \textbf{66.69} & \textbf{68.21} & \textbf{67.86} & \textbf{68.13} & \textbf{68.13} \\

\bottomrule
\end{tabular}

\end{table}

Each I/Q sequence contains $N=100$ complex samples, and we consider four modulation types: BPSK, QPSK, 8-PSK, and 16-QAM. The federated setup includes four clients, each with $140{,}000$ unlabeled and $2{,}800$ labeled training samples. To induce label imbalance and non-IID client distributions, the modulation classes are assigned unevenly across clients. In the order [BPSK, QPSK, 8-PSK, 16-QAM], the class counts are:
\begin{enumerate}
\item \textbf{Client 1:} $[60{,}000, 60{,}000, 10{,}000, 10{,}000]$ unlabeled and $[1{,}200, 1{,}200, 200, 200]$ labeled samples.
\item \textbf{Client 2:} $[10{,}000, 60{,}000, 60{,}000, 10{,}000]$ unlabeled and $[200, 1{,}200, 1{,}200, 200]$ labeled samples.
\item \textbf{Client 3:} $[10{,}000, 10{,}000, 60{,}000, 60{,}000]$ unlabeled and $[200, 200, 1{,}200, 1{,}200]$ labeled samples.
\item \textbf{Client 4:} $[60{,}000, 10{,}000, 10{,}000, 60{,}000]$ unlabeled and $[1{,}200, 200, 200, 1{,}200]$ labeled samples.
\end{enumerate}

Following the original \textbf{FedeAMC} training protocol, all supervised baselines use $1{,}000$ communication rounds with one local epoch per client per round, a batch size of $64$, and a learning rate of $0.001$. We also adopt the label distribution from~\cite{wang2021federated} to ensure a fair and faithful comparison with \textbf{FedeAMC}, the primary prior work on federated automatic modulation classification.

Each client's test set follows the class distribution of its labeled training data and contains one-tenth as many samples. Test SNRs are sampled independently from $\mathcal{U}(-10,10)$.

\item \textbf{MIGOU dataset:} The MIGOU dataset~\cite{utrilla2019migou} contains over-the-air I/Q measurements from 11 modulation classes, including AM, FM, QAM, and PSK signals, transmitted using a USRP B210 and recorded at distances of 1~m and 6~m. These distances correspond to average SNRs of 37~dB and 22~dB, respectively. Its larger number of modulation classes makes MIGOU useful for evaluating robustness under label imbalance and client heterogeneity. To generate non-IID client distributions, we partition each modulation class using Dirichlet sampling:
\[
p([x_1,x_2,x_3,x_4]) = \frac{1}{B(\alpha)} \prod_{i=1}^4 x_i^{\alpha_i - 1},
\]
where $x_i$ is the fraction of samples from a given class assigned to client $i$, and $B(\alpha)$ is the multivariate Beta function. We set
$\alpha=\bar{\alpha}[1,1,1,1]$, where smaller $\bar{\alpha}$ yields more uneven class distributions and larger values produce more uniform splits. Figure~\ref{fig:dist_migou_dirichlet} shows the client-wise distributions for $\bar{\alpha}=0.1$; distributions for other values are provided in Appendix Section~\ref{DirichletDistsSec}. In the main experiments, each client receives $140{,}000$ unlabeled and $2{,}800$ labeled samples. All methods are trained for 10 communication rounds with 100 local epochs per round.

\item \textbf{DeepSig 2018.01A dataset:} The DeepSig 2018.01A dataset~\cite{deepsig2018} contains synthetic and over-the-air recordings from 24 digital and analog modulation classes. SNR ranges from $-20$~dB to $30$~dB in 2~dB increments. For each modulation class and SNR level, the dataset provides $4{,}096$ examples, each consisting of $1{,}024$ complex I/Q samples. We consider four clients and partition each modulation class across them using the same Dirichlet procedure as for MIGOU. Each client receives $2{,}500$ samples per modulation class and SNR level, including $500$ labeled samples; the remainder are treated as unlabeled. As with MIGOU, supervised baselines are trained for 10 communication rounds with 100 local epochs per round.

\end{itemize}

\subsection{Results} \label{4B}
Table~\ref{tab:acc_all_datasets} reports classification accuracy across all three datasets. For MIGOU and DeepSig, we set the Dirichlet concentration parameter to $\bar{\alpha}=0.1$ to induce substantial label heterogeneity across clients. \textbf{FedSSL-AMC} achieves the highest accuracy on all three datasets, demonstrating strong performance under both client heterogeneity and limited label availability.


\begin{figure*}[!t]
    \centering

    \begin{subfigure}[t]{0.48\textwidth}
        \centering
        \includegraphics[width=\linewidth]{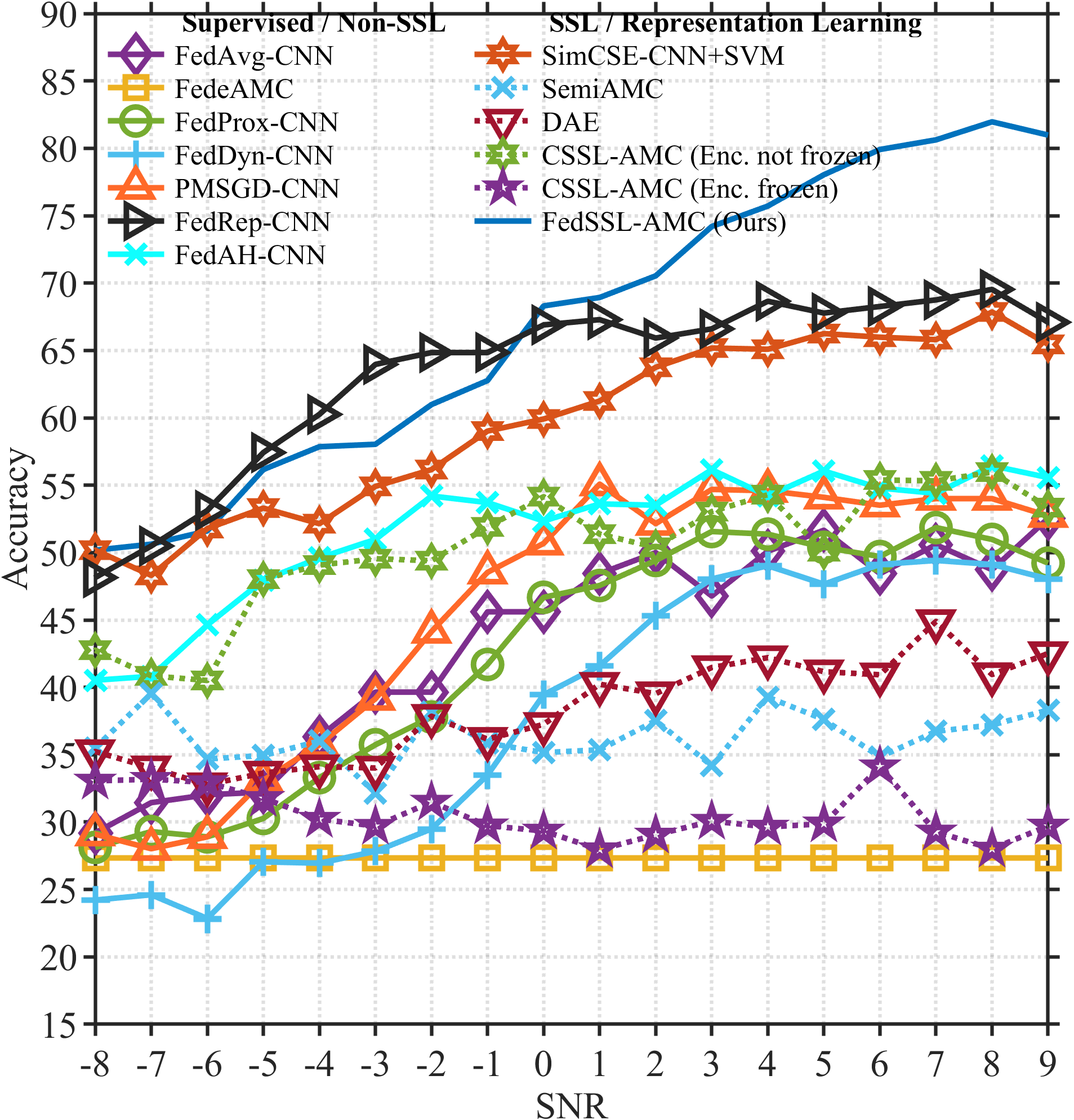}
        \caption{2{,}800 labeled samples per client}
        \label{fig:2a}
    \end{subfigure}
    \hfill
    \begin{subfigure}[t]{0.48\textwidth}
        \centering
        \includegraphics[width=\linewidth]{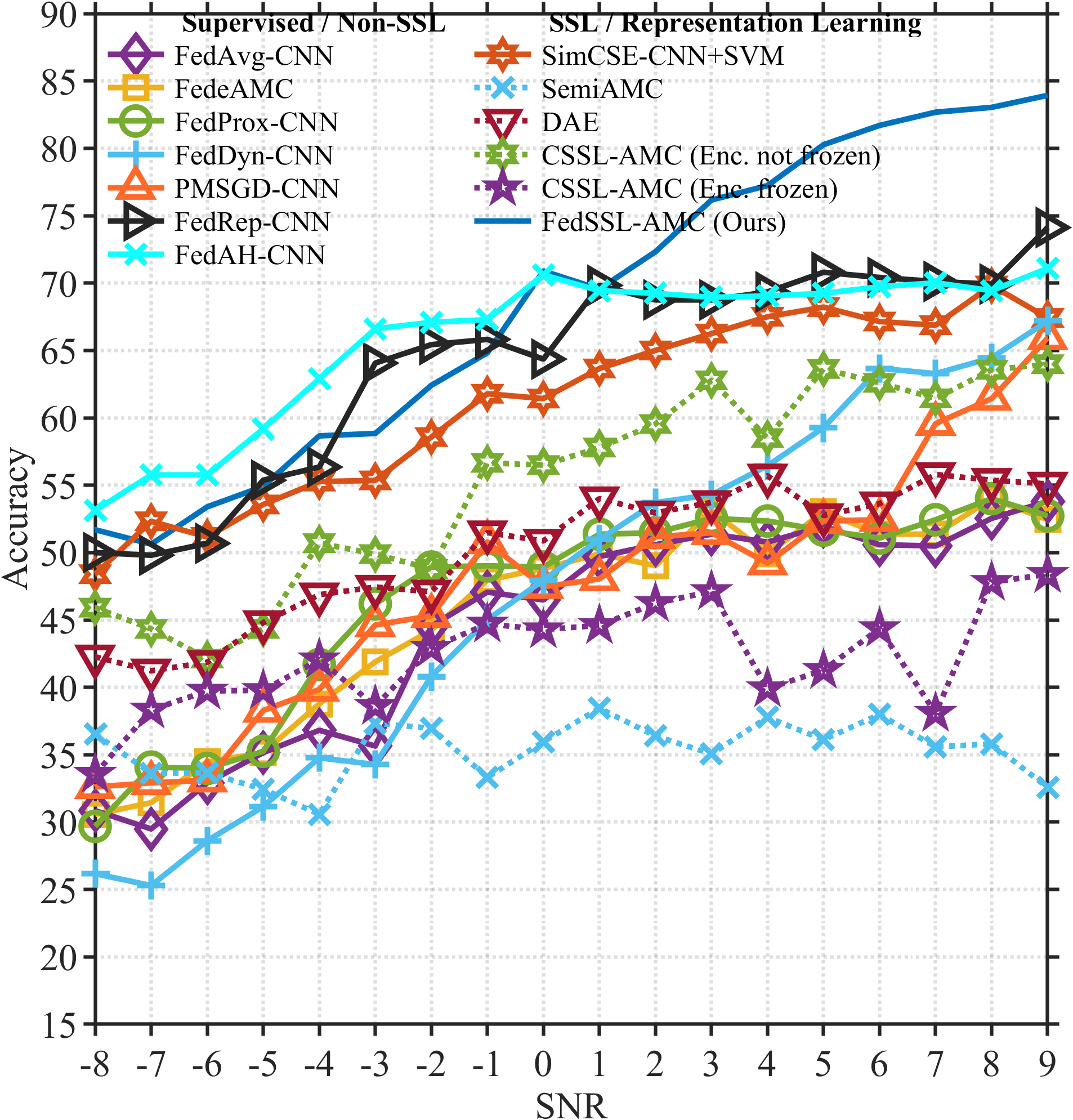}
        \caption{14{,}000 labeled samples per client}
        \label{fig:2b}
    \end{subfigure}

    \caption{Test accuracy versus SNR on the custom synthetic dataset for two labeled-sample budgets.}
    \label{fig:2}

\end{figure*}

Next, we conduct ablation studies, primarily on the custom synthetic dataset, to assess the sensitivity of \textbf{FedSSL-AMC} to SNR, labeled-sample size, carrier-frequency offset, model heterogeneity, and label heterogeneity.

\begin{itemize}

\item \textbf{Number of labeled samples.} We assess label efficiency on the custom synthetic dataset by varying the number of labeled samples per client and reporting client-averaged accuracy in Table~\ref{tab_num_labeled}. \textbf{FedeAMC} is particularly sensitive to label scarcity, with accuracy dropping sharply at smaller labeled-set sizes. Fig.~\ref{fig:confmat} shows confusion matrices for 14{,}000 labeled samples per client. \textbf{FedSSL-AMC} exhibits a stronger diagonal structure and fewer systematic off-diagonal errors than \textbf{SimCSE-CNN+SVM} and \textbf{FedDyn-CNN}, indicating better discrimination among modulation classes.

\begin{table}[!htbp]
\centering
\small
\setlength{\tabcolsep}{3.5pt}
\caption{Client-wise classification accuracy under heterogeneous SNR conditions on the custom synthetic dataset. CSSL-AMC is omitted because its augmentation procedure requires high-SNR data.}
\label{tab:results1b}
\begin{tabular}{lcccc}
\toprule
\textbf{Method} & \textbf{Client 1} & \textbf{Client 2} & \textbf{Client 3} & \textbf{Client 4} \\
\midrule

\multicolumn{5}{l}{\textit{Supervised / non-SSL FL baselines}} \\
\textbf{FedAvg-CNN}   & 31.64 & 27.34 & 58.59 & 69.14 \\
\textbf{FedeAMC}      &  7.81 &  7.81 & 46.88 & 46.88 \\
\textbf{FedProx-CNN}  & 31.64 & 33.20 & 70.31 & 64.84 \\
\textbf{FedDyn-CNN}   & 33.20 & 35.15 & 63.37 & 62.10 \\
\textbf{PMSGD-CNN}    & 34.69 & 34.77 & 57.11 & 58.80 \\
\textbf{FedAH-CNN}    & 46.32 & 46.32 & 79.84 & 90.54 \\
\textbf{FedRep-CNN}   & \textbf{46.80} & 27.19 & 62.73 & 81.72 \\

\midrule
\multicolumn{5}{l}{\textit{SSL / representation-learning methods}} \\
\textbf{SimCSE-CNN+SVM}   & 42.32 & 44.79 & 81.11 & 85.14 \\
\textbf{CSSL-AMC (enc. not frozen)} & -- & -- & -- & -- \\
\textbf{CSSL-AMC (enc. frozen)}   & -- & -- & -- & -- \\
\textbf{SemiAMC}          & 22.46 & 29.36 & 47.98 & 53.32 \\
\textbf{DAE}              & 25.33 & 40.42 & 46.09 & 58.46 \\
\textbf{FedSSL-AMC (Ours)} & 40.36 & \textbf{47.93} & \textbf{84.36} & \textbf{91.43} \\

\bottomrule
\end{tabular}
\end{table}

\begin{figure*}[!htbp]
    \centering

    \begin{subfigure}[t]{0.48\textwidth}
        \centering
        \includegraphics[width=\linewidth]{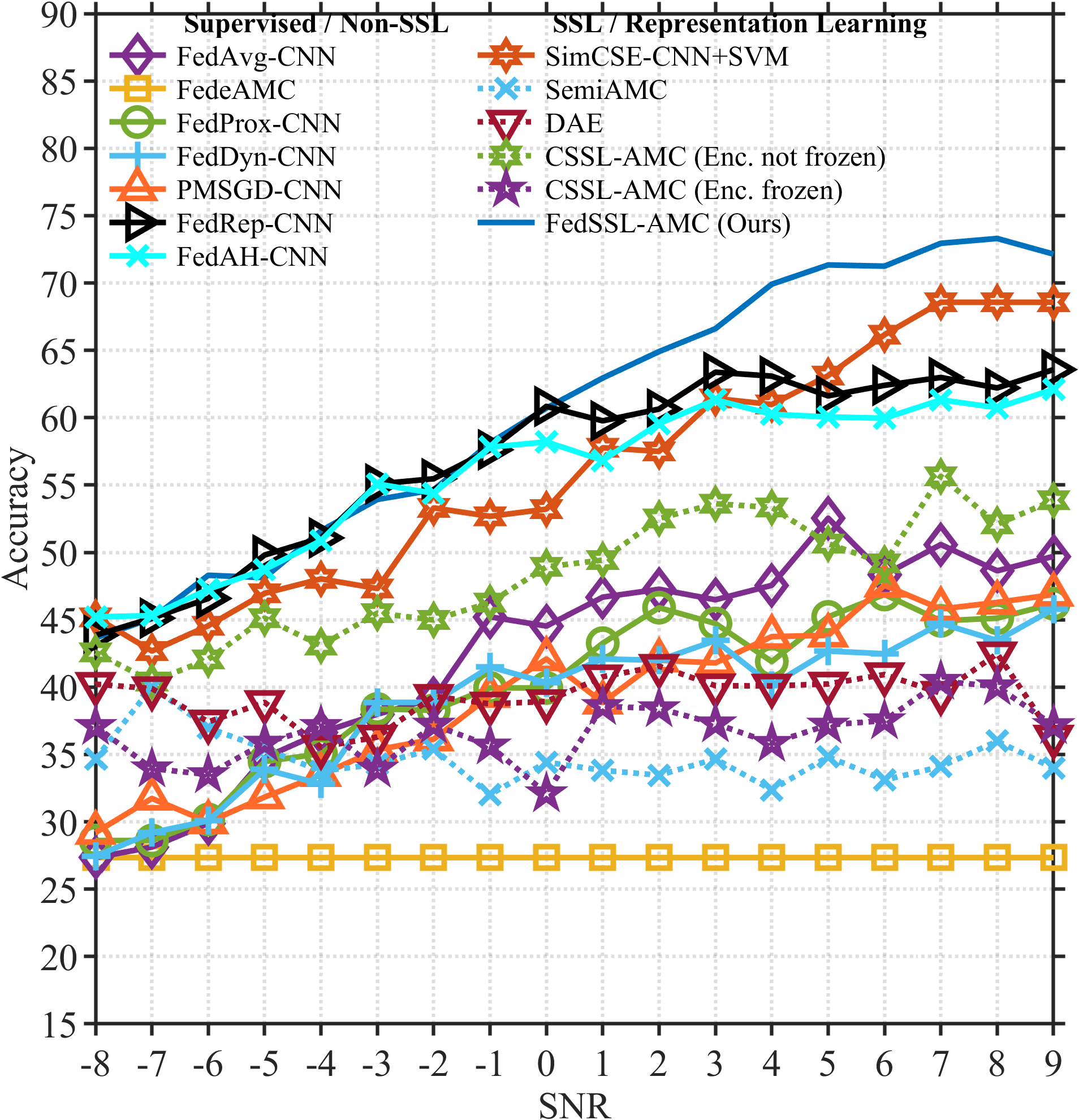}
        \caption{2,800 labeled samples per client}
        \label{fig:4a}
    \end{subfigure}
    \hfill
    \begin{subfigure}[t]{0.48\textwidth}
        \centering
        \includegraphics[width=\linewidth]{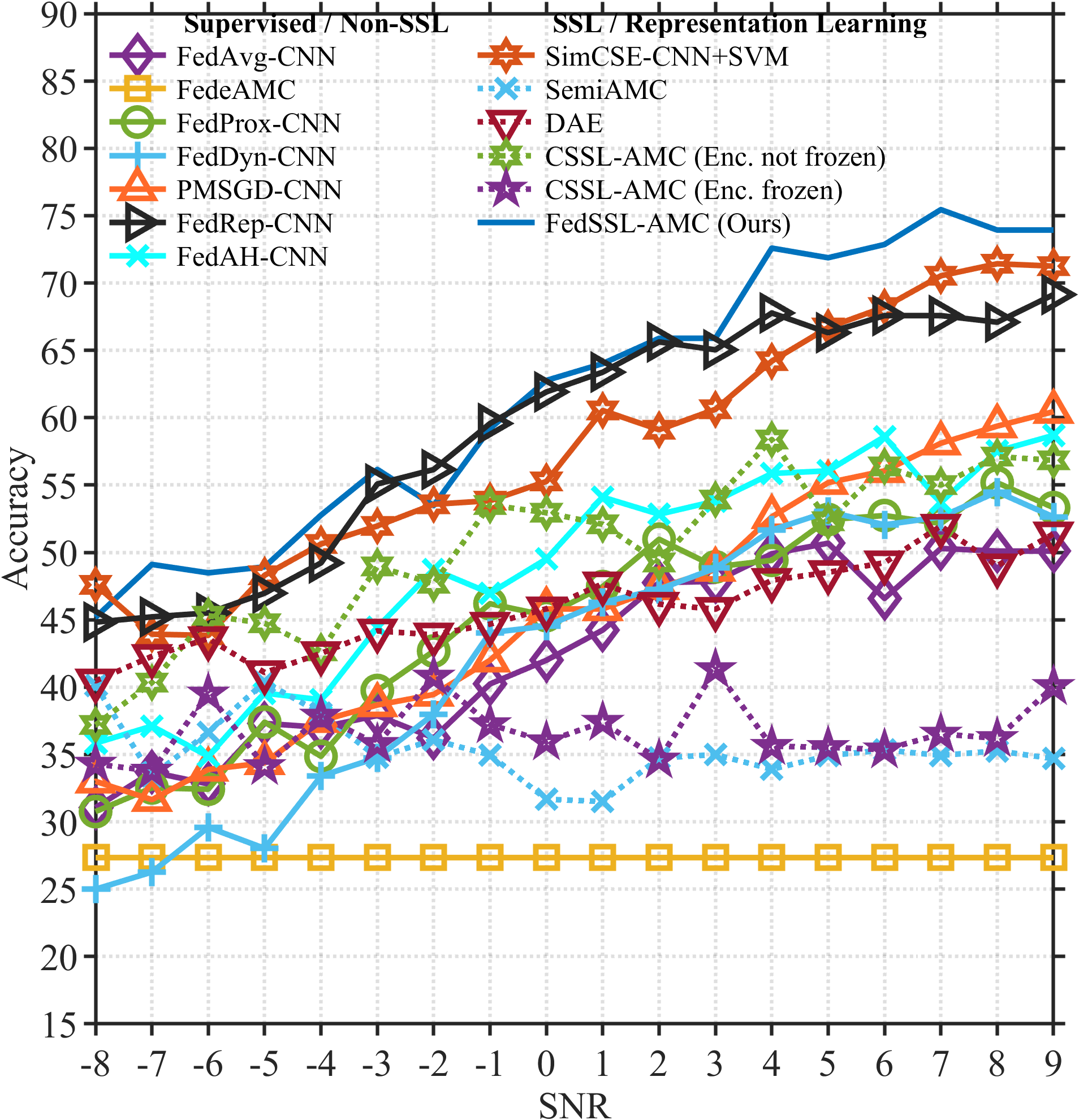}
        \caption{14,000 labeled samples per client}
        \label{fig:4b}
    \end{subfigure}

    \caption{Test accuracy versus SNR on the custom synthetic dataset under combined label-distribution and CFO heterogeneity.}
    \label{fig:CFO_het}

\end{figure*}

\item \textbf{Variable SNR conditions across clients.} We next introduce client-level SNR heterogeneity on the custom synthetic dataset. Each client receives 14{,}000 labeled training sequences, with SNRs sampled from different ranges: $\mathcal{U}(-10,-5)$ for Client~1, $\mathcal{U}(-5,0)$ for Client~2, $\mathcal{U}(0,5)$ for Client~3, and $\mathcal{U}(5,10)$ for Client~4. We also evaluate test accuracy as the SNR varies from $-10$~dB to $9$~dB. As shown in Fig.~\ref{fig:2}, accuracy increases with SNR. Table~\ref{tab:results1b} further shows that the causal-CNN-based self-supervised methods outperform the supervised baselines under heterogeneous SNR conditions.

\item \textbf{Variation in $\Delta f$.} We next examine mobility-induced heterogeneity by varying the carrier-frequency offset (CFO) $\Delta f$. Each client is assigned a different mobility regime, as summarized in Table~\ref{tab:cfo_heterogeneitydist}; the corresponding mapping between mobility and $\Delta f$ is detailed in the appendix. Figure~\ref{fig:CFO_het} reports client-averaged test accuracy for $2{,}800$ and $14{,}000$ labeled samples per client, while Table~\ref{tab:mean_acc_labeled_CFO} presents the corresponding SNR-averaged accuracies. \textbf{FedSSL-AMC} generally outperforms the supervised and self-supervised baselines, although \textbf{FedAH-CNN} and \textbf{FedRep-CNN} remain competitive at lower SNRs.

\begin{table}[!htbp]
\caption{Client-specific mixture weights over four carrier-frequency-offset regimes. Each column corresponds to the indicated distribution of $\Delta f$.}
\label{tab:cfo_heterogeneitydist}
\centering
\small
\setlength{\tabcolsep}{4pt}
\begin{tabular}{ccccc}
\toprule
\textbf{Client}
& $\mathcal{U}[0,0.01]$
& $\mathcal{U}[0.01,0.1]$
& $\mathcal{U}[0.1,1]$
& $\mathcal{U}[1,20]$ \\
\midrule
1 & 0.4 & 0.4 & 0.1 & 0.1 \\
2 & 0.4 & 0.1 & 0.4 & 0.1 \\
3 & 0.1 & 0.4 & 0.4 & 0.1 \\
4 & 0.1 & 0.1 & 0.4 & 0.4 \\
\bottomrule
\end{tabular}
\end{table}

\begin{table}[!htbp]
\centering
\small
\setlength{\tabcolsep}{5pt}
\caption{SNR-averaged classification accuracy (\%) on the custom synthetic dataset under combined label-distribution and carrier-frequency-offset heterogeneity, for $2{,}800$ and $14{,}000$ labeled samples per client.}
\label{tab:mean_acc_labeled_CFO}
\begin{tabular}{lcc}
\toprule
\textbf{Method} & \textbf{2.8k labels} & \textbf{14.0k labels} \\
\midrule

\multicolumn{3}{l}{\textit{Supervised / non-SSL FL baselines}} \\
\textbf{FedAvg-CNN}   & 42.30 & 42.54 \\
\textbf{FedeAMC}      & 27.34 & 27.34 \\
\textbf{FedProx-CNN}  & 39.87 & 44.69 \\
\textbf{FedDyn-CNN}   & 37.66 & 40.47 \\
\textbf{PMSGD-CNN}    & 38.07 & 43.85 \\
\textbf{FedAH-CNN}    & 54.57 & 47.09 \\
\textbf{FedRep-CNN}   & 55.63 & 57.66 \\

\midrule
\multicolumn{3}{l}{\textit{SSL / representation-learning methods}} \\
\textbf{SimCSE-CNN+SVM}       & 54.73 & 56.52 \\
\textbf{CSSL-AMC (enc. not frozen)}  & 47.70 & 48.86 \\
\textbf{CSSL-AMC (enc. frozen)}    & 36.53 & 36.77 \\
\textbf{SemiAMC}              & 35.12 & 35.46 \\
\textbf{DAE}                  & 39.15 & 45.46 \\
\textbf{FedSSL-AMC (Ours)}    & \textbf{58.73} & \textbf{60.02} \\
\bottomrule
\end{tabular}
\end{table}

\item \textbf{Model heterogeneity.} We evaluate \textbf{FedSSL-AMC} under heterogeneous model precision on the custom synthetic dataset. Client~1 uses float32, Client~2 uses float16, and Clients~3 and~4 use int8 quantization during training. Figure~\ref{fig:Quant_het} reports test accuracy with $14{,}000$ labeled samples per client. Despite the differences in bit width, \textbf{FedSSL-AMC} outperforms all baselines.

\begin{figure}[!htbp]
    \centering    \includegraphics[width=0.95\columnwidth]{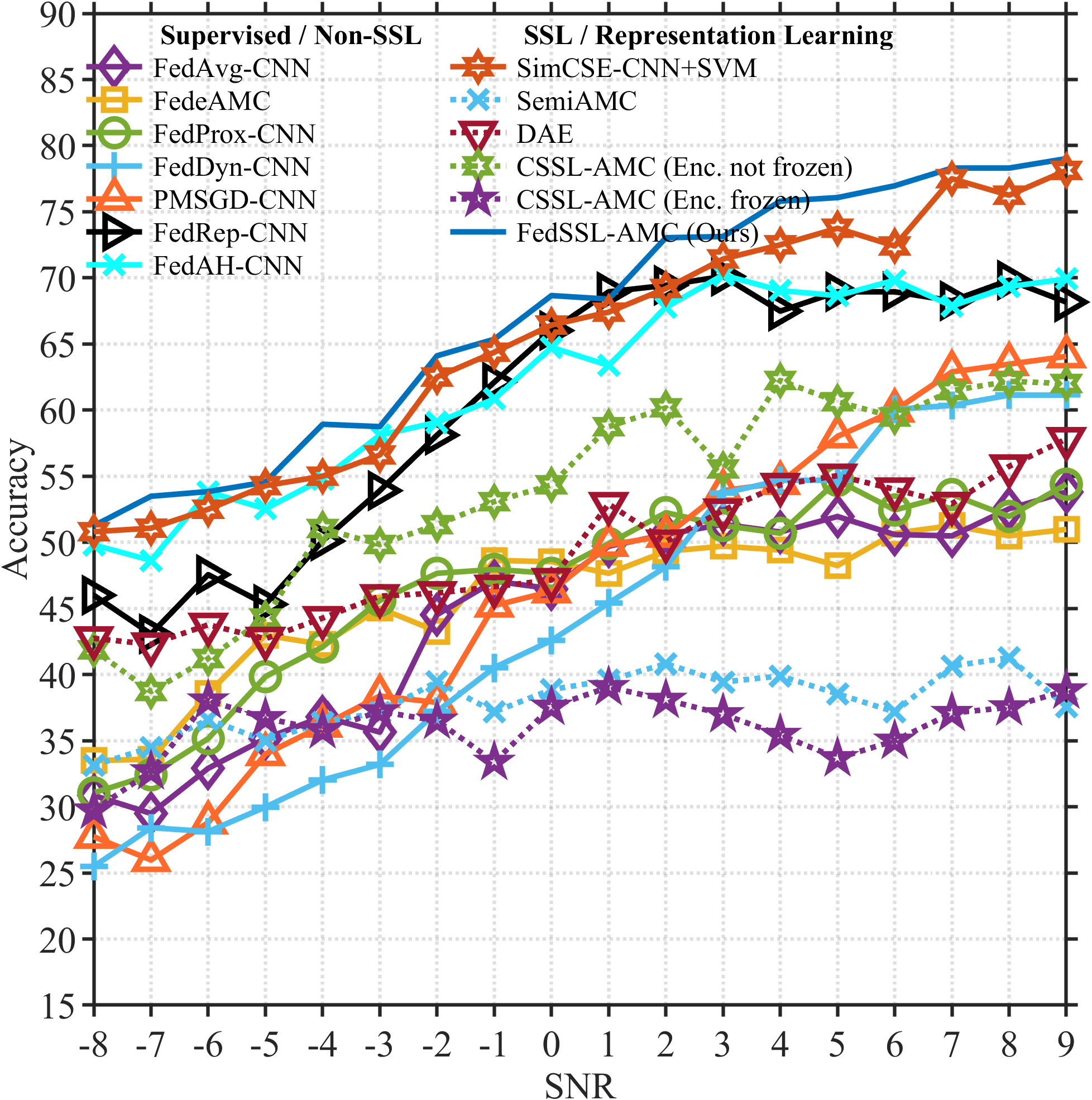}

    \caption{Test accuracy versus SNR on the custom synthetic dataset under combined label-distribution and model-precision heterogeneity.}
    \hfill
    \label{fig:Quant_het}
\end{figure}

\item \textbf{Degree of label heterogeneity $\bar{\alpha}$.} We vary the Dirichlet concentration parameter $\bar{\alpha}$ from $0.05$ to $1.0$ on the MIGOU dataset to assess sensitivity to label heterogeneity. For each value, we report the mean and standard deviation of client-averaged accuracy over 10 independent runs. We also consider a larger setting with 16 clients at $\bar{\alpha}=0.05$, organized into five clusters of sizes 3, 3, 3, 3, and 4. Because the data are distributed across more clients, each client receives 14{,}000 unlabeled samples. Results are reported in Table~\ref{tab:beta_results}. Under severe heterogeneity, \textbf{FedSSL-AMC} generally outperforms the baselines by a clear margin, except for \textbf{CSSL-AMC (Enc. not frozen)}, which fine-tunes the entire encoder during downstream supervised training. The large gap between the encoder-frozen and encoder-not-frozen CSSL-AMC variants indicates that much of its performance gain comes from downstream encoder adaptation rather than from the pretrained representation alone. In contrast, \textbf{FedSSL-AMC} remains competitive while keeping the encoder fixed and training only a lightweight classifier. Because MIGOU contains only two SNR levels, we evaluate SNR sensitivity on DeepSig for $\bar{\alpha}\in\{0.1,0.5\}$. Figure~\ref{fig:deepsig} shows the resulting four-client-averaged accuracy curves. Most baselines degrade under this setting, whereas \textbf{FedSSL-AMC} remains consistently strong and achieves slightly higher accuracy at the more heterogeneous setting, $\bar{\alpha}=0.1$.

\end{itemize}

\begin{table*}[!t]
\centering
\small
\setlength{\tabcolsep}{2.5pt}
\renewcommand{\arraystretch}{1.04}
\caption{Mean client-averaged classification accuracy (\%) with standard deviation (in parentheses) over 10 runs on the MIGOU dataset under varying levels of label heterogeneity, parameterized by the Dirichlet concentration $\bar{\alpha}$. Results for \textbf{CSSL-AMC} are reported with both encoder-frozen and encoder-not-frozen downstream training protocols. The first seven columns correspond to the 4-client setting, while the final column reports the 16-client, 5-cluster setting with $\bar{\alpha}=0.05$.}
\label{tab:beta_results}
\begin{tabular}{@{}p{0.22\textwidth}cccccccc@{}}
\toprule
\multirow{2}{*}{\textbf{Method}}
& \multicolumn{7}{c}{\textbf{4-client setting}}
& \multicolumn{1}{c}{\textbf{16 clients, 5 clusters}} \\
\cmidrule(lr){2-8}\cmidrule(l){9-9}
& $\bar{\alpha}=0.05$
& $0.1$
& $0.25$
& $0.375$
& $0.5$
& $0.625$
& $1.0$
& $\bar{\alpha}=0.05$ \\
\midrule

\multicolumn{9}{l}{\textit{Supervised / non-SSL FL baselines}} \\

\textbf{FedAvg-CNN}
& 30.9 (8.4)
& 37.2 (7.9)
& 50.8 (7.5)
& 53.0 (5.9)
& 53.4 (6.6)
& 58.2 (6.8)
& 62.8 (2.1)
& 31.2 (5.6) \\

\textbf{FedeAMC}
& 29.8 (10.0)
& 33.2 (10.2)
& 45.6 (6.2)
& 39.9 (17.4)
& 44.7 (13.6)
& 52.4 (15.7)
& 59.7 (3.3)
& 14.0 (12.8) \\

\textbf{FedProx-CNN}
& 29.0 (12.9)
& 37.7 (8.0)
& 46.7 (7.7)
& 46.2 (8.7)
& 52.9 (8.3)
& 57.9 (9.8)
& 62.5 (2.3)
& 37.4 (5.1) \\

\textbf{FedDyn-CNN}
& 66.5 (3.2)
& 67.2 (2.1)
& 68.5 (1.1)
& 68.2 (0.6)
& 66.5 (6.4)
& 66.3 (4.1)
& 52.5 (23.2)
& 18.6 (14.0) \\

\textbf{PMSGD-CNN}
& 23.3 (6.1)
& 32.9 (8.8)
& 40.2 (9.3)
& 44.0 (6.7)
& 48.3 (8.0)
& 48.9 (9.8)
& 56.4 (3.8)
& 27.4 (6.4) \\

\textbf{FedAH-CNN}
& 71.2 (13.5)
& 78.4 (7.6)
& 77.0 (7.5)
& 74.5 (4.8)
& 74.7 (5.1)
& 74.7 (4.3)
& 72.9 (2.0)
& 68.9 (8.4) \\

\textbf{FedRep-CNN}
& 77.3 (10.0)
& 79.5 (8.3)
& 73.1 (6.4)
& 75.2 (7.5)
& 74.5 (5.0)
& 75.5 (3.7)
& 72.7 (1.9)
& 75.6 (3.7) \\

\midrule
\multicolumn{9}{l}{\textit{SSL / representation-learning methods}} \\

\textbf{SimCSE-CNN+SVM}
& 81.4 (3.6)
& 76.6 (7.2)
& 71.7 (4.9)
& 68.5 (5.2)
& 66.5 (4.6)
& 64.5 (4.1)
& 58.3 (3.2)
& 80.3 (3.8) \\

\textbf{SemiAMC}
& 83.0 (3.5)
& 63.6 (3.5)
& 75.0 (4.8)
& 72.0 (5.1)
& 70.0 (4.2)
& 68.9 (3.8)
& 63.7 (3.3)
& 81.0 (9.8) \\

\textbf{DAE}
& 49.9 (4.5)
& 27.7 (3.9)
& 41.2 (4.8)
& 38.6 (4.9)
& 37.9 (5.0)
& 32.5 (4.8)
& 28.3 (4.1)
& 49.5 (9.5) \\

\textbf{CSSL-AMC (enc. frozen)}
& 50.3 (6.6)
& 36.4 (5.3)
& 48.5 (5.6)
& 45.9 (6.1)
& 43.3 (5.2)
& 41.0 (5.3)
& 35.8 (5.5)
& 51.1 (6.0) \\

\textbf{CSSL-AMC (enc. not frozen)}
& 90.0 (3.7)
& 74.1 (2.5)
& 84.2 (4.4)
& 81.0 (4.7)
& 80.3 (3.8)
& 78.8 (2.9)
& 74.1 (2.4)
& 88.3 (9.2) \\

\textbf{FedSSL-AMC (Ours)}
& 87.4 (3.7)
& 84.6 (6.0)
& 80.7 (4.5)
& 77.7 (4.8)
& 76.1 (4.2)
& 75.2 (3.6)
& 69.5 (2.9)
& 84.3 (3.8) \\

\bottomrule
\end{tabular}
\end{table*}


\begin{figure}[!htbp]
    \centering
    \begin{subfigure}[!]{\columnwidth}
        \centering
        \includegraphics[width=0.95\columnwidth]{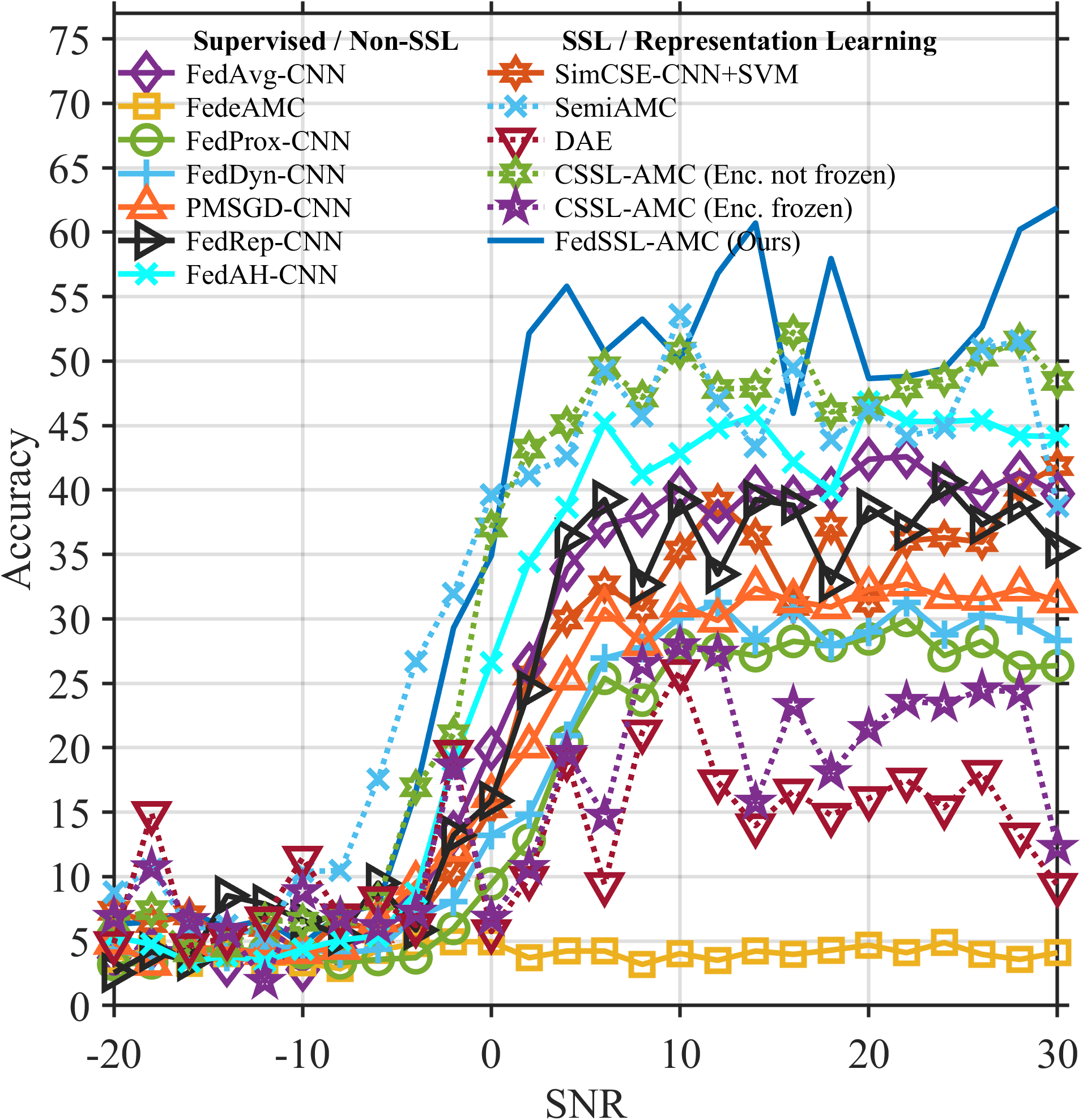 }
        \caption{$\bar{\alpha}=0.1$}
        \label{fig:5a}
    \end{subfigure}
    \hfill
    \begin{subfigure}[!]{\columnwidth}
        \centering
        \includegraphics[width=0.95\columnwidth]{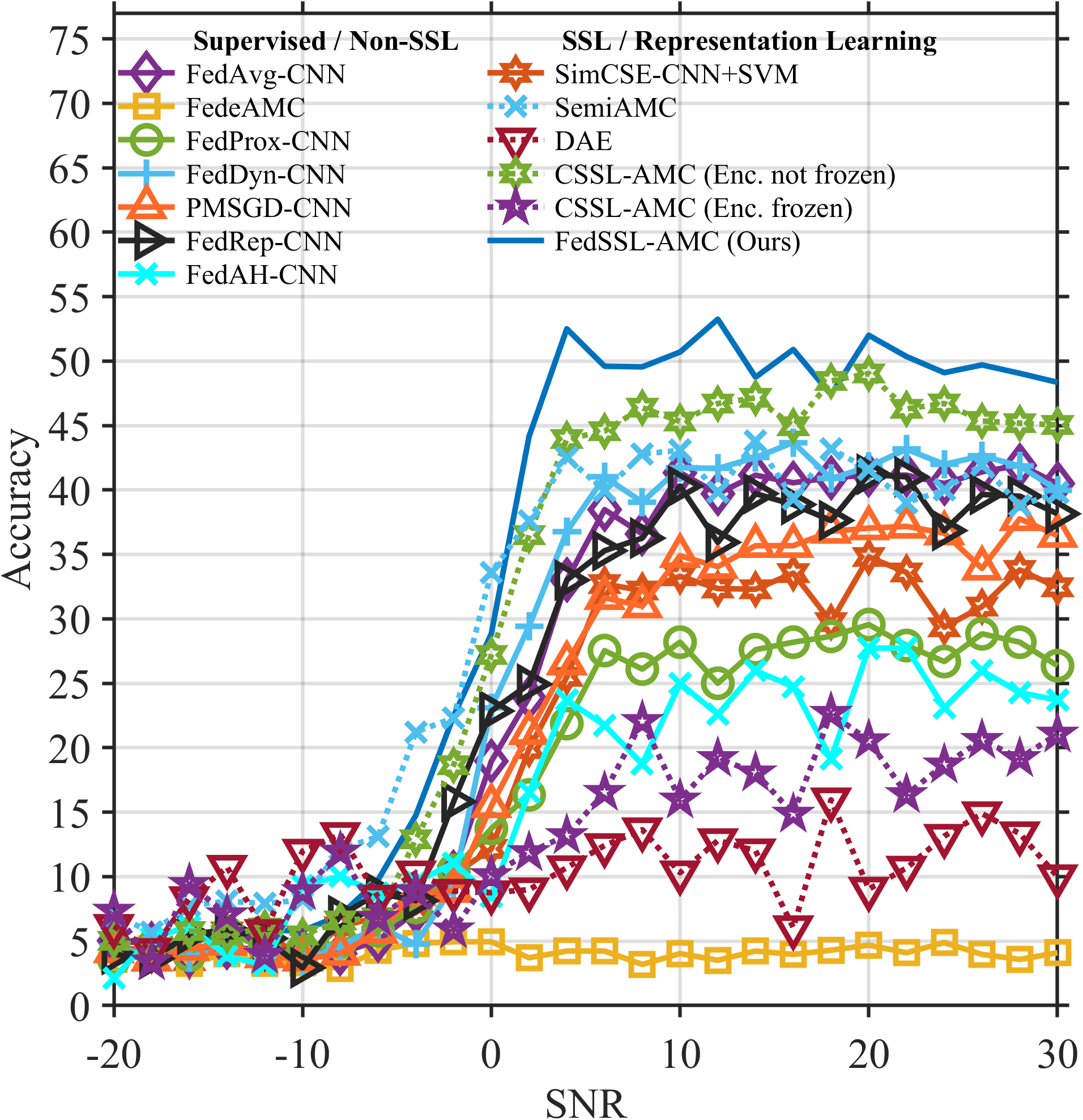}
        \caption{$\bar{\alpha}=0.5$}
        \label{fig:5b}
    \end{subfigure}
    \caption{Test accuracy versus SNR on the DeepSig dataset for $\bar{\alpha}=0.1$ and $\bar{\alpha}=0.5$.}
    \label{fig:deepsig}
\end{figure}

\color{black}

\subsection{Complexity Analysis}

\textbf{Contrastive overhead and communication.} The main additional computational cost of \textbf{FedSSL-AMC} arises from the contrastive objective, which compares each anchor with 10 negative samples. Its communication and server-side aggregation costs per round remain comparable to those of standard FL methods such as \textbf{FedAvg} and \textbf{FedProx}. \textbf{FedDyn} incurs additional client and server computation through its regularization and correction terms, while \textbf{PMSGD}, \textbf{FedAH}, and \textbf{FedRep} do not increase communication volume relative to \textbf{FedAvg}. However, \textbf{PMSGD} requires a client-side grid search over learning rates, and \textbf{FedAH} incurs additional local computation to train aggregation weights. The supervised baselines use 1{,}000 communication rounds, whereas \textbf{FedSSL-AMC} achieves higher accuracy in 10 rounds. This indicates substantially greater communication-round efficiency, although \textbf{FedSSL-AMC} performs more local computation per round. The contrastive baselines \textbf{CSSL-AMC} and \textbf{SemiAMC} have comparable algorithmic complexity when controlling for encoder architecture. However, \textbf{CSSL-AMC}'s augmentation strategy requires high-SNR samples, limiting its applicability when much of the training data has medium or low SNR.

\textbf{Parameter count and computation.} The proposed encoder has substantially fewer trainable parameters than the supervised CNN baseline (0.247M versus 1.78M), but requires more inference computation (473.56 versus 17.76 MFLOPs) because of its deeper causal, time-dilated architecture. The contrastive objective further increases training-time computation. Despite this higher computational cost, the encoder remains practical for edge deployment and offers improved use of unlabeled data together with substantially fewer communication rounds.

\textbf{Encoder design.} Let $\Psi$ denote the number of 1D convolutional layers, $\chi$ the kernel size, and $\Lambda$ the input sequence length. Treating the channel width as fixed, the computational complexity of the causal CNN encoder is $\mathcal{O}(\chi\Psi\Lambda)$. Dilated convolutions expand the receptive field without increasing the number of kernel evaluations per layer, preserving linear scaling with sequence length. By contrast, standard self-attention has quadratic complexity in $\Lambda$, which can be less suitable for long sequences.


\section{Conclusion}

We introduced \textbf{FedSSL-AMC}, a federated self-supervised framework for automatic modulation classification under heterogeneous client data and limited label availability. Our analysis establishes noise-dependent gradient bounds and convergence guarantees for a closely related proximal training variant under non-IID data, together with a high-probability separability result for the downstream classifier. Empirically, \textbf{FedSSL-AMC} consistently outperforms the supervised baselines, particularly when unlabeled data are abundant and labeled samples are scarce or unevenly distributed across clients.

Future work will investigate whether clustering clients by data distribution, for example through group-wise contrastive learning or adaptive aggregation, can further improve performance. The framework may also be applicable to related RF tasks such as spectrum sensing, interference classification, and waveform identification, which share challenges including distributed sensing, privacy constraints, non-IID channel conditions, limited uplink capacity, and scarce labels. In these settings, a shared causal CNN encoder could learn representations from unlabeled I/Q data, while lightweight local classifiers support adaptation to client-specific propagation environments and hardware characteristics.

\bibliographystyle{IEEEbib}
\bibliography{refs}

\appendices
\onecolumn
\section*{Appendix Outline}
This appendix provides proofs of the results omitted from the main text, together with additional modeling and dataset details.
\begin{itemize}
\item Section~\ref{pfslemma} proves Lemmas~\ref{input_enc_lip}, \ref{lem_param_lip}, \ref{enc_out_bound}, and~\ref{Lemma Smooth Loss}.

\item Section~\ref{secthmgradb} proves Theorem~\ref{grad_SNR_bound}, which bounds the expected gradient norms under additive noise.

\item Section~\ref{fproxpf} proves Theorem~\ref{fedprox_th4_adapt}, establishing the convergence result for the proximal federated training variant.

\item Section~\ref{softplusactrespf} proves the corresponding results for a causal CNN encoder with Softplus activation.

\item Section~\ref{misc} describes the mobility-induced carrier-frequency-offset model and illustrates how varying $\bar{\alpha}$ affects the client label distributions.
\end{itemize}


\section{Proofs of theoretical results}

\subsection{Proofs of supporting lemmas}
\label{pfslemma}

This subsection proves the supporting lemmas used in the main theoretical results. We begin with Lemma~\ref{input_enc_lip}, using the following auxiliary fact:
The ReLU activation is $1$-Lipschitz.

\begin{proof}
For any scalars $x,x'\in\mathbb{R}$,
\[
\left|\operatorname{ReLU}(x)-\operatorname{ReLU}(x')\right|
\leq
|x-x'|.
\]
Indeed, if $x$ and $x'$ have the same sign, the inequality follows directly. If, without loss of generality, $x\geq 0$ and $x'<0$, then
\[
\left|\operatorname{ReLU}(x)-\operatorname{ReLU}(x')\right|
=
|x|
\leq
|x-x'|.
\]
Since ReLU acts elementwise, summing the squared scalar inequalities over all coordinates gives
\[
\left\|
\operatorname{ReLU}(x)-\operatorname{ReLU}(x')
\right\|_2
\leq
\left\|x-x'\right\|_2.
\]
Thus, ReLU is $1$-Lipschitz.
\end{proof}


\subsubsection{Proof of Lemma~\ref{input_enc_lip}}

\begin{replemma}{input_enc_lip}
The encoder function $f(x;\bar{\Theta}_1,\dots,\bar{\Theta}_\Psi)$ is $L_x$-Lipschitz in $x$, with $L_x=(R_{\mathrm{mat}})^\Psi$; that is,
\[
\|f(x;\bar{\Theta}_1,\dots,\bar{\Theta}_\Psi)
-f(x';\bar{\Theta}_1,\dots,\bar{\Theta}_\Psi)\|_2
\leq
L_x\|x-x'\|_2.
\]
\end{replemma}

\begin{proof}
Let $\operatorname{Pad}(\cdot)$ denote the zero-padding operation, so that
\[
\bar r_{l-1}=\operatorname{Pad}(\tilde r_{l-1}),
\]
where $\tilde r_{l-1}$ is the unpadded input to layer $l$. Since zero padding preserves the Euclidean norm and ReLU is $1$-Lipschitz, we can bound the Lipschitz constant of the encoder layer by layer.

We prove by induction that, for each $l=1,\ldots,\Psi-1$, the mapping from the encoder input to the output of layer $l$ is
$(\prod_{i=1}^{l}\|\bar{\Theta}_i\|_2)$-Lipschitz.

\textit{Base case:} For $l=1$,
\begin{align}
\|\tilde r_1-\tilde r_1'\|_2
&=
\|\operatorname{ReLU}(\bar{\Theta}_1\bar r_0)
-\operatorname{ReLU}(\bar{\Theta}_1\bar r_0')\|_2
\nonumber\\
&\leq
\|\bar{\Theta}_1\bar r_0-\bar{\Theta}_1\bar r_0'\|_2
\nonumber\\
&=
\|\bar{\Theta}_1(\bar r_0-\bar r_0')\|_2
\nonumber\\
&\leq
\|\bar{\Theta}_1\|_2
\|\bar r_0-\bar r_0'\|_2
\nonumber\\
&=
\|\bar{\Theta}_1\|_2
\|\operatorname{Pad}(\tilde r_0)
-\operatorname{Pad}(\tilde r_0')\|_2
\nonumber\\
&=
\|\bar{\Theta}_1\|_2
\|\tilde r_0-\tilde r_0'\|_2
\nonumber\\
&=
\|\bar{\Theta}_1\|_2
\|x-x'\|_2.
\end{align}

\textit{Inductive step:} Suppose that, for some $j\geq 2$,
\[
\|\tilde r_{j-1}-\tilde r_{j-1}'\|_2
\leq
\prod_{i=1}^{j-1}\|\bar{\Theta}_i\|_2
\|x-x'\|_2.
\]
Then
\begin{align}
\|\tilde r_j-\tilde r_j'\|_2
&=
\|\operatorname{ReLU}(\bar{\Theta}_j\bar r_{j-1})
-\operatorname{ReLU}(\bar{\Theta}_j\bar r_{j-1}')\|_2
\nonumber\\
&\leq
\|\bar{\Theta}_j\bar r_{j-1}
-\bar{\Theta}_j\bar r_{j-1}'\|_2
\nonumber\\
&=
\|\bar{\Theta}_j(\bar r_{j-1}-\bar r_{j-1}')\|_2
\nonumber\\
&\leq
\|\bar{\Theta}_j\|_2
\|\bar r_{j-1}-\bar r_{j-1}'\|_2
\nonumber\\
&=
\|\bar{\Theta}_j\|_2
\|\operatorname{Pad}(\tilde r_{j-1})
-\operatorname{Pad}(\tilde r_{j-1}')\|_2
\nonumber\\
&=
\|\bar{\Theta}_j\|_2
\|\tilde r_{j-1}-\tilde r_{j-1}'\|_2
\nonumber\\
&\leq
\prod_{i=1}^{j}\|\bar{\Theta}_i\|_2
\|x-x'\|_2.
\end{align}

Therefore, the mapping from the encoder input to the output of layer $\Psi-1$ is
$(\prod_{i=1}^{\Psi-1}\|\bar{\Theta}_i\|_2)$-Lipschitz. For the final layer,
\begin{align}
&\|f(x;\bar{\Theta}_1,\dots,\bar{\Theta}_\Psi)
-f(x';\bar{\Theta}_1,\dots,\bar{\Theta}_\Psi)\|_2
\nonumber\\
&=
\|\bar{\Theta}_\Psi
(\bar r_{\Psi-1}-\bar r_{\Psi-1}')\|_2
\nonumber\\
&\leq
\|\bar{\Theta}_\Psi\|_2
\|\bar r_{\Psi-1}-\bar r_{\Psi-1}'\|_2
\nonumber\\
&\leq
\prod_{i=1}^{\Psi}
\|\bar{\Theta}_i\|_2
\|x-x'\|_2.
\end{align}

Finally, Assumption~\ref{SpectralNormBound} gives
\[
\|\bar{\Theta}_i\|_2\leq R_{\mathrm{mat}}
\qquad\text{for all }i,
\]
and hence
\[
\|f(x;\bar{\Theta}_1,\dots,\bar{\Theta}_\Psi)
-f(x';\bar{\Theta}_1,\dots,\bar{\Theta}_\Psi)\|_2
\leq
(R_{\mathrm{mat}})^\Psi
\|x-x'\|_2.
\]
Thus, the encoder is $L_x$-Lipschitz in $x$ with
$L_x=(R_{\mathrm{mat}})^\Psi$.
\end{proof}

\subsubsection{Proofs of Lemmas~\ref{lem_param_lip}--\ref{enc_out_bound}
and related corollaries}

To prove Lemma~\ref{lem_param_lip}, we first establish the following
bound on the layer activations.

\begin{lemma}[Activation bound]
\label{lem_activation_bound}
Every element of the encoder activation at layer $l$ is bounded in
magnitude by
\[
\zeta\sqrt{P}\,(\hat{c}\chi R)^l.
\]
\end{lemma}

\begin{proof}
We proceed by induction on $l$.

\textit{Base case:} For $l=1$, each convolutional output is the inner
product of a kernel vector and an input vector, each of dimension
$\hat{c}\chi$. Every kernel entry is bounded in magnitude by $R$, and
every input entry is bounded by $\zeta\sqrt{P}$. Therefore, by the
Cauchy--Schwarz inequality,
\begin{align}
\left|\left\langle \theta,u\right\rangle\right|
&\leq
\|\theta\|_2\|u\|_2 \nonumber\\
&\leq
R\sqrt{\hat{c}\chi}\,
\zeta\sqrt{P}\sqrt{\hat{c}\chi}
\nonumber\\
&=
\zeta\sqrt{P}\,\hat{c}\chi R.
\end{align}
Because ReLU does not increase the magnitude of its input, every
activation at the first layer is bounded by
$\zeta\sqrt{P}(\hat{c}\chi R)$.

\textit{Inductive step:} Suppose that every activation at layer $l-1$
is bounded in magnitude by
\[
\zeta\sqrt{P}\,(\hat{c}\chi R)^{l-1}.
\]
Each pre-activation at layer $l$ is again an inner product of dimension
$\hat{c}\chi$. Its kernel entries are bounded by $R$, while its input
entries are bounded by
$\zeta\sqrt{P}(\hat{c}\chi R)^{l-1}$. Hence,
\begin{align}
\left|\left\langle \theta,u\right\rangle\right|
&\leq
R\sqrt{\hat{c}\chi}\,
\zeta\sqrt{P}(\hat{c}\chi R)^{l-1}
\sqrt{\hat{c}\chi}
\nonumber\\
&=
\zeta\sqrt{P}\,(\hat{c}\chi R)^l.
\end{align}
Applying ReLU preserves this upper bound. The result therefore holds
for every layer $l$.
\end{proof}

\begin{replemma}{lem_param_lip}
The encoder function
$f(x;\bar{\Theta}_1,\dots,\bar{\Theta}_\Psi)
\equiv
f(x;\bar{\theta}_1,\dots,\bar{\theta}_\Psi)$
is $L_{\theta}(l)$-Lipschitz with respect to the parameter vector
$\bar{\theta}_l$, where
{\color{black}{\[
L_{\theta}(l)
=
\sqrt{T_l\hat{c}\chi}\,\zeta\sqrt{P}\,
(\hat{c}\chi R)^{l-1}
(R_{\mathrm{mat}})^{\Psi-l}.
\]}}
\end{replemma}

\begin{proof}
Consider two encoders that differ only in their layer-$l$ parameter
vectors, denoted by $\bar{\theta}_l$ and $\bar{\theta}_l'$. By
\eqref{eqnactivationparam} and the $1$-Lipschitz continuity of ReLU,
\begin{align}
\|\bar r_l-\bar r_l'\|_2
&=
\left\|
\operatorname{ReLU}
\left(\bar{\mathbf R}_{l-1}\bar{\theta}_l\right)
-
\operatorname{ReLU}
\left(\bar{\mathbf R}_{l-1}\bar{\theta}_l'\right)
\right\|_2
\nonumber\\
&\leq
\left\|
\bar{\mathbf R}_{l-1}
(\bar{\theta}_l-\bar{\theta}_l')
\right\|_2
\nonumber\\
&\leq
\|\bar{\mathbf R}_{l-1}\|_2
\|\bar{\theta}_l-\bar{\theta}_l'\|_2.
\end{align}

By Lemma~\ref{lem_activation_bound}, every entry of the input to layer
$l$ is bounded in magnitude by
\[
\zeta\sqrt{P}(\hat{c}\chi R)^{l-1}.
\]
Each block $\hat{\mathbf R}_{l-1}$ has $T_l$ rows and
$\hat{c}\chi$ entries per row. Since
$\bar{\mathbf R}_{l-1}$ is block diagonal with copies of
$\hat{\mathbf R}_{l-1}$,
\[
\|\bar{\mathbf R}_{l-1}\|_2
=
\|\hat{\mathbf R}_{l-1}\|_2
\leq
\|\hat{\mathbf R}_{l-1}\|_F.
\]
Therefore,
\begin{align}
\|\bar{\mathbf R}_{l-1}\|_2
&\leq
\sqrt{T_l\hat{c}\chi}\,
\zeta\sqrt{P}
(\hat{c}\chi R)^{l-1}
\end{align}
\begin{align}
\color{black}\|\bar r_l-\bar r_l'\|_2
\leq
\sqrt{T_l\hat{c}\chi}\zeta\sqrt{P}
(\hat{c}\chi R)^{l-1}
\|\bar{\theta}_l-\bar{\theta}_l'\|_2.
\end{align}

The perturbation then propagates through layers $l+1,\dots,\Psi$.
As in the proof of Lemma~\ref{input_enc_lip}, each subsequent layer
contributes at most its spectral norm. Thus,
\begin{align}
&\|f(x;\bar{\theta}_1,\dots,\bar{\theta}_l,\dots,\bar{\theta}_\Psi)
-
f(x;\bar{\theta}_1,\dots,\bar{\theta}_l',\dots,\bar{\theta}_\Psi)\|_2
\nonumber\\
& \color{black} \leq
\left(
\prod_{j=l+1}^{\Psi}
\|\bar{\Theta}_j\|_2
\right)
\sqrt{T_l\hat{c}\chi}\zeta\sqrt{P}
(\hat{c}\chi R)^{l-1}
\|\bar{\theta}_l-\bar{\theta}_l'\|_2
\nonumber\\
&\color{black} \leq
(R_{\mathrm{mat}})^{\Psi-l}
\sqrt{T_l\hat{c}\chi}\zeta\sqrt{P}
(\hat{c}\chi R)^{l-1}
\|\bar{\theta}_l-\bar{\theta}_l'\|_2.
\end{align}

Therefore, the encoder is $L_\theta(l)$-Lipschitz with respect to
$\bar{\theta}_l$.
\end{proof}

\begin{replemma}{enc_out_bound}
The encoder output is bounded elementwise by
\[
\zeta\sqrt{P}\,(\hat{c}\chi R)^\Psi.
\]
\end{replemma}

\begin{proof}
By Lemma~\ref{lem_activation_bound}, every activation at layer $l$ is bounded in magnitude by
\[
\zeta\sqrt{P}\,(\hat{c}\chi R)^l.
\]
Setting $l=\Psi$ gives the stated bound on the encoder output.
\end{proof}

\begin{repcorollary}{col_jac_bound}
Wherever it exists, the Jacobian of the encoder
$f(x;\bar{\theta}_1,\dots,\bar{\theta}_\Psi)$ with respect to
$\bar{\theta}_l$ has spectral norm bounded by $L_\theta(l)$; that is,
\[
\left\|J_{\bar{\theta}_l}(x)\right\|_2
\leq
L_\theta(l).
\]
\end{repcorollary}

\begin{proof}
Fix an input $x$ and a layer index $l$, and define
\[
g(\bar{\theta}_l)
\triangleq
f(x;\bar{\theta}_1,\dots,\bar{\theta}_l,\dots,\bar{\theta}_\Psi),
\]
with all parameter blocks $\{\bar{\theta}_j\}_{j\neq l}$ held fixed.
By Lemma~\ref{lem_param_lip}, $g$ is $L_\theta(l)$-Lipschitz; hence, for
all $\bar{\theta}_l$ and $\bar{\theta}_l'$,
\[
\|g(\bar{\theta}_l)-g(\bar{\theta}_l')\|_2
\leq
L_\theta(l)\,
\|\bar{\theta}_l-\bar{\theta}_l'\|_2.
\]

Assume that $g$ is differentiable at the point of interest. For any
direction $u$ with $\|u\|_2=1$ and any $t\neq 0$,
\[
\frac{
\|g(\bar{\theta}_l+t u)-g(\bar{\theta}_l)\|_2
}{|t|}
\leq
L_\theta(l).
\]
Taking the limit as $t\to 0$ gives
\[
\|J_{\bar{\theta}_l}(x)u\|_2
\leq
L_\theta(l).
\]
Therefore,
\[
\|J_{\bar{\theta}_l}(x)\|_2
=
\sup_{\|u\|_2=1}
\|J_{\bar{\theta}_l}(x)u\|_2
\leq
L_\theta(l).
\]
\end{proof}

\begin{repcorollary}{col_grad_loss_bound}
The gradients of the loss function with respect to the encoder outputs
are bounded by some constant $G>0$; that is,
\begin{align}
\left\|\nabla_{z^{ref}}
\mathcal{L}(z^{ref},z^{pos},\{z_k^{neg}\}_{k=1}^K)
\right\|_2
&\leq G,\\
\left\|\nabla_{z^{pos}}
\mathcal{L}(z^{ref},z^{pos},\{z_k^{neg}\}_{k=1}^K)
\right\|_2
&\leq G,\\
\left\|\nabla_{z_k^{neg}}
\mathcal{L}(z^{ref},z^{pos},\{z_k^{neg}\}_{k=1}^K)
\right\|_2
&\leq G,
\qquad k=1,\dots,K.
\end{align}
\end{repcorollary}

\begin{proof}
Let $B_z>0$ satisfy
\[
\|z^{ref}\|_2,\;
\|z^{pos}\|_2,\;
\|z_k^{neg}\|_2
\leq B_z,
\qquad k=1,\dots,K.
\]
For the triplet loss
\[
\mathcal{L}
=
-\log \sigma\big((z^{ref})^Tz^{pos}\big)
-\sum_{k=1}^K
\log \sigma\big(-(z^{ref})^Tz_k^{neg}\big),
\]
the gradients are
\begin{align}
\nabla_{z^{ref}}\mathcal{L}
&=
-\left[1-\sigma\big((z^{ref})^Tz^{pos}\big)\right]z^{pos}
+\sum_{k=1}^K
\sigma\big((z^{ref})^Tz_k^{neg}\big)z_k^{neg},
\\
\nabla_{z^{pos}}\mathcal{L}
&=
-\left[1-\sigma\big((z^{ref})^Tz^{pos}\big)\right]z^{ref},
\\
\nabla_{z_k^{neg}}\mathcal{L}
&=
\sigma\big((z^{ref})^Tz_k^{neg}\big)z^{ref}.
\end{align}
Since $0\leq \sigma(a)\leq 1$ for all $a\in\mathbb{R}$,
\begin{align}
\|\nabla_{z^{ref}}\mathcal{L}\|_2
&\leq
\|z^{pos}\|_2
+\sum_{k=1}^K\|z_k^{neg}\|_2
\leq
(K+1)B_z,
\\
\|\nabla_{z^{pos}}\mathcal{L}\|_2
&\leq
\|z^{ref}\|_2
\leq B_z,
\\
\|\nabla_{z_k^{neg}}\mathcal{L}\|_2
&\leq
\|z^{ref}\|_2
\leq B_z.
\end{align}
Thus, the claimed bounds hold with
\[
G=(K+1)B_z.
\]
\end{proof}

\subsubsection{Proof of Lemma~\ref{Lemma Smooth Loss}}

\begin{replemma}{Lemma Smooth Loss}
The triplet loss function
$\mathcal{L}(z^{ref},z^{pos},\{z_k^{neg}\}_{k=1}^K)$ is
$\Gamma$-smooth with respect to the concatenated vector
$[z^{ref},z^{pos},z_1^{neg},\dots,z_K^{neg}]$, where
$\Gamma=\sqrt{K+2}\max\{\Gamma_{ref},\Gamma_{pn}\}$ and
{\setlength{\abovedisplayskip}{4pt}
 \setlength{\belowdisplayskip}{4pt}
 \setlength{\abovedisplayshortskip}{2pt}
 \setlength{\belowdisplayshortskip}{2pt}
\begin{align}
\Gamma_{ref}
&=
2(T_{\Psi}\zeta)^2P(\hat{c}\chi R)^{2\Psi}K^2
\nonumber\\
&\quad+
\left(4(T_{\Psi}\zeta)^2P(\hat{c}\chi R)^{2\Psi}+1\right)K
\nonumber\\
&\quad+
2(T_{\Psi}\zeta)^2P(\hat{c}\chi R)^{2\Psi}+1,
\\
\Gamma_{pn}
&=
2(K+1)(T_{\Psi}\zeta)^2P(\hat{c}\chi R)^{2\Psi}+1.
\end{align}}
\end{replemma}

\begin{proof}
We prove that the triplet loss is smooth with respect to the encoder
outputs associated with the reference, positive, and negative signals.
Define
\begin{align}
g_{\mathrm{loss}}
\left(\tilde z^+,\{\tilde z_k^-\}_{k=1}^K\right)
&=
-\log\left(\sigma(\tilde z^+)\right)
-\sum_{k=1}^K
\log\left(\sigma(-\tilde z_k^-)\right).
\end{align}
Each term of $g_{\mathrm{loss}}$ is a scalar logistic loss whose second
derivative is bounded by $1/4$. Hence, the Hessian of
$g_{\mathrm{loss}}$ is diagonal with spectral norm at most $1/4$.
In particular, $g_{\mathrm{loss}}$ is $1$-smooth with respect to
$\tilde z^+,\tilde z_1^-,\dots,\tilde z_K^-$.

The triplet loss can be written as
\[
\mathcal{L}
\left(z^{ref},z^{pos},\{z_k^{neg}\}_{k=1}^K\right)
=
g_{\mathrm{loss}}
\left(
(z^{ref})^Tz^{pos},
\{(z^{ref})^Tz_k^{neg}\}_{k=1}^K
\right).
\]
Its smoothness with respect to the encoder outputs requires these
outputs to be bounded. By Lemma~\ref{enc_out_bound}, every encoder
output is elementwise bounded by
$\zeta\sqrt{P}(\hat{c}\chi R)^\Psi$. Since the encoder output contains
$T_\Psi$ entries,
\[
\|z\|_2
\leq
\sqrt{T_\Psi}\,
\zeta\sqrt{P}(\hat{c}\chi R)^\Psi
\leq
T_\Psi\zeta\sqrt{P}(\hat{c}\chi R)^\Psi.
\]
For notational convenience, define
\[
B_{\mathrm{enc}}
=
T_\Psi\zeta\sqrt{P}(\hat{c}\chi R)^\Psi.
\]

The gradient of the loss with respect to the concatenated encoder
outputs is
\begin{align}
&\nabla
\mathcal{L}
\left(z^{ref},z^{pos},\{z_k^{neg}\}_{k=1}^K\right)
\nonumber\\
&=
\begin{bmatrix}
z^{pos}\nabla_{\tilde z^+}g_{\mathrm{loss}}
+
\displaystyle\sum_{k=1}^K
z_k^{neg}\nabla_{\tilde z_k^-}g_{\mathrm{loss}}
\\[2mm]
z^{ref}\nabla_{\tilde z^+}g_{\mathrm{loss}}
\\
z^{ref}\nabla_{\tilde z_1^-}g_{\mathrm{loss}}
\\
\vdots
\\
z^{ref}\nabla_{\tilde z_K^-}g_{\mathrm{loss}}
\end{bmatrix},
\end{align}
where all derivatives of $g_{\mathrm{loss}}$ are evaluated at
\[
\left(
(z^{ref})^Tz^{pos},
\{(z^{ref})^Tz_k^{neg}\}_{k=1}^K
\right).
\]

Consider another collection of encoder outputs
$z^{ref'}$, $z^{pos'}$, and $\{z_k^{neg'}\}_{k=1}^K$. By the triangle
inequality,
\begin{align}
&\left\|
\nabla\mathcal{L}
\left(z^{ref},z^{pos},\{z_k^{neg}\}_{k=1}^K\right)
-
\nabla\mathcal{L}
\left(z^{ref'},z^{pos'},\{z_k^{neg'}\}_{k=1}^K\right)
\right\|_2
\nonumber\\
&\leq
\left\|
z^{pos}\nabla_{\tilde z^+}g_{\mathrm{loss}}
-
z^{pos'}\nabla_{\tilde z^+}g_{\mathrm{loss}}'
\right\|_2
\nonumber\\
&\quad+
\sum_{k=1}^K
\left\|
z_k^{neg}\nabla_{\tilde z_k^-}g_{\mathrm{loss}}
-
z_k^{neg'}\nabla_{\tilde z_k^-}g_{\mathrm{loss}}'
\right\|_2
\nonumber\\
&\quad+
\left\|
z^{ref}\nabla_{\tilde z^+}g_{\mathrm{loss}}
-
z^{ref'}\nabla_{\tilde z^+}g_{\mathrm{loss}}'
\right\|_2
\nonumber\\
&\quad+
\sum_{k=1}^K
\left\|
z^{ref}\nabla_{\tilde z_k^-}g_{\mathrm{loss}}
-
z^{ref'}\nabla_{\tilde z_k^-}g_{\mathrm{loss}}'
\right\|_2,
\end{align}
where the unprimed derivatives are evaluated at
\[
\left(
(z^{ref})^Tz^{pos},
\{(z^{ref})^Tz_k^{neg}\}_{k=1}^K
\right),
\]
and the primed derivatives are evaluated at
\[
\left(
(z^{ref'})^Tz^{pos'},
\{(z^{ref'})^Tz_k^{neg'}\}_{k=1}^K
\right).
\]

For the first term, add and subtract
$z^{pos'}\nabla_{\tilde z^+}g_{\mathrm{loss}}$ to obtain
\begin{align}
&\left\|
z^{pos}\nabla_{\tilde z^+}g_{\mathrm{loss}}
-
z^{pos'}\nabla_{\tilde z^+}g_{\mathrm{loss}}'
\right\|_2
\nonumber\\
&\leq
\|z^{pos}-z^{pos'}\|_2
\left\|\nabla_{\tilde z^+}g_{\mathrm{loss}}\right\|_2
\nonumber\\
&\quad+
\|z^{pos'}\|_2
\left\|
\nabla_{\tilde z^+}g_{\mathrm{loss}}
-
\nabla_{\tilde z^+}g_{\mathrm{loss}}'
\right\|_2
\nonumber\\
&\leq
\|z^{pos}-z^{pos'}\|_2
+
B_{\mathrm{enc}}
\left\|
\nabla_{\tilde z^+}g_{\mathrm{loss}}
-
\nabla_{\tilde z^+}g_{\mathrm{loss}}'
\right\|_2,
\end{align}
where we used
$\|\nabla_{\tilde z^+}g_{\mathrm{loss}}\|_2\leq1$ and
$\|z^{pos'}\|_2\leq B_{\mathrm{enc}}$.

Similarly,
\begin{align}
&\sum_{k=1}^K
\left\|
z_k^{neg}\nabla_{\tilde z_k^-}g_{\mathrm{loss}}
-
z_k^{neg'}\nabla_{\tilde z_k^-}g_{\mathrm{loss}}'
\right\|_2
\nonumber\\
&\leq
\sum_{k=1}^K
\Bigg(
\|z_k^{neg}-z_k^{neg'}\|_2
+
B_{\mathrm{enc}}
\left\|
\nabla_{\tilde z_k^-}g_{\mathrm{loss}}
-
\nabla_{\tilde z_k^-}g_{\mathrm{loss}}'
\right\|_2
\Bigg),
\end{align}
and
\begin{align}
&\left\|
z^{ref}\nabla_{\tilde z^+}g_{\mathrm{loss}}
-
z^{ref'}\nabla_{\tilde z^+}g_{\mathrm{loss}}'
\right\|_2
\nonumber\\
&\leq
\|z^{ref}-z^{ref'}\|_2
+
B_{\mathrm{enc}}
\left\|
\nabla_{\tilde z^+}g_{\mathrm{loss}}
-
\nabla_{\tilde z^+}g_{\mathrm{loss}}'
\right\|_2,
\end{align}
as well as
\begin{align}
&\sum_{k=1}^K
\left\|
z^{ref}\nabla_{\tilde z_k^-}g_{\mathrm{loss}}
-
z^{ref'}\nabla_{\tilde z_k^-}g_{\mathrm{loss}}'
\right\|_2
\nonumber\\
&\leq
\sum_{k=1}^K
\Bigg(
\|z^{ref}-z^{ref'}\|_2
+
B_{\mathrm{enc}}
\left\|
\nabla_{\tilde z_k^-}g_{\mathrm{loss}}
-
\nabla_{\tilde z_k^-}g_{\mathrm{loss}}'
\right\|_2
\Bigg).
\end{align}

By the $1$-smoothness of $g_{\mathrm{loss}}$,
\begin{align}
&\left\|
\nabla_{\tilde z_k^-}g_{\mathrm{loss}}
-
\nabla_{\tilde z_k^-}g_{\mathrm{loss}}'
\right\|_2
\nonumber\\
&\leq
\left|
(z^{ref})^Tz^{pos}
-
(z^{ref'})^Tz^{pos'}
\right|
\nonumber\\
&\quad+
\sum_{j=1}^K
\left|
(z^{ref})^Tz_j^{neg}
-
(z^{ref'})^Tz_j^{neg'}
\right|.
\end{align}
The same bound holds for
$\|\nabla_{\tilde z^+}g_{\mathrm{loss}}
-\nabla_{\tilde z^+}g_{\mathrm{loss}}'\|_2$.

For the positive-pair inner product,
\begin{align}
&(z^{ref})^Tz^{pos}
-
(z^{ref'})^Tz^{pos'}
\nonumber\\
&=
(z^{ref}-z^{ref'})^Tz^{pos}
+
(z^{ref'})^T(z^{pos}-z^{pos'}),
\end{align}
and therefore
\begin{align}
&\left|
(z^{ref})^Tz^{pos}
-
(z^{ref'})^Tz^{pos'}
\right|
\nonumber\\
&\leq
\|z^{pos}\|_2\|z^{ref}-z^{ref'}\|_2
+
\|z^{ref'}\|_2\|z^{pos}-z^{pos'}\|_2
\nonumber\\
&\leq
B_{\mathrm{enc}}\|z^{ref}-z^{ref'}\|_2
+
B_{\mathrm{enc}}\|z^{pos}-z^{pos'}\|_2.
\end{align}
Likewise, for each negative sample,
\begin{align}
&\left|
(z^{ref})^Tz_k^{neg}
-
(z^{ref'})^Tz_k^{neg'}
\right|
\nonumber\\
&\leq
B_{\mathrm{enc}}\|z^{ref}-z^{ref'}\|_2
+
B_{\mathrm{enc}}\|z_k^{neg}-z_k^{neg'}\|_2.
\end{align}
Consequently,
\begin{align}
&\left\|
\nabla_{\tilde z_k^-}g_{\mathrm{loss}}
-
\nabla_{\tilde z_k^-}g_{\mathrm{loss}}'
\right\|_2
\nonumber\\
&\leq
B_{\mathrm{enc}}(K+1)
\|z^{ref}-z^{ref'}\|_2
+
B_{\mathrm{enc}}
\|z^{pos}-z^{pos'}\|_2
\nonumber\\
&\quad+
B_{\mathrm{enc}}
\sum_{j=1}^K
\|z_j^{neg}-z_j^{neg'}\|_2.
\end{align}
The same bound holds for
$\|\nabla_{\tilde z^+}g_{\mathrm{loss}}
-\nabla_{\tilde z^+}g_{\mathrm{loss}}'\|_2$.

Substituting these bounds into the preceding inequalities gives
\begin{align}
&\left\|
z^{pos}\nabla_{\tilde z^+}g_{\mathrm{loss}}
-
z^{pos'}\nabla_{\tilde z^+}g_{\mathrm{loss}}'
\right\|_2
\nonumber\\
&\leq
B_{\mathrm{enc}}^2(K+1)
\|z^{ref}-z^{ref'}\|_2
\nonumber\\
&\quad+
(B_{\mathrm{enc}}^2+1)
\|z^{pos}-z^{pos'}\|_2
\nonumber\\
&\quad+
B_{\mathrm{enc}}^2
\sum_{k=1}^K
\|z_k^{neg}-z_k^{neg'}\|_2,
\end{align}
\begin{align}
&\sum_{k=1}^K
\left\|
z_k^{neg}\nabla_{\tilde z_k^-}g_{\mathrm{loss}}
-
z_k^{neg'}\nabla_{\tilde z_k^-}g_{\mathrm{loss}}'
\right\|_2
\nonumber\\
&\leq
K(K+1)B_{\mathrm{enc}}^2
\|z^{ref}-z^{ref'}\|_2
\nonumber\\
&\quad+
KB_{\mathrm{enc}}^2
\|z^{pos}-z^{pos'}\|_2
\nonumber\\
&\quad+
(KB_{\mathrm{enc}}^2+1)
\sum_{k=1}^K
\|z_k^{neg}-z_k^{neg'}\|_2,
\end{align}
\begin{align}
&\left\|
z^{ref}\nabla_{\tilde z^+}g_{\mathrm{loss}}
-
z^{ref'}\nabla_{\tilde z^+}g_{\mathrm{loss}}'
\right\|_2
\nonumber\\
&\leq
(B_{\mathrm{enc}}^2K+B_{\mathrm{enc}}^2+1)
\|z^{ref}-z^{ref'}\|_2
\nonumber\\
&\quad+
B_{\mathrm{enc}}^2
\|z^{pos}-z^{pos'}\|_2
\nonumber\\
&\quad+
B_{\mathrm{enc}}^2
\sum_{k=1}^K
\|z_k^{neg}-z_k^{neg'}\|_2,
\end{align}
and
\begin{align}
\sum_{k=1}^K
\left\|
z^{ref}\nabla_{\tilde z_k^-}g_{\mathrm{loss}}
-
z^{ref'}\nabla_{\tilde z_k^-}g_{\mathrm{loss}}'
\right\|_2
& \leq
K(B_{\mathrm{enc}}^2K+B_{\mathrm{enc}}^2+1)
\|z^{ref}-z^{ref'}\|_2
+
KB_{\mathrm{enc}}^2
\|z^{pos}-z^{pos'}\|_2 \nonumber \\
& +
KB_{\mathrm{enc}}^2
\sum_{k=1}^K
\|z_k^{neg}-z_k^{neg'}\|_2.
\end{align}

Grouping the terms yields
\begin{align}
&\left\|
\nabla\mathcal{L}
\left(z^{ref},z^{pos},\{z_k^{neg}\}_{k=1}^K\right)
-
\nabla\mathcal{L}
\left(z^{ref'},z^{pos'},\{z_k^{neg'}\}_{k=1}^K\right)
\right\|_2
\nonumber\\
&\leq
\left[
2B_{\mathrm{enc}}^2K^2
+
(4B_{\mathrm{enc}}^2+1)K
+
2B_{\mathrm{enc}}^2+1
\right]
\|z^{ref}-z^{ref'}\|_2
\nonumber\\
&\quad+
\left[
2(K+1)B_{\mathrm{enc}}^2+1
\right]
\|z^{pos}-z^{pos'}\|_2
\nonumber\\
&\quad+
\left[
2(K+1)B_{\mathrm{enc}}^2+1
\right]
\sum_{k=1}^K
\|z_k^{neg}-z_k^{neg'}\|_2.
\end{align}

Substituting
$B_{\mathrm{enc}}
=T_\Psi\zeta\sqrt{P}(\hat{c}\chi R)^\Psi$ gives
\begin{align}
\Gamma_{ref}
&=
2(T_{\Psi}\zeta)^2P(\hat{c}\chi R)^{2\Psi}K^2
\nonumber\\
&\quad+
\left(
4(T_{\Psi}\zeta)^2P(\hat{c}\chi R)^{2\Psi}+1
\right)K
\nonumber\\
&\quad+
2(T_{\Psi}\zeta)^2P(\hat{c}\chi R)^{2\Psi}+1,
\\
\Gamma_{pn}
&=
2(K+1)(T_{\Psi}\zeta)^2P(\hat{c}\chi R)^{2\Psi}+1.
\end{align}
Let
\[
\widetilde{\Gamma}
=
\max\{\Gamma_{ref},\Gamma_{pn}\}.
\]
Then
\begin{align}
&\left\|
\nabla\mathcal{L}
\left(z^{ref},z^{pos},\{z_k^{neg}\}_{k=1}^K\right)
-
\nabla\mathcal{L}
\left(z^{ref'},z^{pos'},\{z_k^{neg'}\}_{k=1}^K\right)
\right\|_2
\nonumber\\
&\leq
\widetilde{\Gamma}
\left(
\|z^{ref}-z^{ref'}\|_2
+
\|z^{pos}-z^{pos'}\|_2
+
\sum_{k=1}^K
\|z_k^{neg}-z_k^{neg'}\|_2
\right).
\end{align}

Define
\[
z_{\mathrm{stacked}}
=
\begin{bmatrix}
z^{ref}\\
z^{pos}\\
z_1^{neg}\\
\vdots\\
z_K^{neg}
\end{bmatrix},
\qquad
z_{\mathrm{stacked}}'
=
\begin{bmatrix}
z^{ref'}\\
z^{pos'}\\
z_1^{neg'}\\
\vdots\\
z_K^{neg'}
\end{bmatrix}.
\]
By the Cauchy--Schwarz inequality,
\begin{align}
&\|z^{ref}-z^{ref'}\|_2
+
\|z^{pos}-z^{pos'}\|_2
+
\sum_{k=1}^K
\|z_k^{neg}-z_k^{neg'}\|_2
\nonumber\\
&\leq
\sqrt{K+2}\,
\|z_{\mathrm{stacked}}-z_{\mathrm{stacked}}'\|_2.
\end{align}
Therefore,
\begin{align}
&\left\|
\nabla\mathcal{L}(z_{\mathrm{stacked}})
-
\nabla\mathcal{L}(z_{\mathrm{stacked}}')
\right\|_2
\nonumber\\
&\leq
\sqrt{K+2}\,
\widetilde{\Gamma}\,
\|z_{\mathrm{stacked}}-z_{\mathrm{stacked}}'\|_2.
\end{align}
Thus, the triplet loss is $\Gamma$-smooth with
\[
\Gamma
=
\sqrt{K+2}\max\{\Gamma_{ref},\Gamma_{pn}\}.
\]
\end{proof}

\subsection{Proof of Theorem~\ref{grad_SNR_bound}}
\label{secthmgradb}

We first recall the statement of Theorem~\ref{grad_SNR_bound}.

\begin{reptheorem}{grad_SNR_bound}
Let $\gamma$ denote the SNR and $P$ the expected signal power per
channel. Let
\[
z^{ref'}=f(x^{ref'};\Theta),\qquad
z^{pos'}=f(x^{pos'};\Theta),\qquad
z_k^{neg'}=f(x_k^{neg'};\Theta)
\]
denote the noise-perturbed embeddings, and let
$z^{ref}$, $z^{pos}$, and $z_k^{neg}$ denote their noiseless
counterparts. Under the preceding assumptions and lemmas, the following
bounds hold, where the expectation is taken over the additive signal
noise.

\noindent
(1) The expected $\ell_2$-norm of the gradient of the loss with respect
to the layer-$l$ encoder parameters satisfies
\begin{align}
&\mathbb{E}\left[
\left\|
\nabla_{\bar{\theta}_l}
\mathcal{L}
\left(z^{ref'},z^{pos'},\{z_k^{neg'}\}_{k=1}^K\right)
\right\|_2
\right]
\nonumber\\
&\leq
\left\|
\nabla_{\bar{\theta}_l}
\mathcal{L}
\left(z^{ref},z^{pos},\{z_k^{neg}\}_{k=1}^K\right)
\right\|_2
+
2(K+2)L_\theta(l)G
\nonumber\\
&\quad+
L_\theta(l)\Gamma L_x
\sqrt{T(K+2)^3\gamma^{-1}P}.
\end{align}

\noindent
(2) The corresponding expected squared $\ell_2$-norm satisfies
\begin{align}
&\mathbb{E}\left[
\left\|
\nabla_{\bar{\theta}_l}
\mathcal{L}
\left(z^{ref'},z^{pos'},\{z_k^{neg'}\}_{k=1}^K\right)
\right\|_2^2
\right]
\nonumber\\
&\leq
2\left\|
\nabla_{\bar{\theta}_l}
\mathcal{L}
\left(z^{ref},z^{pos},\{z_k^{neg}\}_{k=1}^K\right)
\right\|_2^2
+
16(K+2)^2L_\theta^2(l)G^2
\nonumber\\
&\quad+
4L_\theta^2(l)\Gamma^2L_x^2
T(K+2)^3\gamma^{-1}P.
\end{align}

\noindent
(3) The expected $\ell_2$-norm of the gradient with respect to the full
encoder parameter vector satisfies
\begin{align}
&\mathbb{E}\left[
\left\|
\nabla_{\Theta}
\mathcal{L}
\left(z^{ref'},z^{pos'},\{z_k^{neg'}\}_{k=1}^K\right)
\right\|_2
\right]
\nonumber\\
&\leq
\left\|
\nabla_{\Theta}
\mathcal{L}
\left(z^{ref},z^{pos},\{z_k^{neg}\}_{k=1}^K\right)
\right\|_2
+
2(K+2)G\sum_{l=1}^{\Psi}L_\theta(l)
\nonumber\\
&\quad+
\Gamma L_x
\sqrt{T(K+2)^3\gamma^{-1}P}
\sum_{l=1}^{\Psi}L_\theta(l).
\end{align}

\noindent
(4) The corresponding expected squared $\ell_2$-norm satisfies
\begin{align}
&\mathbb{E}\left[
\left\|
\nabla_{\Theta}
\mathcal{L}
\left(z^{ref'},z^{pos'},\{z_k^{neg'}\}_{k=1}^K\right)
\right\|_2^2
\right]
\nonumber\\
&\leq
2\left\|
\nabla_{\Theta}
\mathcal{L}
\left(z^{ref},z^{pos},\{z_k^{neg}\}_{k=1}^K\right)
\right\|_2^2
+
16(K+2)^2G^2
\sum_{l=1}^{\Psi}L_\theta^2(l)
\nonumber\\
&\quad+
4\Gamma^2L_x^2T(K+2)^3\gamma^{-1}P
\sum_{l=1}^{\Psi}L_\theta^2(l).
\end{align}

Here, $L_\theta(l)$ is the Lipschitz constant of the encoder with
respect to its layer-$l$ parameters, $G$ bounds the loss gradients with
respect to the encoder outputs
(Corollary~\ref{col_grad_loss_bound}), $\Gamma$ is the smoothness
constant of the triplet loss (Lemma~\ref{Lemma Smooth Loss}), $L_x$ is
the Lipschitz constant of the encoder with respect to its inputs, $T$
is the signal length, and $\Psi$ is the number of encoder layers.
\end{reptheorem}

\begin{proof}
For conciseness, define
\begin{align}
\mathcal{L}'
&\triangleq
\mathcal{L}
\left(z^{ref'},z^{pos'},\{z_k^{neg'}\}_{k=1}^K\right),
\\
\mathcal{L}
&\triangleq
\mathcal{L}
\left(z^{ref},z^{pos},\{z_k^{neg}\}_{k=1}^K\right).
\end{align}
By the chain rule,
\begin{align}
\nabla_{\bar{\theta}_l}\mathcal{L}'
&=
J_{\bar{\theta}_l}(x^{ref'})^T
\nabla_{z^{ref'}}\mathcal{L}'
+
J_{\bar{\theta}_l}(x^{pos'})^T
\nabla_{z^{pos'}}\mathcal{L}'
\nonumber\\
&\quad+
\sum_{k=1}^K
J_{\bar{\theta}_l}(x_k^{neg'})^T
\nabla_{z_k^{neg'}}\mathcal{L}'.
\end{align}
Similarly,
\begin{align}
\nabla_{\bar{\theta}_l}\mathcal{L}
&=
J_{\bar{\theta}_l}(x^{ref})^T
\nabla_{z^{ref}}\mathcal{L}
+
J_{\bar{\theta}_l}(x^{pos})^T
\nabla_{z^{pos}}\mathcal{L}
\nonumber\\
&\quad+
\sum_{k=1}^K
J_{\bar{\theta}_l}(x_k^{neg})^T
\nabla_{z_k^{neg}}\mathcal{L}.
\end{align}

Consider first the reference-sample contribution. Adding and
subtracting
\[
J_{\bar{\theta}_l}(x^{ref})^T
\nabla_{z^{ref'}}\mathcal{L}'
\]
gives
\begin{align}
&J_{\bar{\theta}_l}(x^{ref'})^T
\nabla_{z^{ref'}}\mathcal{L}'
-
J_{\bar{\theta}_l}(x^{ref})^T
\nabla_{z^{ref}}\mathcal{L}
\nonumber\\
&=
\left[
J_{\bar{\theta}_l}(x^{ref'})
-
J_{\bar{\theta}_l}(x^{ref})
\right]^T
\nabla_{z^{ref'}}\mathcal{L}'
\nonumber\\
&\quad+
J_{\bar{\theta}_l}(x^{ref})^T
\left[
\nabla_{z^{ref'}}\mathcal{L}'
-
\nabla_{z^{ref}}\mathcal{L}
\right].
\end{align}
Therefore,
\begin{align}
&\left\|
J_{\bar{\theta}_l}(x^{ref'})^T
\nabla_{z^{ref'}}\mathcal{L}'
-
J_{\bar{\theta}_l}(x^{ref})^T
\nabla_{z^{ref}}\mathcal{L}
\right\|_2
\nonumber\\
&\leq
\left\|
J_{\bar{\theta}_l}(x^{ref'})
-
J_{\bar{\theta}_l}(x^{ref})
\right\|_2
\left\|
\nabla_{z^{ref'}}\mathcal{L}'
\right\|_2
\nonumber\\
&\quad+
\left\|
J_{\bar{\theta}_l}(x^{ref})
\right\|_2
\left\|
\nabla_{z^{ref'}}\mathcal{L}'
-
\nabla_{z^{ref}}\mathcal{L}
\right\|_2.
\end{align}

By Corollaries~\ref{col_jac_bound} and
\ref{col_grad_loss_bound},
\begin{align}
\left\|
J_{\bar{\theta}_l}(x^{ref'})
-
J_{\bar{\theta}_l}(x^{ref})
\right\|_2
\left\|
\nabla_{z^{ref'}}\mathcal{L}'
\right\|_2
\leq
2L_\theta(l)G,
\end{align}
where we used
\[
\left\|
J_{\bar{\theta}_l}(x^{ref'})
-
J_{\bar{\theta}_l}(x^{ref})
\right\|_2
\leq
\left\|J_{\bar{\theta}_l}(x^{ref'})\right\|_2
+
\left\|J_{\bar{\theta}_l}(x^{ref})\right\|_2
\leq
2L_\theta(l).
\]

By Lemmas~\ref{Lemma Smooth Loss} and~\ref{input_enc_lip},
\begin{align}
&\left\|
\nabla_{z^{ref'}}\mathcal{L}'
-
\nabla_{z^{ref}}\mathcal{L}
\right\|_2
\nonumber\\
&\leq
\Gamma
\left(
\|z^{ref'}-z^{ref}\|_2^2
+
\|z^{pos'}-z^{pos}\|_2^2
+
\sum_{k=1}^K
\|z_k^{neg'}-z_k^{neg}\|_2^2
\right)^{1/2}
\nonumber\\
&\leq
\Gamma L_x
\left(
\|n^{ref}\|_2^2
+
\|n^{pos}\|_2^2
+
\sum_{k=1}^K
\|n_k^{neg}\|_2^2
\right)^{1/2}.
\end{align}
Since
$\|J_{\bar{\theta}_l}(x^{ref})\|_2\leq L_\theta(l)$,
\begin{align}
&\left\|
J_{\bar{\theta}_l}(x^{ref})
\right\|_2
\left\|
\nabla_{z^{ref'}}\mathcal{L}'
-
\nabla_{z^{ref}}\mathcal{L}
\right\|_2
\nonumber\\
&\leq
L_\theta(l)\Gamma L_x
\left(
\|n^{ref}\|_2^2
+
\|n^{pos}\|_2^2
+
\sum_{k=1}^K
\|n_k^{neg}\|_2^2
\right)^{1/2}.
\end{align}

The same argument applies to the positive sample and to each of the
$K$ negative samples. Hence,
\begin{align}
&\left\|
\nabla_{\bar{\theta}_l}\mathcal{L}'
-
\nabla_{\bar{\theta}_l}\mathcal{L}
\right\|_2
\nonumber\\
&\leq
2(K+2)L_\theta(l)G
\nonumber\\
&\quad+
(K+2)L_\theta(l)\Gamma L_x
\left(
\|n^{ref}\|_2^2
+
\|n^{pos}\|_2^2
+
\sum_{k=1}^K
\|n_k^{neg}\|_2^2
\right)^{1/2}.
\label{eq__raw_grad_diff_norm_1}
\end{align}

By the reverse triangle inequality,
\begin{align}
\left\|
\nabla_{\bar{\theta}_l}\mathcal{L}'
\right\|_2
\leq
\left\|
\nabla_{\bar{\theta}_l}\mathcal{L}
\right\|_2
+
\left\|
\nabla_{\bar{\theta}_l}\mathcal{L}'
-
\nabla_{\bar{\theta}_l}\mathcal{L}
\right\|_2.
\end{align}
Combining this inequality with
\eqref{eq__raw_grad_diff_norm_1} yields
\begin{align}
\left\|
\nabla_{\bar{\theta}_l}\mathcal{L}'
\right\|_2
&\leq
\left\|
\nabla_{\bar{\theta}_l}\mathcal{L}
\right\|_2
+
2(K+2)L_\theta(l)G
\nonumber\\
&\quad+
(K+2)L_\theta(l)\Gamma L_x
\left(
\|n^{ref}\|_2^2
+
\|n^{pos}\|_2^2
+
\sum_{k=1}^K
\|n_k^{neg}\|_2^2
\right)^{1/2}.
\label{eqn_b1_raw}
\end{align}

For independent Gaussian noise with per-coordinate variance
$\gamma^{-1}P$,
\begin{align}
&\mathbb{E}\left[
\|n^{ref}\|_2^2
+
\|n^{pos}\|_2^2
+
\sum_{k=1}^K
\|n_k^{neg}\|_2^2
\right]
\nonumber\\
&=
T(K+2)\gamma^{-1}P.
\end{align}
By Jensen's inequality,
\begin{align}
&\mathbb{E}\left[
\left(
\|n^{ref}\|_2^2
+
\|n^{pos}\|_2^2
+
\sum_{k=1}^K
\|n_k^{neg}\|_2^2
\right)^{1/2}
\right]
\nonumber\\
&\leq
\sqrt{T(K+2)\gamma^{-1}P}.
\end{align}
Taking expectations in \eqref{eqn_b1_raw} therefore gives
\begin{align}
&\mathbb{E}\left[
\left\|
\nabla_{\bar{\theta}_l}\mathcal{L}'
\right\|_2
\right]
\nonumber\\
&\leq
\left\|
\nabla_{\bar{\theta}_l}\mathcal{L}
\right\|_2
+
2(K+2)L_\theta(l)G
\nonumber\\
&\quad+
L_\theta(l)\Gamma L_x
\sqrt{T(K+2)^3\gamma^{-1}P},
\end{align}
which proves statement (1).

To prove statement (2), let
\[
d_l
\triangleq
\nabla_{\bar{\theta}_l}\mathcal{L}'
-
\nabla_{\bar{\theta}_l}\mathcal{L}.
\]
From \eqref{eq__raw_grad_diff_norm_1} and
$(a+b)^2\leq2a^2+2b^2$,
\begin{align}
\|d_l\|_2^2
&\leq
8(K+2)^2L_\theta^2(l)G^2
\nonumber\\
&\quad+
2(K+2)^2L_\theta^2(l)\Gamma^2L_x^2
\left(
\|n^{ref}\|_2^2
+
\|n^{pos}\|_2^2
+
\sum_{k=1}^K
\|n_k^{neg}\|_2^2
\right).
\end{align}
Moreover,
\begin{align}
\left\|
\nabla_{\bar{\theta}_l}\mathcal{L}'
\right\|_2^2
&=
\left\|
\nabla_{\bar{\theta}_l}\mathcal{L}
+d_l
\right\|_2^2
\nonumber\\
&\leq
2\left\|
\nabla_{\bar{\theta}_l}\mathcal{L}
\right\|_2^2
+
2\|d_l\|_2^2.
\end{align}
Taking expectations yields
\begin{align}
&\mathbb{E}\left[
\left\|
\nabla_{\bar{\theta}_l}\mathcal{L}'
\right\|_2^2
\right]
\nonumber\\
&\leq
2\left\|
\nabla_{\bar{\theta}_l}\mathcal{L}
\right\|_2^2
+
16(K+2)^2L_\theta^2(l)G^2
\nonumber\\
&\quad+
4L_\theta^2(l)\Gamma^2L_x^2
T(K+2)^3\gamma^{-1}P,
\end{align}
which proves statement (2).

For the full encoder parameter vector,
\[
\nabla_{\Theta}\mathcal{L}'
=
\begin{bmatrix}
\nabla_{\bar{\theta}_1}\mathcal{L}'\\
\vdots\\
\nabla_{\bar{\theta}_\Psi}\mathcal{L}'
\end{bmatrix},
\qquad
\nabla_{\Theta}\mathcal{L}
=
\begin{bmatrix}
\nabla_{\bar{\theta}_1}\mathcal{L}\\
\vdots\\
\nabla_{\bar{\theta}_\Psi}\mathcal{L}
\end{bmatrix}.
\]
Using
\[
\left\|
\nabla_{\Theta}\mathcal{L}'
-
\nabla_{\Theta}\mathcal{L}
\right\|_2
\leq
\sum_{l=1}^{\Psi}\|d_l\|_2
\]
together with \eqref{eq__raw_grad_diff_norm_1}, we obtain
\begin{align}
&\left\|
\nabla_{\Theta}\mathcal{L}'
\right\|_2
\nonumber\\
&\leq
\left\|
\nabla_{\Theta}\mathcal{L}
\right\|_2
+
2(K+2)G\sum_{l=1}^{\Psi}L_\theta(l)
\nonumber\\
&\quad+
(K+2)\Gamma L_x
\left(\sum_{l=1}^{\Psi}L_\theta(l)\right)
\left(
\|n^{ref}\|_2^2
+
\|n^{pos}\|_2^2
+
\sum_{k=1}^K
\|n_k^{neg}\|_2^2
\right)^{1/2}.
\end{align}
Taking expectations proves statement (3).

Finally, since
\[
\left\|
\nabla_{\Theta}\mathcal{L}'
\right\|_2^2
=
\sum_{l=1}^{\Psi}
\left\|
\nabla_{\bar{\theta}_l}\mathcal{L}'
\right\|_2^2,
\]
summing the layerwise bounds from statement (2) gives
\begin{align}
&\mathbb{E}\left[
\left\|
\nabla_{\Theta}\mathcal{L}'
\right\|_2^2
\right]
\nonumber\\
&\leq
2\left\|
\nabla_{\Theta}\mathcal{L}
\right\|_2^2
+
16(K+2)^2G^2
\sum_{l=1}^{\Psi}L_\theta^2(l)
\nonumber\\
&\quad+
4\Gamma^2L_x^2T(K+2)^3\gamma^{-1}P
\sum_{l=1}^{\Psi}L_\theta^2(l),
\end{align}
which proves statement (4).
\end{proof}

\subsection{Proof of Theorem \ref{fedprox_th4_adapt}}
\label{fproxpf}

\noindent

\begin{reptheorem}{fedprox_th4_adapt}
 Assume that the local losses are $B_{\text{diss}}$-locally dissimilar at $\Theta^t$ {\color{black}{, and that
\[
B_{\mathrm{diss}}\eta_{\mathrm{ie}}<1.
\]}} Then for any SNR $\gamma$, signal power $P$, and heterogeneous client data distributions, we can choose $\mu_{\text{prox}} > L_-$ large enough such that
\begin{align}
\mathcal{L}_{\text{global}}(\Theta^{t+1}) 
&\leq \mathcal{L}_{\text{global}}(\Theta^t)
- \mathcal{O}\left(\frac{1}{\mu_{\text{prox}}}\right)
\|\nabla \mathcal{L}_{\text{global}}(\Theta^t)\|^2_2 \\
&\quad + \mathcal{O}\left(\frac{1}{\mu_{\text{prox}}}\right) \cdot \mathrm{Err}(\gamma, P, B_{\text{diss}}, \eta_{\text{ie}}, \Theta^t),
\end{align}
where the variance term captures the impact of signal noise, client heterogeneity, and inexact updates.
\end{reptheorem}

\noindent \textbf{Proof:}

{\color{black}{We begin by recalling important results from our proofs presented for Theorem \ref{grad_SNR_bound} which we will use to prove this theorem. From our definitions
\begin{align}
\mathcal{L}_c(\Theta)
=
\mathbb{E}_{x\sim\mathcal{D}_c,\,n}
\big[
\mathcal{L}(\Theta;x+n)
\big].
\end{align}
and 
\begin{align}
\mathcal{L}_c^{\mathrm{clean}}(\Theta)
=
\mathbb{E}_{x\sim\mathcal{D}_c}
\big[
\mathcal{L}(\Theta;x)
\big],
\end{align}
and applying Jensen's inequality, we have 
\begin{align}
||\nabla\mathcal{L}_c(\Theta^{t})-\nabla \mathcal{L}^{clean}_c(\Theta^{t}) ||_2&\leq \mathbb{E}_{x\sim\mathcal{D}_c}
\big[
||\mathbb{E}_{n|x}\big[\nabla\mathcal{L}(\Theta^t;x+n)-\nabla\mathcal{L}(\Theta^t;x)
\big]||_2\big].\nonumber \\
&\leq\mathbb{E}_{x\sim\mathcal{D}_c}
\big[
\mathbb{E}_{n|x}\big[||\nabla\mathcal{L}(\Theta^t;x+n)-\nabla\mathcal{L}(\Theta^t;x)||_2
\big]\big] \nonumber \\
&=\mathbb{E}_{x\sim\mathcal{D}_c}
\big[
\mathbb{E}_{n}\big[||\nabla\mathcal{L}(\Theta^t;x+n)-\nabla\mathcal{L}(\Theta^t;x)||_2
\big]\big]\nonumber \\ 
&\leq\mathbb{E}_{x\sim\mathcal{D}_c}\big[2(K+2)G\sum_{l=1}^{\Psi}L_\theta(l)
\nonumber\\
&\quad+
\Gamma L_x
\sqrt{T(K+2)^3\gamma^{-1}P}
\sum_{l=1}^{\Psi}L_\theta(l).\big] \nonumber \\
&=2(K+2)G\sum_{l=1}^{\Psi}L_\theta(l)+
\Gamma L_x
\sqrt{T(K+2)^3\gamma^{-1}P}
\sum_{l=1}^{\Psi}L_\theta(l) \nonumber
\end{align}
where we use the fact that the noise is independent of signal. We now proceed to the main proof.
}}

Introducing $\bar{\mu}_{prox}=\mu_{prox}-L_-$, we seek to prove
\begin{align}
&\mathcal{L}_{global}({\Theta}^{t+1}) \nonumber \\
&\leq \mathcal{L}_{global}(\Theta^{t})-       \bigg(\frac{1}{\mu_{prox}}-\frac{B_{diss}\eta_{ie}}{\mu_{prox}}-\frac{B_{diss}L^{all}_{loss}(1+\eta_{ie})}{\mu_{prox}\bar{\mu}_{prox}}-\frac{3B_{diss}^2L^{all}_{loss}(1+\eta_{ie})^2}{2 \bar{\mu}^2_{prox}}\bigg) ||\nabla\mathcal{L}_{global}(\Theta^{t})||^2_2 \nonumber \\
&+\bigg(2(K+2)G\bigg(\frac{L^{all}_{loss}(1+\eta_{ie})}{\mu_{prox}\bar{\mu}_{prox}}\bigg) \sum_{l=1}^{\Psi} L_{\theta}(l) + \Gamma L_x \sqrt{T(K+2)^3\gamma^{-1}P}\bigg(\frac{L^{all}_{loss}(1+\eta_{ie})}{\mu_{prox}\bar{\mu}_{prox}}\bigg)\sum_{l=1}^{\Psi} L_{\theta}(l)\bigg) ||\nabla\mathcal{L}_{global}(\Theta^{t})||_2  \nonumber \\
&+\bigg(\frac{12(K+2)^2G^2L^{all}_{loss}(1+\eta_{ie})^2}{\bar{\mu}_{prox}^2} (\sum_{l=1}^{\Psi} L_{\theta}(l))^2 +  \frac{3 \Gamma^2 L_x^2 T(K+2)^3\gamma^{-1}P(1+\eta_{ie})^2}{2\bar{\mu}_{prox}^2}(\sum_{l=1}^{\Psi} L_{\theta}(l))^2 \bigg)+ \nonumber \\
&\bigg(2(K+2)G\bigg(\frac{1+\eta_{ie}}{\mu_{prox}}\bigg) \sum_{l=1}^{\Psi} L_{\theta}(l) + \Gamma L_x \sqrt{T(K+2)^3\gamma^{-1}P}\bigg(\frac{1+\eta_{ie}}{\mu_{prox}}\bigg)\sum_{l=1}^{\Psi} L_{\theta}(l)\bigg) ||\nabla\mathcal{L}_{global}(\Theta^{t})||_2 
\end{align}
Let us define 
\begin{align}
\color{black} e_c^{t+1}=\nabla \mathcal{L}_{c}(\Theta_c^{t+1})+\mu_{prox}(\Theta_c^{t+1}-\Theta^t).
\end{align}
By definition of inexactness and Assumption \ref{inexact_ass}, we obtain
\begin{align}
|| e_{c}^{t+1}|| \leq \eta_{ie}||\nabla \mathcal{L}_{c}(\Theta^{t})||
\end{align}
Taking expectation, we get 
\begin{align}
\bar{\Theta}^{t+1}=\mathbb{E}_c[\Theta_c^{t+1}]
\end{align} 
which yields 
\begin{align}
\bar{\Theta}^{t+1}-\Theta^t=-\frac{1}{\mu_{prox}}\mathbb{E}_c[\nabla \mathcal{L}_{c}(\Theta_c^{t+1})]+\frac{1}{\mu_{prox}}\mathbb{E}_c[e_c^{t+1}] \label{eq_grad_and_error}
\end{align}
Recall that we have defined $\bar{\mu}_{prox}=\mu_{prox}-L_{-}$. By the optimality condition
\begin{align}
\color{black}\nabla h_c(\hat{\Theta}_c^{t+1}, \Theta^{t})=0
\end{align}
and by $\bar{\mu}_{prox}$-strong convexity of $h_c()$, for any $\Theta$
\begin{align}
\color{black} h_c(\Theta, \Theta^{t})\geq h_c(\hat{\Theta}_c^{t+1}, \Theta^{t})+\frac{\bar{\mu}_{prox}}{2}||\Theta-\hat{\Theta}_c^{t+1}||_2^2
\end{align}
Strong convexity also implies 
\begin{align}
||\hat{\Theta}_c^{t+1} -\Theta_c^{t+1}||_2&\leq \frac{\eta_{ie}}{\bar{\mu}_{prox}}||\nabla \mathcal{L}_{c}(\Theta^t) ||_2\\
||\hat{\Theta}_c^{t+1} -\Theta^{t}||_2&\leq \frac{1}{\bar{\mu}_{prox}}||\nabla \mathcal{L}_{c}(\Theta^t) ||_2
\end{align}
Consequently,
\begin{align}
||\Theta_c^{t+1}-\Theta^{t}||_2&=||\Theta_c^{t+1}-\hat{\Theta}_c^{t+1}+\hat{\Theta}_c^{t+1}-\Theta^{t}||_2 \nonumber \\ & \leq||\Theta_c^{t+1}-\hat{\Theta}_c^{t+1}||_2+||\hat{\Theta}_c^{t+1}-\Theta^{t}||_2 \nonumber \\
&\leq \frac{1+\eta_{ie}}{\bar{\mu}_{prox}} ||\nabla \mathcal{L}_{c}(\Theta^t) ||_2 \nonumber \\
&\leq \frac{1+\eta_{ie}}{\bar{\mu}_{prox}} \bigg( ||\nabla\mathcal{L}^{clean}_c(\Theta^t)||_2+ 2(K+2)G \sum_{l=1}^{\Psi} L_{\theta}(l) +  \Gamma L_x \sqrt{T(K+2)^3\gamma^{-1}P}\sum_{l=1}^{\Psi} L_{\theta}(l)\bigg) \nonumber \\
& = \frac{1+\eta_{ie}}{\bar{\mu}_{prox}}||\nabla\mathcal{L}^{clean}_c(\Theta^t)||_2+ 2(K+2)G\frac{1+\eta_{ie}}{\bar{\mu}_{prox}} \sum_{l=1}^{\Psi} L_{\theta}(l) +  \Gamma L_x \sqrt{T(K+2)^3\gamma^{-1}P}\frac{1+\eta_{ie}}{\bar{\mu}_{prox}}\sum_{l=1}^{\Psi} L_{\theta}(l)
\end{align}
where the third inequality is obtained by utilizing theorem \ref{grad_SNR_bound}.
By Jensen's inequality, 
\begin{align}
||\bar{\Theta}^{t+1}-\Theta^t||_2 &\leq \mathbb{E}_c\bigg[||\Theta_c^{t+1}-\Theta^{t}||_2\bigg] \nonumber \\
  & \leq \frac{1+\eta_{ie}}{\bar{\mu}_{prox}}\mathbb{E}_c\bigg[||\nabla\mathcal{L}^{clean}_c(\Theta^t)||_2\bigg]+ 2(K+2)G\frac{1+\eta_{ie}}{\bar{\mu}_{prox}} \sum_{l=1}^{\Psi} L_{\theta}(l) +  \Gamma L_x \sqrt{T(K+2)^3\gamma^{-1}P}\frac{1+\eta_{ie}}{\bar{\mu}_{prox}}\sum_{l=1}^{\Psi} L_{\theta}(l)
\end{align}
Since the square root is a concave function, invoking Jensen's inequality again followed by $B_{diss}$-dissimilarity at $\Theta^t$  yields
\begin{align}
\mathbb{E}_c\bigg[||\nabla\mathcal{L}^{clean}_c(\Theta^t)||_2\bigg] &\leq \sqrt{\mathbb{E}_c[||\nabla\mathcal{L}^{clean}_c(\Theta^t)||^2_2]} \nonumber \\
&\leq B_{diss}||\nabla \mathcal{L}_{global}(\Theta^t)||_2
\end{align}
Consequently, 
\begin{align}
&||\bar{\Theta}^{t+1}-\Theta^t||_2 \leq \frac{1+\eta_{ie}}{\bar{\mu}_{prox}}B_{diss}||\nabla \mathcal{L}_{global}(\Theta^t)||_2+ 2(K+2)G\frac{1+\eta_{ie}}{\bar{\mu}_{prox}} \sum_{l=1}^{\Psi} L_{\theta}(l) \nonumber \\
&+  \Gamma L_x \sqrt{T(K+2)^3\gamma^{-1}P}\frac{1+\eta_{ie}}{\bar{\mu}_{prox}}\sum_{l=1}^{\Psi} L_{\theta}(l)
\end{align}
Next, define
\begin{align}
\color{black} M_{gap}^{t+1}=\nabla\mathcal{L}_{global}(\Theta^t)+\mu_{prox} (\bar{\Theta}^{t+1}-\Theta^t) 
\end{align}
so that 
\begin{align}
 &\color{black} ||M_{gap}^{t+1}||_2=||\mathbb{E}_c\big[ -\nabla \mathcal{L}_c(\Theta_c^{t+1})+\nabla \mathcal{L}^{clean}_c(\Theta^{t}) + e_c^{t+1}\big] ||_2 \nonumber \\
&\color{black} \leq \mathbb{E}_c\big[ ||-\nabla \mathcal{L}_c(\Theta_c^{t+1})+\nabla \mathcal{L}^{clean}_c(\Theta^{t}) ||_2 +\eta_{ie}||\nabla\mathcal{L}_{c}(\Theta^{t})||_2 \big] \label{gap_norm} 
\end{align}

Now, 
\begin{align}
&||\nabla \mathcal{L}_c(\Theta_c^{t+1})-\nabla \mathcal{L}^{clean}_c(\Theta^{t}) ||_2 \nonumber \\
&=||\nabla \mathcal{L}_c(\Theta_c^{t+1})-\nabla\mathcal{L}_c(\Theta^{t})+\nabla\mathcal{L}_c(\Theta^{t})-\nabla \mathcal{L}^{clean}_c(\Theta^{t}) ||_2\nonumber\\
&\leq ||\nabla \mathcal{L}_c(\Theta_c^{t+1})-\nabla\mathcal{L}_c(\Theta^{t})||_2+||\nabla\mathcal{L}_c(\Theta^{t})-\nabla \mathcal{L}^{clean}_c(\Theta^{t}) ||_2 \nonumber \\
&\leq L^{all}_{loss} ||\Theta_c^{t+1}-\Theta^t||_2+ 2(K+2)G\sum_{l=1}^{\Psi}L_{\theta}(l) +\Gamma L_x  \sqrt{T(K+2)^3\gamma^{-1}P}\sum_{l=1}^{\Psi} L_{\theta}(l) \nonumber \\
&\leq L^{all}_{loss} ||\Theta_c^{t+1}-\Theta^t||_2+ 2(K+2)G\sum_{l=1}^{\Psi}L_{\theta}(l) +\Gamma L_x  \sqrt{T(K+2)^3\gamma^{-1}P}\sum_{l=1}^{\Psi} L_{\theta}(l)
\end{align}
Substituting this back in \eqref{gap_norm} yields
\begin{align}
&||M_{gap}^{t+1}||_2\leq \big(\frac{L^{all}_{loss}(1+\eta_{ie})}{\bar{\mu}_{prox}}+\eta_{ie}\big)\mathbb{E}_c\bigg[||\nabla\mathcal{L}^{clean}_c(\Theta^t)||_2\bigg] + 2(K+2)G\bigg(\frac{L^{all}_{loss}(1+\eta_{ie})}{\bar{\mu}_{prox}}+1+\eta_{ie}\bigg) \sum_{l=1}^{\Psi} L_{\theta}(l)  \nonumber \\ &+\Gamma L_x \sqrt{T(K+2)^3\gamma^{-1}P}\bigg(\frac{L^{all}_{loss}(1+\eta_{ie})}{\bar{\mu}_{prox}}+1+\eta_{ie}\bigg)\sum_{l=1}^{\Psi} L_{\theta}(l) \nonumber \\
&\leq B_{diss}\big(\frac{L^{all}_{loss}(1+\eta_{ie})}{\bar{\mu}_{prox}}+\eta_{ie}\big)||\nabla \mathcal{L}_{global}(\Theta^t)||_2 + 2(K+2)G\bigg(\frac{L^{all}_{loss}(1+\eta_{ie})}{\bar{\mu}_{prox}}+1+\eta_{ie}\bigg) \sum_{l=1}^{\Psi} L_{\theta}(l)  \nonumber \\ &+\Gamma L_x \sqrt{T(K+2)^3\gamma^{-1}P}\bigg(\frac{L^{all}_{loss}(1+\eta_{ie})}{\bar{\mu}_{prox}}+1+\eta_{ie}\bigg)\sum_{l=1}^{\Psi} L_{\theta}(l)
\end{align}
Finally, the $L^{all}_{loss}$ smoothness of the global loss function and first order Taylor expansion,
\begin{align}
&\mathcal{L}_{global}(\bar{\Theta}^{t+1})\leq \mathcal{L}_{global}(\Theta^{t}) + <\nabla\mathcal{L}_{global}(\Theta^{t}),\bar{\Theta}^{t+1} - \Theta^{t} > +\frac{L^{all}_{loss}}{2}||\bar{\Theta}^{t+1}-\Theta^t||^2_2 \nonumber \\
&\color{black}=\mathcal{L}_{global}(\Theta^{t}) -\frac{1}{\mu_{prox}} ||\nabla \mathcal{L}_{global}(\Theta^t)||^2_2 +\frac{1}{\mu_{prox}} <\nabla \mathcal{L}_{global}(\Theta^t),M^{t+1}_{gap}> +\frac{L^{all}_{loss}}{2}||\bar{\Theta}^{t+1}-\Theta^t||^2_2 \nonumber \\
&\color{black}\leq \mathcal{L}_{global}(\Theta^{t}) -\frac{1}{\mu_{prox}} ||\nabla \mathcal{L}_{global}(\Theta^t)||^2_2 +\frac{1}{\mu_{prox}} ||\nabla\mathcal{L}_{global}(\Theta^t)||_2||M^{t+1}_{gap}||_2 +\frac{L^{all}_{loss}}{2}||\bar{\Theta}^{t+1}-\Theta^t||^2_2 \nonumber \\
&\leq \mathcal{L}_{global}(\Theta^{t})-       \bigg(\frac{1}{\mu_{prox}}-\frac{B_{diss}\eta_{ie}}{\mu_{prox}}-\frac{B_{diss}L^{all}_{loss}(1+\eta_{ie})}{\mu_{prox}\bar{\mu}_{prox}}-\frac{3B_{diss}^2L^{all}_{loss}(1+\eta_{ie})^2}{2 \bar{\mu}^2_{prox}}\bigg) ||\nabla\mathcal{L}_{global}(\Theta^{t})||^2_2 \nonumber \\
&+\bigg(2(K+2)G\bigg(\frac{L^{all}_{loss}(1+\eta_{ie})}{\mu_{prox}\bar{\mu}_{prox}}\bigg) \sum_{l=1}^{\Psi} L_{\theta}(l) + \Gamma L_x \sqrt{T(K+2)^3\gamma^{-1}P}\bigg(\frac{L^{all}_{loss}(1+\eta_{ie})}{\mu_{prox}\bar{\mu}_{prox}}\bigg)\sum_{l=1}^{\Psi} L_{\theta}(l)\bigg) ||\nabla\mathcal{L}_{global}(\Theta^{t})||_2  \nonumber \\
&+\bigg(\frac{12(K+2)^2G^2L^{all}_{loss}(1+\eta_{ie})^2}{\bar{\mu}_{prox}^2} (\sum_{l=1}^{\Psi} L_{\theta}(l))^2 +  \frac{3 \Gamma^2 L_x^2 T(K+2)^3\gamma^{-1}P(1+\eta_{ie})^2}{2\bar{\mu}_{prox}^2}(\sum_{l=1}^{\Psi} L_{\theta}(l))^2 \bigg)+ \nonumber \\
&\bigg(2(K+2)G\bigg(\frac{1+\eta_{ie}}{\mu_{prox}}\bigg) \sum_{l=1}^{\Psi} L_{\theta}(l) + \Gamma L_x \sqrt{T(K+2)^3\gamma^{-1}P}\bigg(\frac{1+\eta_{ie}}{\mu_{prox}}\bigg)\sum_{l=1}^{\Psi} L_{\theta}(l)\bigg) ||\nabla\mathcal{L}_{global}(\Theta^{t})||_2 
\end{align}
Finally, since we have complete client participation, we have $\bar{\Theta}^{t+1}=\Theta^{t+1}$. Plugging this into the inequality above yields the desired result. 
\hfill $\blacksquare$s

\subsection{Proofs of Theorems \ref{grad_SNR_bound_softplus} and \ref{fedprox_th4_adapt_softplus}}

\noindent
\begin{reptheorem}{grad_SNR_bound_softplus}
\label{softplusactrespf}
Suppose that the encoder uses Softplus, or another activation function
satisfying the conditions of Lemma~\ref{jac_liptz}. Let
$z^{ref'}$, $z^{pos'}$, and $\{z_k^{neg'}\}_{k=1}^K$ denote the
noise-perturbed embeddings, and let $z^{ref}$, $z^{pos}$, and
$\{z_k^{neg}\}_{k=1}^K$ denote their noiseless counterparts. Then the
excess terms in the expected gradient-norm bounds scale as
$\mathcal{O}(\gamma^{-1/2}P^{1/2})$, while those in the expected
squared-gradient-norm bounds scale as
$\mathcal{O}(\gamma^{-1}P)$. Specifically,
\begin{align}
&\mathbb{E}\left[
\left\|
\nabla_{\bar{\theta}_l}
\mathcal{L}(z^{ref'},z^{pos'},\{z_k^{neg'}\}_{k=1}^K)
\right\|_2
\right]
\nonumber\\
&\leq
\left\|
\nabla_{\bar{\theta}_l}
\mathcal{L}(z^{ref},z^{pos},\{z_k^{neg}\}_{k=1}^K)
\right\|_2
+
(K+2)\Gamma_JG\sqrt{T\gamma^{-1}P}
\nonumber\\
&\quad+
L_\theta(l)\Gamma L_x
\sqrt{T(K+2)^3\gamma^{-1}P},
\end{align}
\begin{align}
&\mathbb{E}\left[
\left\|
\nabla_{\bar{\theta}_l}
\mathcal{L}(z^{ref'},z^{pos'},\{z_k^{neg'}\}_{k=1}^K)
\right\|_2^2
\right]
\nonumber\\
&\leq
2\left\|
\nabla_{\bar{\theta}_l}
\mathcal{L}(z^{ref},z^{pos},\{z_k^{neg}\}_{k=1}^K)
\right\|_2^2
+
4T\gamma^{-1}P(K+2)^2\Gamma_J^2G^2
\nonumber\\
&\quad+
4L_\theta^2(l)\Gamma^2L_x^2
T(K+2)^3\gamma^{-1}P,
\end{align}
\begin{align}
&\mathbb{E}\left[
\left\|
\nabla_{\Theta}
\mathcal{L}(z^{ref'},z^{pos'},\{z_k^{neg'}\}_{k=1}^K)
\right\|_2
\right]
\nonumber\\
&\leq
\left\|
\nabla_{\Theta}
\mathcal{L}(z^{ref},z^{pos},\{z_k^{neg}\}_{k=1}^K)
\right\|_2
+
(K+2)\Gamma_JG\Psi\sqrt{T\gamma^{-1}P}
\nonumber\\
&\quad+
\Gamma L_x
\sqrt{T(K+2)^3\gamma^{-1}P}
\sum_{l=1}^{\Psi}L_\theta(l),
\end{align}
\begin{align}
&\mathbb{E}\left[
\left\|
\nabla_{\Theta}
\mathcal{L}(z^{ref'},z^{pos'},\{z_k^{neg'}\}_{k=1}^K)
\right\|_2^2
\right]
\nonumber\\
&\leq
2\left\|
\nabla_{\Theta}
\mathcal{L}(z^{ref},z^{pos},\{z_k^{neg}\}_{k=1}^K)
\right\|_2^2
+
4T\gamma^{-1}P(K+2)^2\Psi\Gamma_J^2G^2
\nonumber\\
&\quad+
4\Gamma^2L_x^2T(K+2)^3\gamma^{-1}P
\sum_{l=1}^{\Psi}L_\theta^2(l).
\end{align}
\end{reptheorem}

\begin{proof}
Define
\[
\mathcal{L}'
\triangleq
\mathcal{L}(z^{ref'},z^{pos'},\{z_k^{neg'}\}_{k=1}^K),
\qquad
\mathcal{L}
\triangleq
\mathcal{L}(z^{ref},z^{pos},\{z_k^{neg}\}_{k=1}^K).
\]
Consider the reference-sample contribution to the layer-$l$ gradient.
Adding and subtracting
$J_{\bar{\theta}_l}(x^{ref})^T\nabla_{z^{ref'}}\mathcal{L}'$
gives
\begin{align}
&J_{\bar{\theta}_l}(x^{ref'})^T\nabla_{z^{ref'}}\mathcal{L}'
-
J_{\bar{\theta}_l}(x^{ref})^T\nabla_{z^{ref}}\mathcal{L}
\nonumber\\
&=
\left[
J_{\bar{\theta}_l}(x^{ref'})
-
J_{\bar{\theta}_l}(x^{ref})
\right]^T
\nabla_{z^{ref'}}\mathcal{L}'
\nonumber\\
&\quad+
J_{\bar{\theta}_l}(x^{ref})^T
\left[
\nabla_{z^{ref'}}\mathcal{L}'
-
\nabla_{z^{ref}}\mathcal{L}
\right].
\end{align}
By Lemma~\ref{jac_liptz} and
Corollary~\ref{col_grad_loss_bound},
\begin{align}
&\left\|
\left[
J_{\bar{\theta}_l}(x^{ref'})
-
J_{\bar{\theta}_l}(x^{ref})
\right]^T
\nabla_{z^{ref'}}\mathcal{L}'
\right\|_2
\nonumber\\
&\leq
\Gamma_JG
\|x^{ref'}-x^{ref}\|_2
=
\Gamma_JG\|n^{ref}\|_2.
\end{align}
By Corollary~\ref{col_jac_bound},
Lemma~\ref{Lemma Smooth Loss}, and
Lemma~\ref{input_enc_lip},
\begin{align}
&\left\|
J_{\bar{\theta}_l}(x^{ref})^T
\left[
\nabla_{z^{ref'}}\mathcal{L}'
-
\nabla_{z^{ref}}\mathcal{L}
\right]
\right\|_2
\nonumber\\
&\leq
L_\theta(l)\Gamma L_x
\left(
\|n^{ref}\|_2^2
+
\|n^{pos}\|_2^2
+
\sum_{k=1}^K\|n_k^{neg}\|_2^2
\right)^{1/2}.
\end{align}
Applying the same argument to the positive sample and all $K$ negative
samples yields
\begin{align}
&\left\|
\nabla_{\bar{\theta}_l}\mathcal{L}'
-
\nabla_{\bar{\theta}_l}\mathcal{L}
\right\|_2
\nonumber\\
&\leq
\Gamma_JG
\left(
\|n^{ref}\|_2+\|n^{pos}\|_2
+\sum_{k=1}^K\|n_k^{neg}\|_2
\right)
\nonumber\\
&\quad+
(K+2)L_\theta(l)\Gamma L_x
\left(
\|n^{ref}\|_2^2
+
\|n^{pos}\|_2^2
+
\sum_{k=1}^K\|n_k^{neg}\|_2^2
\right)^{1/2}.
\label{eq:softplus_grad_diff}
\end{align}

For independent Gaussian noise with per-coordinate variance
$\gamma^{-1}P$,
\[
\mathbb{E}\|n\|_2
\leq
\sqrt{\mathbb{E}\|n\|_2^2}
=
\sqrt{T\gamma^{-1}P},
\]
and
\[
\mathbb{E}\left[
\|n^{ref}\|_2^2+\|n^{pos}\|_2^2+
\sum_{k=1}^K\|n_k^{neg}\|_2^2
\right]
=
T(K+2)\gamma^{-1}P.
\]
Taking expectations in \eqref{eq:softplus_grad_diff} and applying the
reverse triangle inequality proves the first bound.

For the squared-gradient bound, let
\[
d_l
=
\nabla_{\bar{\theta}_l}\mathcal{L}'
-
\nabla_{\bar{\theta}_l}\mathcal{L}.
\]
By Cauchy--Schwarz,
\[
\left(
\|n^{ref}\|_2+\|n^{pos}\|_2+
\sum_{k=1}^K\|n_k^{neg}\|_2
\right)^2
\leq
(K+2)
\left(
\|n^{ref}\|_2^2+\|n^{pos}\|_2^2+
\sum_{k=1}^K\|n_k^{neg}\|_2^2
\right).
\]
Using \((a+b)^2\leq2a^2+2b^2\) in
\eqref{eq:softplus_grad_diff}, followed by
\[
\|\nabla_{\bar{\theta}_l}\mathcal{L}'\|_2^2
\leq
2\|\nabla_{\bar{\theta}_l}\mathcal{L}\|_2^2
+
2\|d_l\|_2^2,
\]
gives the second bound after taking expectations.

The full-parameter gradient is obtained by stacking the layerwise
gradients. Summing the first-order layerwise estimates gives the third
bound, while summing the corrected squared layerwise estimates gives
the fourth bound.
\end{proof}

\begin{reptheorem}{fedprox_th4_adapt_softplus}
Assume that the clean local objectives are
$B_{\mathrm{diss}}$-locally dissimilar at $\Theta^t$, and that
$B_{\mathrm{diss}}\eta_{\mathrm{ie}}<1$. Then, for any signal power
$P$ and heterogeneous client data distributions, there exist a
sufficiently large proximal coefficient
$\mu_{\mathrm{prox}}>L_-$ and a sufficiently large SNR $\gamma$ such
that
\begin{align}
\mathcal{L}_{\mathrm{global}}(\Theta^{t+1})
&\leq
\mathcal{L}_{\mathrm{global}}(\Theta^t)
-
\mathcal{O}\left(\frac{1}{\mu_{\mathrm{prox}}}\right)
\|\nabla \mathcal{L}_{\mathrm{global}}(\Theta^t)\|_2^2
\nonumber\\
&\quad+
\mathcal{O}\left(\frac{1}{\mu_{\mathrm{prox}}}\right)
\cdot
\mathcal{O}\left(\gamma^{-1/2}P^{1/2}\right)
\|\nabla \mathcal{L}_{\mathrm{global}}(\Theta^t)\|_2
\nonumber\\
&\quad+
\mathcal{O}\left(\frac{1}{\mu_{\mathrm{prox}}}\right)
\cdot
\mathcal{O}\left(\gamma^{-1}P\right).
\end{align}
The hidden constants depend on the client heterogeneity and inexactness
level, while the explicit SNR-dependent terms quantify the effect of
signal noise.
    
\end{reptheorem}

\noindent\textbf{Proof: } Starting from Theorem \ref{grad_SNR_bound_softplus}, we re-trace the steps of Theorem \ref{fedprox_th4_adapt}, making appropriate substitutions for the expected norm of gradients according to the bounds provided by Theorem \ref{grad_SNR_bound_softplus} and show
\begin{align}
&\mathcal{L}_{global}({\Theta}^{t+1}) \leq \mathcal{L}_{global}(\Theta^{t})-       ||\nabla\mathcal{L}_{global}(\Theta^{t})||^2_2\bigg(\frac{1}{\mu_{prox}}\nonumber \\&-\frac{B_{diss}\eta_{ie}}{\mu_{prox}}-\frac{B_{diss}L^{all}_{loss}(1+\eta_{ie})}{\mu_{prox}\bar{\mu}_{prox}}-\frac{3B_{diss}^2L^{all}_{loss}(1+\eta_{ie})^2}{2 \bar{\mu}^2_{prox}}\bigg)  \nonumber \\
&+||\nabla\mathcal{L}_{global}(\Theta^{t})||_2  \bigg(\sqrt{T\gamma^{-1}P}(K+2)\Gamma_{J}G\Psi\frac{L^{all}_{loss}(1+\eta_{ie})}{\mu_{prox}\bar{\mu}_{prox}}  \nonumber \\&+ \Gamma L_x \sqrt{T(K+2)^3\gamma^{-1}P}\frac{L^{all}_{loss}(1+\eta_{ie})}{\mu_{prox}\bar{\mu}_{prox}}\sum_{l=1}^{\Psi} L_{\theta}(l)\bigg) \nonumber \\
&+\bigg(\frac{3T\gamma^{-1}P(K+2)^2\Gamma_{J}^2G^2\Psi^2L^{all}_{loss}(1+\eta_{ie})^2}{\bar{\mu}_{prox}^2}  \nonumber \\&+  \frac{3 \Gamma^2 L_x^2 T(K+2)^3\gamma^{-1}P(1+\eta_{ie})^2}{2\bar{\mu}_{prox}^2}(\sum_{l=1}^{\Psi} L_{\theta}(l))^2 \bigg) \nonumber \\
&+||\nabla\mathcal{L}_{global}(\Theta^{t})||_2\bigg(\sqrt{T\gamma^{-1}P}\Psi(K+2)\Gamma_{J}G\bigg(\frac{1+\eta_{ie}}{\mu_{prox}}\bigg)  \nonumber \\&+ \Gamma L_x \sqrt{T(K+2)^3\gamma^{-1}P}\bigg(\frac{1+\eta_{ie}}{\mu_{prox}}\bigg)\sum_{l=1}^{\Psi} L_{\theta}(l)\bigg) 
\end{align}
The remaining steps and details are identical to those utilized in the proof of Theorem \ref{fedprox_th4_adapt}, and are therefore omitted.
\hfill $\blacksquare$

\section{Miscellaneous}
\label{misc}

\subsection{Mobility-induced carrier frequency offset}

This subsection provides a simplified relation between carrier frequency offset (CFO) and relative motion between a transmitter and receiver. Such mobility-induced offsets may vary across clients and thereby introduce heterogeneity into federated representation learning.

Let $f_c$ denote the nominal carrier frequency to which the receiver's local oscillator is tuned, and let $\nu_r$ denote the radial velocity of the transmitter relative to the receiver. Under the classical Doppler model for a moving transmitter and stationary receiver, the observed carrier frequency is
\begin{align}
f_o
=
f_c\left(1\mp\frac{\nu_r}{c}\right)^{-1},
\end{align}
where $c$ is the speed of light. The minus sign corresponds to motion toward the receiver and the plus sign to motion away from it. For $\nu_r\ll c$, a first-order approximation gives
\begin{align}
f_o
\approx
f_c\left(1\pm\frac{\nu_r}{c}\right),
\end{align}
where the positive sign corresponds to an approaching transmitter.

Consider the approaching case, for which the Doppler shift is positive. The received passband signal can be represented as
\[
\tilde{A}(t)
\cos\left(2\pi f_o t+\tilde{\theta}(t)\right),
\]
where $\tilde{A}(t)$ and $\tilde{\theta}(t)$ represent the amplitude and phase modulation, respectively. After downconversion using a local oscillator at frequency $f_c$ and sampling at rate $f_s$, the corresponding complex baseband signal is
\begin{align}
r[n]
=
\tilde{A}[n]
e^{j\left(
2\pi\frac{f_c\nu_r}{f_sc}n+\tilde{\theta}[n]
\right)}
+
w[n],
\end{align}
where $w[n]$ denotes complex baseband noise. The resulting normalized CFO is therefore
\begin{align}
\Delta f
=
\frac{f_c\nu_r}{f_sc}.
\end{align}
This offset produces a phase rotation that accumulates linearly with the sample index. Consequently, differences in carrier frequency, sampling rate, and radial velocity across clients can produce different values of $\Delta f$, introducing client heterogeneity during federated encoder training. 

\subsection{Effect of varying $\bar{\alpha}$ on label distributions}
\label{DirichletDistsSec}

Figures~\ref{fig:dist_migou_dirichlet_part1}
and~\ref{fig:dist_migou_dirichlet_part2} show the client-wise label
distributions obtained by Dirichlet partitioning of the MIGOU dataset
across four clients for different values of $\bar{\alpha}$. Smaller
values of $\bar{\alpha}$ produce more highly skewed client
distributions, whereas larger values yield increasingly uniform
partitions.

\begin{figure}[t]
    \centering
    
    \begin{subfigure}[b]{0.32\columnwidth}
        \centering
        \includegraphics[width=\linewidth]{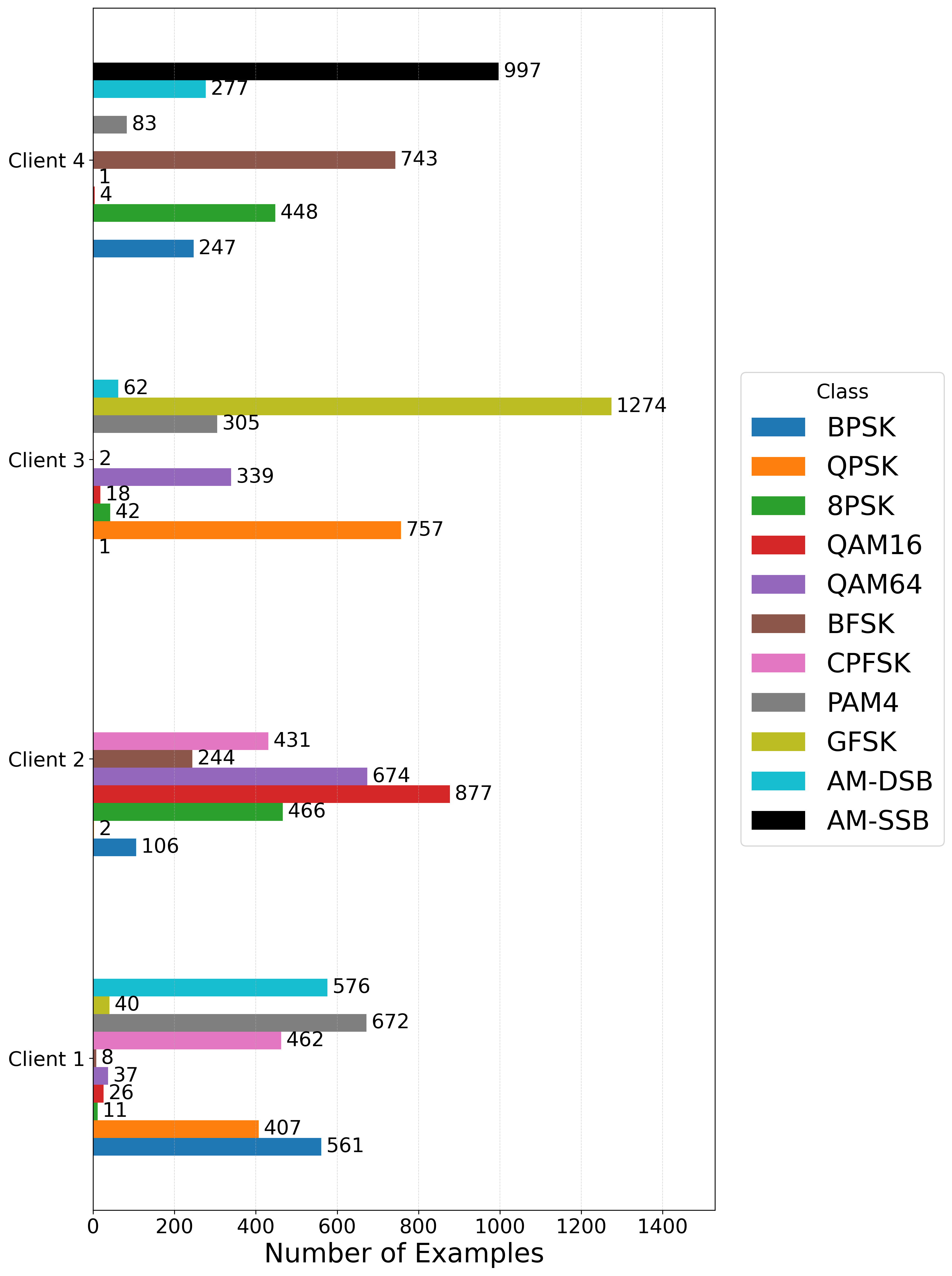}
        \caption{$\bar{\alpha}=0.25$}
    \end{subfigure}
    \hfill
    \begin{subfigure}[b]{0.32\columnwidth}
        \centering
        \includegraphics[width=\linewidth]{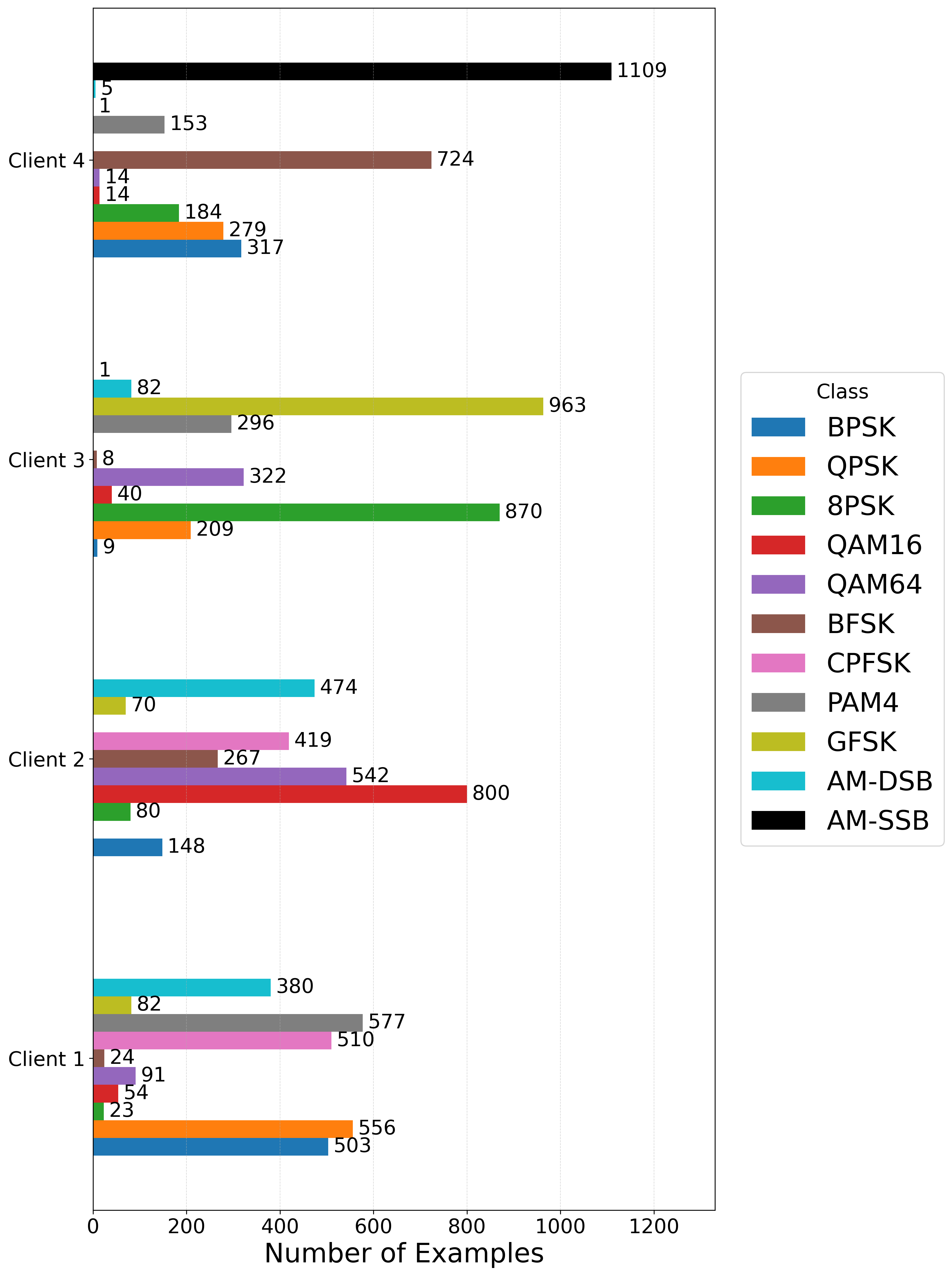}
        \caption{$\bar{\alpha}=0.375$}
    \end{subfigure}
    \hfill
    \begin{subfigure}[b]{0.32\columnwidth}
        \centering
        \includegraphics[width=\linewidth]{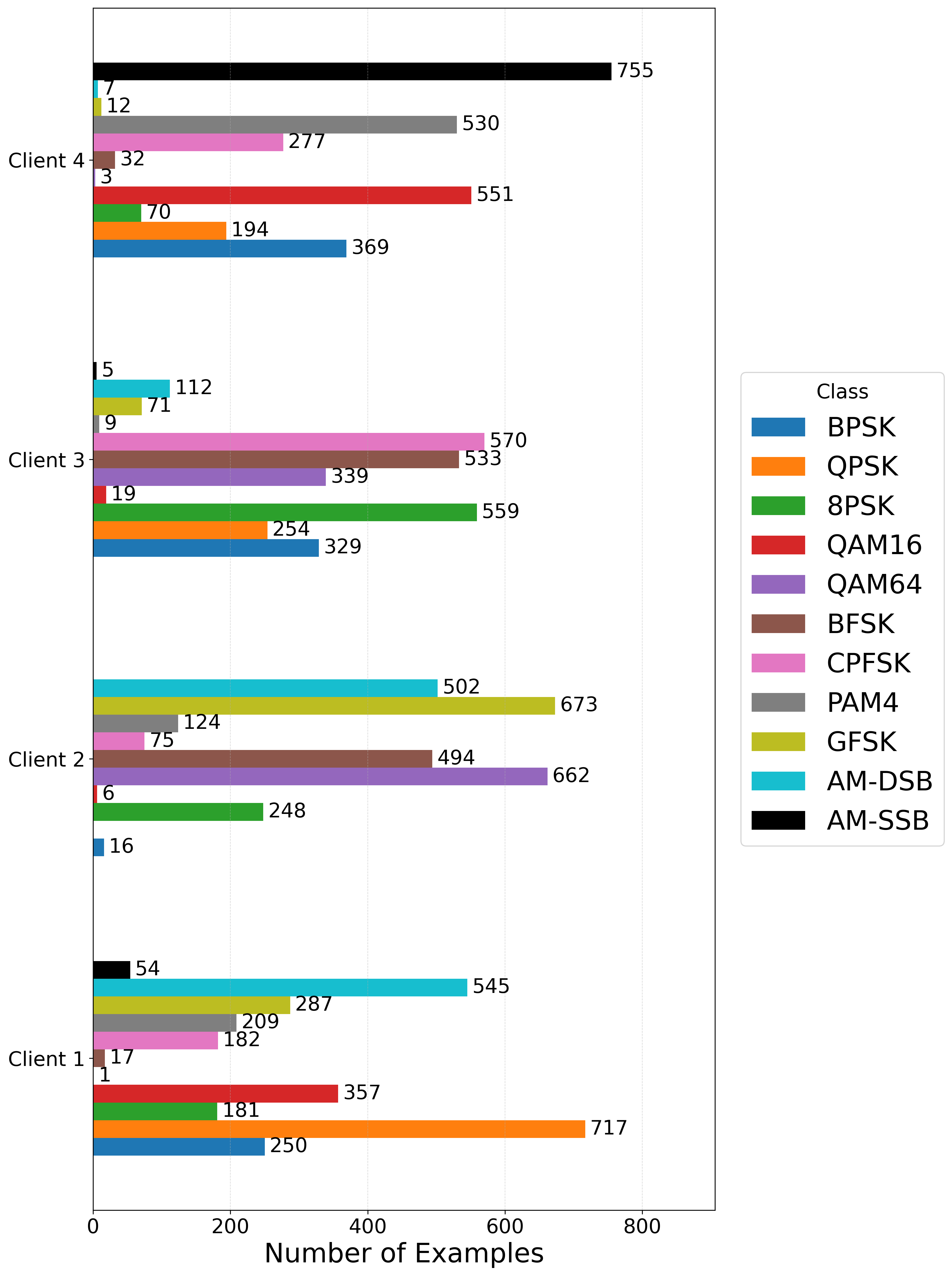}
        \caption{$\bar{\alpha}=0.5$}
    \end{subfigure}

    \caption{Client-wise label distributions in the MIGOU dataset under Dirichlet partitioning for $\bar{\alpha}\in\{0.25,0.375,0.5\}$.}
    \label{fig:dist_migou_dirichlet_part1}
\end{figure}

\begin{figure}[t]
    \centering
    
    \begin{subfigure}[b]{0.32\columnwidth}
        \centering
        \includegraphics[width=\linewidth]{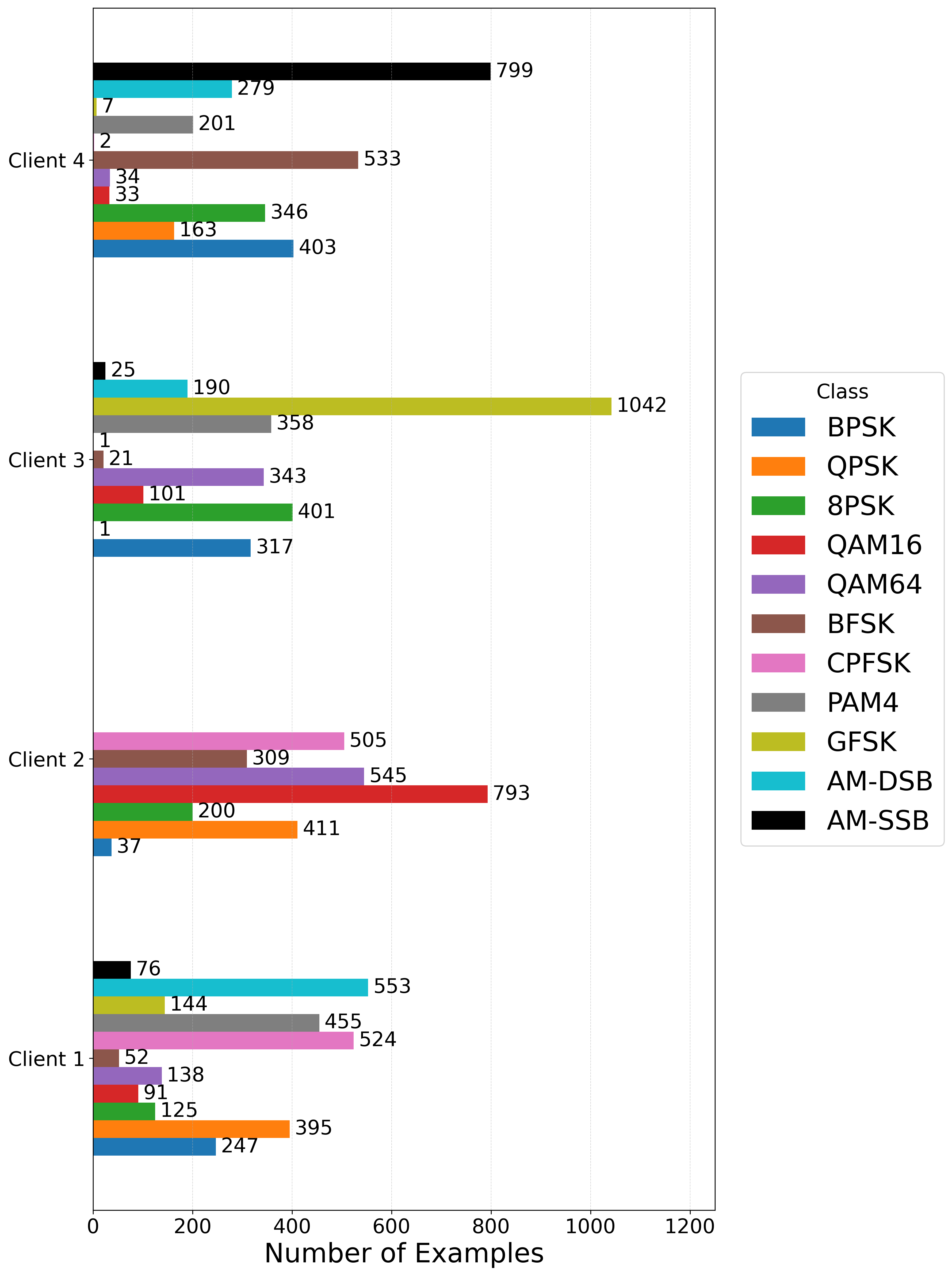}
        \caption{$\bar{\alpha}=0.625$}
    \end{subfigure}
    \hfill
    \begin{subfigure}[b]{0.32\columnwidth}
        \centering
        \includegraphics[width=\linewidth]{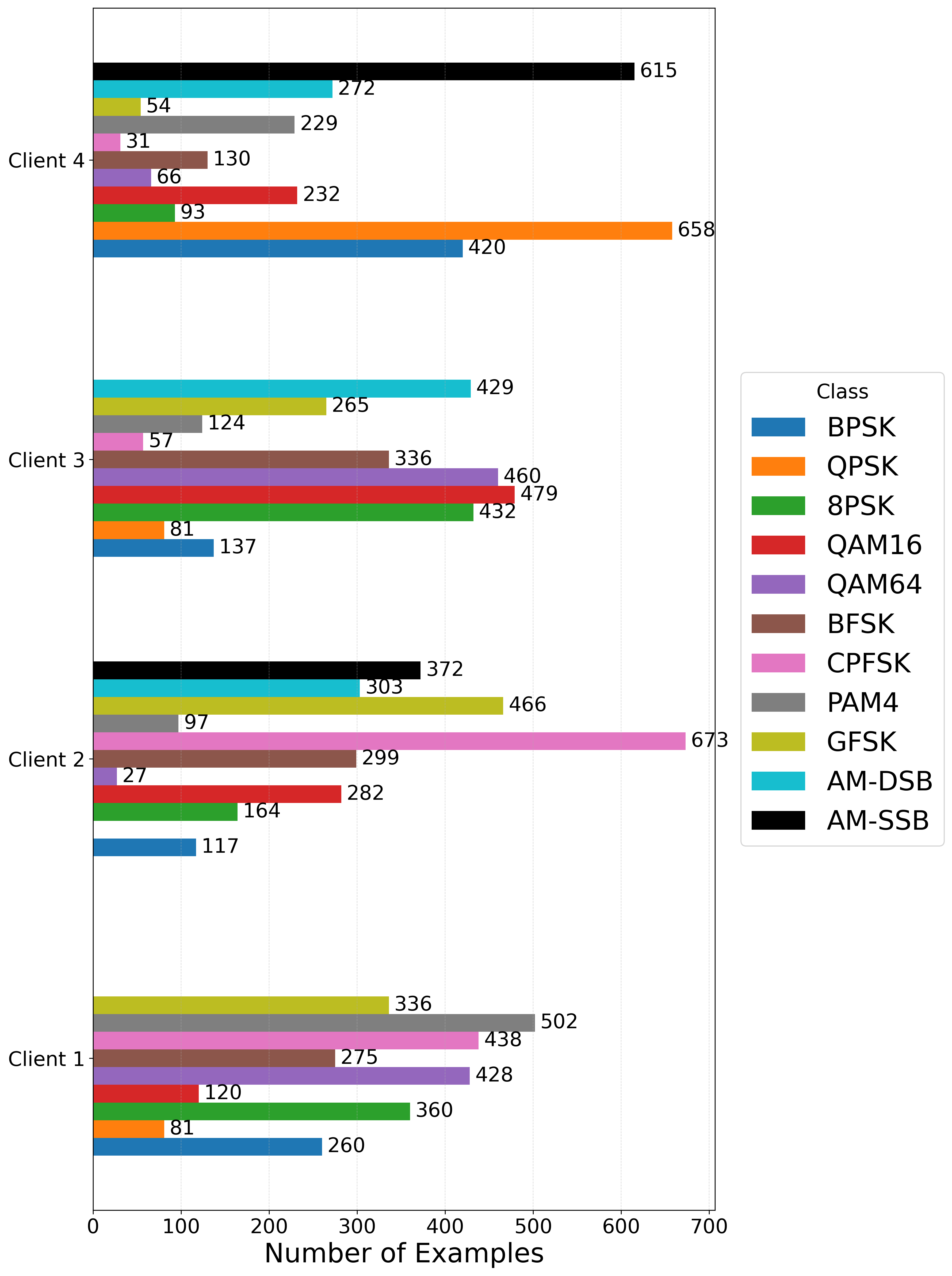}
        \caption{$\bar{\alpha}=1.0$}
    \end{subfigure}

    \caption{Client-wise label distributions in the MIGOU dataset under Dirichlet partitioning for $\bar{\alpha}\in\{0.625,1.0\}$.}
    \label{fig:dist_migou_dirichlet_part2}
\end{figure}

\end{document}